\PassOptionsToPackage{table}{xcolor}
\documentclass[]{fairmeta}

\usepackage{amsmath,amsfonts,bm}

\def\eqref#1{equation~\ref{#1}}

\def\1{\bm{1}}

\DeclareMathAlphabet{\mathsfit}{\encodingdefault}{\sfdefault}{m}{sl}
\SetMathAlphabet{\mathsfit}{bold}{\encodingdefault}{\sfdefault}{bx}{n}

\usepackage{wrapfig}
\usepackage{needspace}
\usepackage{makecell}
\usepackage{array}
\usepackage{arydshln}
\usepackage{xparse}
\usepackage{url}
\usepackage{pifont}

\makeatletter
\newcommand{\legodashvline}{%
  \begingroup
  \color{black}%
  \lower\dp\@arstrutbox\vbox to
    \dimexpr\ht\@arstrutbox+\dp\@arstrutbox\relax{%
      \xleaders\vbox{%
        \hrule width\arrayrulewidth height\dashlinedash
        \kern\dashlinegap
      }\vfill
    }%
  \endgroup
}
\makeatother

\newcommand{\cmark}{\textcolor{green!50!black}{\ding{51}}}
\newcommand{\xmark}{\textcolor{red!70!black}{\ding{55}}}

\makeatletter
\newcommand{\thickhline}{%
  \noalign{\ifnum0=`}\fi\hrule \@height 1.2pt \futurelet\reserved@a\@xhline
}
\makeatother

\newcommand{\legotablestyle}[1][6pt]{%
    \centering
    \scriptsize
    \setlength{\tabcolsep}{#1}%
    \renewcommand{\arraystretch}{1}%
    \rowcolors{2}{gray!11}{white}%
    \ADLinactivate
}
\newcommand{\legotabletoprule}{\thickhline\toprule}
\newcommand{\legotablebottomrule}{\bottomrule\thickhline}
\providecommand{\meanstd}[2]{\mbox{#1\,(#2)}}

\newcounter{finding}
\newcommand{\finding}[1]{%
  \refstepcounter{finding}%
  \begin{tcolorbox}[
    enhanced jigsaw,
    colback=white!90!gray,
    colframe=teal!60!black,
    arc=5pt,
    boxsep=5pt,
    left=1mm,
    right=1mm,
    top=1mm,
    bottom=1mm,
    boxrule=0.8pt,
    drop shadow=gray!50!white
  ]
  \textbf{\textit{Finding \thefinding:}} #1
  \end{tcolorbox}%
}

\definecolor{PromptGray}{gray}{0.85}
\tcbset{
  promptboxstyle/.style={
    enhanced,
    boxrule=0.9pt,
    colback=gray!0,
    colframe=black!50,
    colbacktitle=PromptGray,
    coltitle=black!80,
    fonttitle=\sffamily\bfseries,
    attach boxed title to top center={
      yshift=-3mm,
      yshifttext=-1mm
    },
    boxed title style={
      size=small,
      colframe=PromptGray
    },
    left=2mm,
    right=2mm,
    top=4mm,
    bottom=2mm,
  }
}

\NewDocumentEnvironment{promptfigure}{O{htbp} m m m +b}
{%
  \begin{figure}[#1]
    \centering
    \begin{tcolorbox}[
      promptboxstyle,
      title={#2}
    ]
      \small
      #5
    \end{tcolorbox}
    \caption{#3}
    \label{#4}
  \end{figure}
}
{}

\makeatletter
\def\blfootnote{\xdef\@thefnmark{}\@footnotetext}
\makeatother

\title{LEGO-Anything: Coding Agents for 3D Scene Reconstruction}

\author[1,2,*]{Xirui Li}
\author[2]{Peng Shi}
\author[2]{Mingwen Dong}
\author[2]{Sheng Zhang}
\author[2]{Zhuoyan Xu}
\author[2]{Dongkyu Lee}
\author[2]{Shuaichen Chang}
\author[2]{Yi Xiang}
\author[2]{Lin Pan}
\author[2]{Jiarong Jiang}

\affiliation[1]{University of Maryland, College Park}
\affiliation[2]{AWS}

\contribution[*]{Work done during an internship at AWS}

\abstract{
A 3D scene reconstructed from a single image is most useful when represented not as a rendering or a fixed 3D output, but as an explicit scene program whose execution yields a scene that can be inspected, edited, and queried. We present \textbf{LEGO-Anything}, an \emph{Image-to-Code} framework in which a coding agent builds such a program by iteratively writing and executing Blender code, inspecting the evolving scene and its renderings, and revising the program.
To evaluate how well such agents recover scenes end to end, we introduce \textbf{LEGO-Bench}, a simulator-grounded benchmark with 208 images from 104 diverse scenes spanning indoor and outdoor environments. LEGO-Bench separately scores artifact validity, visible-surface geometry, and rendered appearance, and its simulator-grounded design makes the benchmark naturally extensible while retaining precise automatic evaluation. Across the evaluated coding agents, GPT-6-astra achieves the strongest overall results, with 53.4\% indoor and 39.6\% outdoor scores, yet substantial gaps remain between delivering valid scene artifacts and faithfully recovering scene geometry and appearance.
To understand these failures, we analyze agent construction trajectories and identify three recurring issues: weak scene initialization, regressive edits during iteration, and unreliable self-evaluation. These findings motivate \textbf{LEGO-Plugin}, a training-free harness plugin for more controlled iterative scene construction with relative gains of up to 62.7\% in overall score.
Finally, we ask whether reconstructed scenes are faithful enough to represent natural images and support vision tasks. In \textbf{LEGO-World}, we take scenes reconstructed by GPT-6-astra and derive object detections, instance masks, and relative depth as deterministic queries on each scene. These readouts show non-trivial performance across all three tasks but fall well short of specialized vision models, suggesting that program-constructed scenes from current coding agents are a promising but not yet sufficiently precise representation of natural images.

}

\date{September 2026}
\authoremails{\email{xiruili@umd.edu}, \email{penshi@amazon.com}}
\metadata[Project Page]{\url{https://lego-anything.com}}

\begin{document}

\maketitle

\section{Introduction}
\vspace{-1mm}
The representation a 3D reconstruction takes determines what can be done with it. Compared with meshes~\citep{gkioxari2020meshrcnn}, point maps~\citep{wang2024dust3rgeometric3dvision}, or object sets~\citep{ardelean2025gen3dsrgeneralizable3dscene}, an executable scene program makes objects, geometry, layout, and camera explicit, and can be run, inspected, edited, and queried like any other code. Yet existing approaches rarely adopt this form. Modular pipelines compose specialized components for perception, reconstruction, asset retrieval, and scene assembly~\citep{kim2026sceneconductor3dscenegeneration,kang2026simworldstudioautomaticenvironment}, while learned models map images directly to fixed 3D outputs~\citep{sam3dteam2026sam3d3dfyimages,cao2025physxanythingsimulationreadyphysical3d,dahnert2021panoptic3dscene,zadaianchuk2026reconstructiongeneration}. Neither offers a direct way to check the result against the input and revise it. Building on recent program-based approaches~\citep{hu2024scenecraftllmagentsynthesizing,li2026scenixsparseview3dscene,yin2026visionasinversegraphicsagentinterleavedmultimodal}, we treat single-image scene reconstruction as the iterative construction of an explicit, executable scene program, in which each step is executed, compared with the input image, and refined.

\begin{figure}[t]
\centering
\includegraphics[width=1\linewidth]{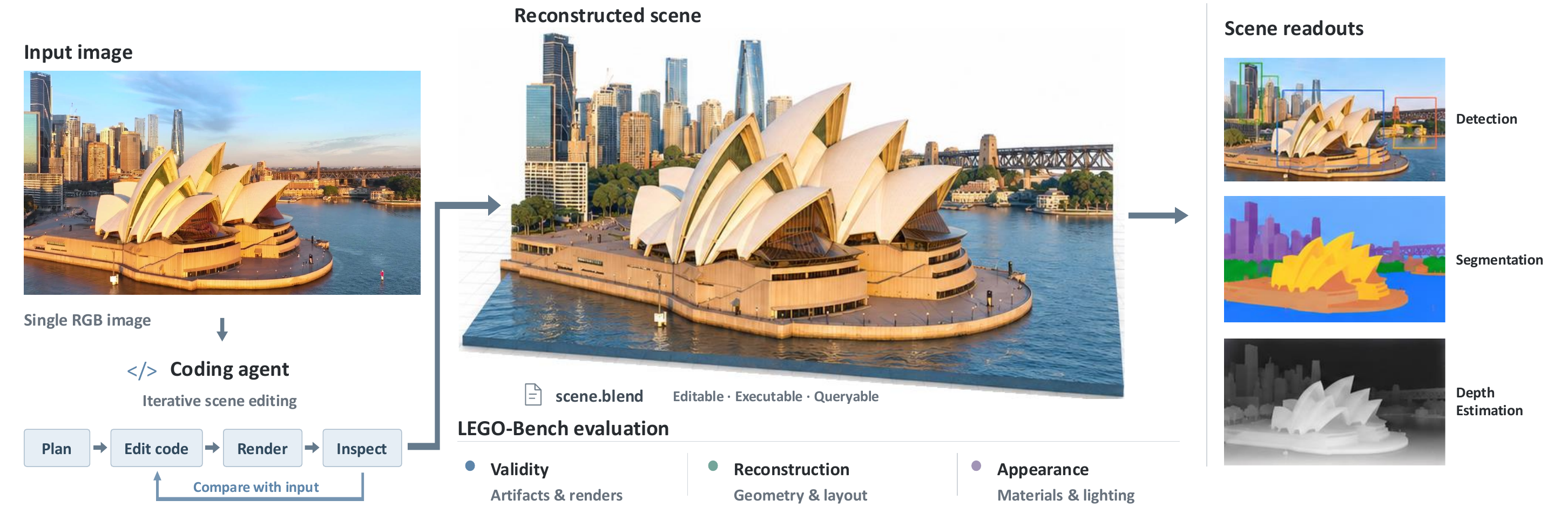}
\caption{\textbf{From image to scene program.} Given a single RGB image, a coding agent iteratively plans, edits Blender code, renders, and inspects the result against the input, producing an editable and executable scene program. The resulting scene program can also serve as a representation that can be queried directly for detection, segmentation, and depth estimation.}
\label{fig:main_graph}
\end{figure}

Coding agents~\citep{anthropic_claude_code_2026, openai_codex_2026} offer a natural way to instantiate this representation: rather than predicting a scene in one shot, they can build it incrementally, using execution feedback to guide each revision~\citep{yin2026visionasinversegraphicsagentinterleavedmultimodal}. We study this setting as \textbf{LEGO-Anything}, an \emph{Image-to-Code (Image2Code)} framework in which a general-purpose coding agent writes and executes Blender programs, inspects the evolving scene and its renderings, and revises the construction to match a single reference image. This formulation lets us study three questions: (1) how faithfully coding agents reconstruct scenes from a single image, (2) what limits their iterative construction process, and (3) whether the resulting scene programs are precise enough to serve as representations of natural images. Figure~\ref{fig:main_graph} illustrates the formulation, from iterative scene construction to querying the reconstructed scene for detection, segmentation, and depth.

To evaluate agentic end-to-end reconstruction, we need a benchmark whose inputs resemble the natural and diverse scenes that users would actually ask a system to reconstruct, while still supporting precise automatic evaluation. Existing benchmarks cover parts of this setting but not all of it: they focus on object- or part-level modeling~\citep{gao20263dcodebenchbenchmarkingagenticprocedural,yang2026p3dbenchbenchmarkingmllmsparametric}, specify worlds through text rather than images~\citep{lu2026worldcoderbenchbenchmarkingphysicallygrounded}, or rely on calibrated multi-view indoor observations~\citep{zhao2026sceneactbenchagentsact3d}. We therefore introduce \textbf{LEGO-Bench}, a simulator-grounded diagnostic benchmark with 208 images rendered from 104 diverse indoor and outdoor simulator scenes. Rendering from fully specified 3D scenes yields natural-image-style inputs while retaining private geometry, depth, and instance masks for evaluation, and allows scene complexity to be varied in a controlled way. LEGO-Bench separately scores artifact validity, visible-surface geometry, and rendered appearance, distinguishing failures to deliver valid scene artifacts from failures of geometric and visual fidelity. Among the evaluated coding agents and scene-construction baselines, \texttt{GPT-6-astra} achieves the strongest overall results, with 53.4\% indoor and 39.6\% outdoor scores.

To diagnose why end-to-end scene reconstruction still falls short, we analyze agent construction trajectories and identify three recurring failures: weak scene initialization, regressive edits during iteration, and unreliable self-evaluation. These findings motivate \textbf{LEGO-Plugin}, a training-free harness plugin with three modules, each targeting one failure: \emph{Enhanced Initialization} grounds the starting scene in the reference image, \emph{Grounded Refinement} replaces unreliable self-judgment with evidence from the reference, and \emph{Version Control} protects previously correct progress from regressive edits. Rather than introducing a separate planner, LEGO-Plugin plugs into existing harnesses as tools, skills, and runtime hooks. It improves all six evaluated models on the LEGO-Bench Office subset, by up to 62.7\% relative, with the largest gains for weaker agents.

We next ask whether reconstructed scenes are precise enough to serve as structured representations of natural images~\citep{zhang2025scene, wang2026codeworldsagenticdiscovery, zhu2022scenegraphgenerationcomprehensive} for downstream vision tasks. To this end, we introduce \textbf{LEGO-World}, an evaluation setting that takes the program reconstructed from a natural image and derives object detections, instance masks, and relative depth through deterministic readouts. Without any task-specific training, these readouts achieve non-trivial performance on all three tasks but remain well below SOTA vision models, suggesting that executable scenes from current coding agents are a promising but not yet sufficiently precise representation of natural images.

\vspace{-1mm}

\section{Related Work}
\subsection{Coding Agents}

Coding agents combine language models with tools for code editing, execution, and iterative verification
\citep{anthropic_claude_code_2026, openai_codex_2026,liu2026diveclaudecodedesign}.
Their harnesses organize tool access, execution context, and environmental feedback, allowing code to serve not only as an output, but also as an interface for planning and interaction~\citep{ning2026codeagentharness}.
Recent work extends this paradigm to visual and 3D settings. VIGA reconstructs and edits visual programs through a code--render--inspect loop~\citep{yin2026visionasinversegraphicsagentinterleavedmultimodal}; SEIG reconstructs images as editable Blender programs through staged executable inverse graphics~\citep{he2026thinkingblenderstagedexecutable}; ArtiCraft~\citep{zhou2026articraftagenticscalablearticulated} develops an agentic coding interface for articulated 3D assets; and Code-CoT~\citep{liu20263dprimitivesspatiallanguage} uses executable 3D programs as intermediate representations for spatial reasoning.
In contrast, we study general-purpose coding agents as a testbed for single-image executable scene construction, emphasizing scene fidelity, iterative failure modes, and the perceptual sufficiency of frozen reconstructed scenes.

\subsection{3D Scene Construction and Evaluation}

Prior work addresses important components of scene construction under different assumptions. Procedural systems generate natural, indoor, and urban environments from structured specifications and reusable assets~\citep{deitke2022procthorlargescaleembodiedai, raistrick2023infinitephotorealisticworldsusing, raistrick2024infinigenindoorsphotorealisticindoor, deng2024citycraftrealcrafter3d}, while image-conditioned methods recover geometry and appearance from visual observations~\citep{cao2026collaborativemultimodalcodinghighquality, sam3dteam2026sam3d3dfyimages}. Other work targets editable indoor scenes from RGB-D scans~\citep{huang2026literealitygraphicsready3dscene} or simulation-ready articulated assets~\citep{jiang2022dittobuildingdigitaltwins, liu2023parispartlevelreconstructionmotion, wu2026urdfanythingendtoendgenerationsimulationready, zhang2026simartdecomposingmonolithicmeshes, le2025articulateanythingautomaticmodelingarticulated, cao2025physxanythingsimulationreadyphysical3d}. These directions provide increasingly capable components, but do not directly answer whether a single natural image can be reconstructed as a faithful executable scene.

Existing benchmarks leave different parts of this setting untested. 3DCodeBench~\citep{gao20263dcodebenchbenchmarkingagenticprocedural} focuses on procedural object modeling, P3D-Bench~\citep{yang2026p3dbenchbenchmarkingmllmsparametric} targets parametric parts and assemblies, and WorldCoder-Bench~\citep{lu2026worldcoderbenchbenchmarkingphysicallygrounded} evaluates text-specified interactive worlds rather than fidelity to a reference image. SEIG is closely related in its executable Blender-program formulation, but its quantitative evaluations are object-centric. SceneActBench~\citep{zhao2026sceneactbenchagentsact3d} comes closest, but its reconstruction track uses calibrated multi-view indoor observations and excludes room structure and texture recovery. \textbf{LEGO-Bench} instead evaluates single-image, end-to-end executable scene reconstruction across diverse and realistic environments, separating artifact validity, visible-surface reconstruction, and rendered appearance. See Table~\ref{tab:related_work_comparison} and Appendix~\ref{app:rw_3d_benchmarks} for additional comparisons.

\begin{table*}[t]
    \legotablestyle
    \caption{\textbf{Comparison of related 3D benchmarks.}
    \cmark: supported; \xmark: not supported. LEGO-Bench uniquely targets end-to-end evaluation of executable scene reconstruction across diverse, realistic scenes with controlled difficulty.}
    \label{tab:related_work_comparison}

    \resizebox{\textwidth}{!}{%
        \begin{tabular}{l|cccccc}
            \legotabletoprule
            \textbf{Benchmark}
            & \textbf{Image Input}
            & \textbf{Scene-level Eval.}
            & \textbf{Executable / Interactive}
            & \textbf{Indoor/Outdoor Coverage}
            & \textbf{Natural-Image Style}
            & \textbf{Controlled Difficulty} \\
            \midrule

            \texttt{3DCodeBench}~\citep{gao20263dcodebenchbenchmarkingagenticprocedural}
            & \cmark & \xmark & \xmark & \xmark & \xmark & \xmark \\

            \texttt{WorldCoder-Bench}~\citep{lu2026worldcoder}
            & \xmark & \cmark & \cmark & \xmark & \xmark & \xmark \\

            \texttt{P3D-Bench}~\citep{yang2026p3dbenchbenchmarkingmllmsparametric}
            & \cmark & \xmark & \xmark & \xmark & \xmark & \xmark \\

            \texttt{SEIG}~\citep{he2026thinkingblenderstagedexecutable}
            & \cmark & \xmark & \cmark & \xmark & \xmark & \xmark \\

            \texttt{SceneActBench}~\citep{zhao2026sceneactbenchagentsact3d}
            & \cmark & \cmark & \cmark & \xmark & \xmark & \xmark \\

            \textbf{\texttt{LEGO-Bench} (Ours)}
            & \cmark & \cmark & \cmark & \cmark & \cmark &\cmark \\

            \legotablebottomrule
        \end{tabular}%
    }
\end{table*}

\section{LEGO-Anything: Scene Reconstruction as Image-to-Code}
\label{sec:lego-anything}

\begin{figure}[t]
\centering
\includegraphics[width=\linewidth]{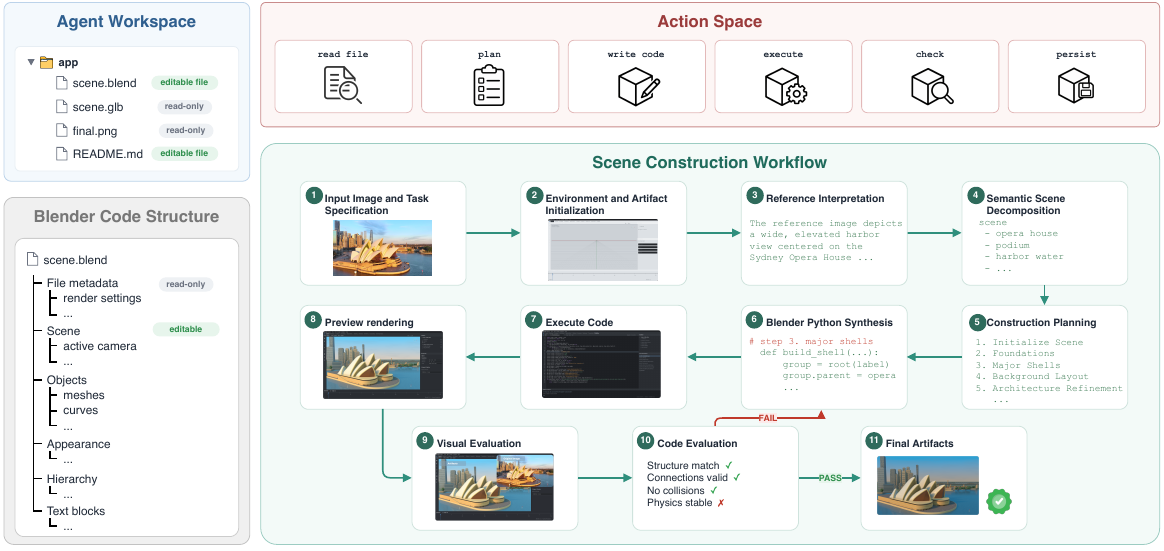}
\caption{\textbf{Agentic scene construction, abstracted from a \texttt{GPT-6-astra} trajectory.} Four blocks define the process: the Agent Workspace manages artifacts, the Action Space exposes agent operations, the Blender Code Structure organizes editable scene state, and the Scene Construction Workflow iterates from reference interpretation to validated final artifacts. Together, they enable executable, feedback-driven scene reconstruction.}
\label{fig:workflow}
\end{figure}

LEGO-Anything formulates single-image scene reconstruction as an \emph{Image-to-Code (Image2Code)} problem. Given a single RGB image $I$, a general-purpose coding agent $\pi$ interacts with a 3D editing environment $\mathcal{E}$ to produce a scene program $P$, whose execution constructs a 3D scene $S$:
\begin{equation}
P = \pi(I;\mathcal{E}), \qquad
S = \operatorname{Exec}_{\mathcal{E}}(P).
\end{equation}
In our setting, $\mathcal{E}$ is Blender. Rather than producing $P$ in one shot, the agent alternates between editing code, executing it, and inspecting the resulting scene and renderings, yielding a \emph{construction trajectory} $\tau = \{(P_t, S_t, o_t)\}_{t=1}^{T}$ of intermediate programs, scenes, and observations, with $P = P_T$. The agent submits the final scene, its export, and a rendered view, which together form an executable artifact that can be evaluated for fidelity and queried for downstream perception. Figure~\ref{fig:workflow} illustrates this workflow as abstracted from a GPT-6-astra trajectory. We evaluate final scenes in LEGO-Bench (Section~\ref{sec:LEGO-Bench}), analyze trajectories to diagnose failures and design LEGO-Plugin (Section~\ref{sec:lego_plugin}), and query frozen scenes in LEGO-World (Section~\ref{sec:lego-world}).

\section{LEGO-Bench: A Simulator-Grounded Benchmark for End-to-End Scene Reconstruction}
\label{sec:LEGO-Bench}

LEGO-Bench evaluates end-to-end 3D scene reconstruction in a setting that mirrors real use: given a single natural image, an agent must reconstruct the scene as a complete, executable 3D artifact, jointly recovering its contents, geometry, spatial layout, and appearance from one view. Because every case is rendered from a fully specified simulator scene, ground truth comes for free, and the benchmark scales naturally: users can extend it with their own scenes and assets through the same construction pipeline while retaining precise automatic evaluation.

\subsection{Benchmark Design and Construction}

Real photographs lack precise 3D ground truth, while simple synthetic scenes lack realism. LEGO-Bench resolves this tension by rendering inputs from professionally authored simulator scenes, which (1) provides realistic, natural-image-style observations, (2) retains private geometry, object identities, camera parameters, depth, and instance masks for automatic evaluation, and (3) allows scene complexity to be controlled directly.

\begin{figure}[t]
    \centering
    \begin{minipage}[c]{0.70\linewidth}
        \centering

        \includegraphics[
            width=\linewidth
        ]{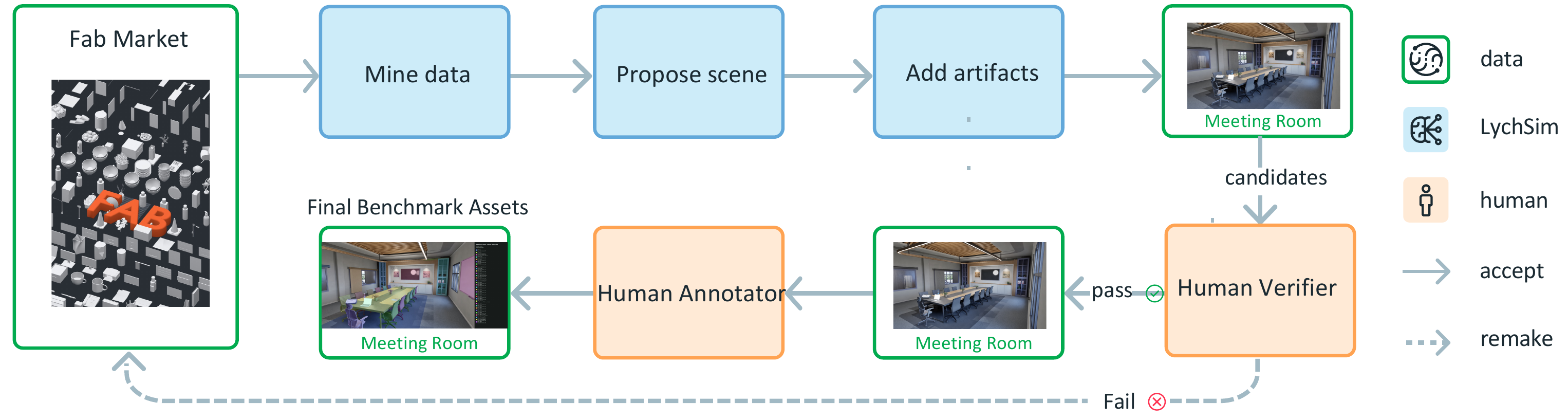}

        \captionof{figure}{\textbf{Extensible benchmark-construction pipeline.} Fab assets are mined and assembled into candidate scenes; human verification triggers either remaking or annotation before acceptance. This human-in-the-loop process scales benchmark construction while retaining simulator ground truth and scene quality.}
        \label{fig:lego-bench-pipeline}
    \end{minipage}
    \hfill
    \begin{minipage}[c]{0.27\linewidth}
        \legotablestyle
        \captionof{table}{\textbf{Statistics of LEGO-Bench.} The benchmark spans 104 scenes, eight environments, 17 themes, and 443 assets.}
        \label{tab:lego_bench_statistics}

        \resizebox{\linewidth}{!}{%
        \begin{tabular}{l|c}
            \legotabletoprule
            \textbf{Statistic} & \textbf{LEGO-Bench} \\
            \midrule
            $\#$ RGB Inputs
                & 208 \\
            \midrule
            \hspace{1em}$\#$ Environments
                & 8 \\
            \hspace{1em}$\#$ Themes
                & 17 \\
            \hspace{1em}$\#$ Logical Scenes
                & 104  \\
            \hspace{1em}$\#$ Registered Assets
                & 443 \\
            \legotablebottomrule
        \end{tabular}%
        }

    \end{minipage}
\end{figure}

As shown in Figure~\ref{fig:lego-bench-pipeline}, we build LEGO-Bench in LychSim~\citep{ma2026lychsimcontrollableinteractivesimulation} using scene and asset packs from Fab, and organize indoor and outdoor scenes into themes with nested object sets (Easy $\subset$ Medium $\subset$ Hard). We select only Fab assets whose usage terms allow AI use. Within each theme, architecture, materials, lighting, and camera remain fixed while visible scene content increases, allowing matched comparisons across difficulty levels. Candidate scenes undergo collision and stability checks and human review before inclusion. Because this pipeline only requires registered assets and a scene specification, LEGO-Bench scales with available asset libraries: new environments, difficulty levels, and views can be added without changing the evaluation protocol. As an example, we include an NYC aerial split without difficulty pairing as a large-scale reconstruction stress test (Appendix~\ref{app:birds_eye_stress}).

LEGO-Bench contains 208 RGB inputs from 104 scenes across 8 environments and 17 themes, using 443 registered assets (Table~\ref{tab:lego_bench_statistics}). Through the construction pipeline in Figure~\ref{fig:lego-bench-pipeline}, users can convert their own scenes and assets into new evaluation cases with the same ground truth and scoring protocol. Each trial provides an RGB image and public metadata, including image dimensions, horizontal field of view, a category taxonomy, and an output schema, while reference geometry, depth, instance masks, object correspondences, and scene transforms remain private to the evaluator.

\subsection{Evaluation Metrics}

We evaluate each trial along three complementary axes: submission \textbf{V}alidity $V_i$, visible-surface \textbf{R}econstruction $R_i$, and rendered \textbf{A}ppearance $A_i$. Together, these metrics distinguish failure to deliver an evaluable scene from failures of geometric and visual fidelity. Appendix~\ref{app:metric_details} provides implementation details, auxiliary metric protocols, and a human-validation study of the headline metrics (Appendix~\ref{app:human_metric_alignment}).

\paragraph{\textbf{Validity.}}
Validity checks whether the submission delivers usable scene artifacts: \texttt{scene.blend} must open and contain renderable geometry and an active camera, \texttt{scene.glb} must be a well-formed, non-empty export, and \texttt{final.png} must be a parseable, non-degenerate image. Submissions that fail these checks, or for which the headline evaluation cannot be completed, receive $V_i=0$ and $R_i=A_i=0$. Low reconstruction quality alone does not make a submission invalid.

\paragraph{\textbf{Reconstruction.}}
Reconstruction measures geometric agreement on object surfaces visible in the reference view. We extract visible surfaces from the submitted scene under its active camera and derive reference-visible surfaces from private depth. Both point sets are projected onto the image plane and partitioned into scored objects by the private instance masks, then compared directly in camera coordinates without alignment or rescaling. For each object $o$, precision is the fraction of submitted points within a depth-scaled tolerance $\tau(g)=0.05\,z(g)$ of the nearest reference point $g$, where $z(g)$ is its forward depth, and recall is defined symmetrically. The Reconstruction score (F@5\%) averages object-level F1 scores:
\begin{equation}
F_{i,o} = \frac{2\,\mathrm{Prec}_{i,o}\,\mathrm{Rec}_{i,o}}{\mathrm{Prec}_{i,o}+\mathrm{Rec}_{i,o}}, \qquad
R_i = \frac{1}{|\mathcal{O}_i|}\sum_{o\in\mathcal{O}_i} F_{i,o},
\label{eq:reconstruction}
\end{equation}
where $\mathcal{O}_i$ contains the annotated scored objects (Appendix~\ref{app:reconstruction_scoring}), and $F_{i,o}=0$ when its denominator is zero. Missing surfaces reduce recall, while extra geometry within a scored mask reduces precision. Appendix~\ref{app:reconstruction_scoring} also reports stricter F@2\% and relaxed F@10\% variants.
\paragraph{\textbf{Appearance.}}
Appearance evaluates visual similarity between the reconstructed scene and the reference image. Rather than scoring the agent's own \texttt{final.png}, the evaluator re-renders the submitted scene, preserving its camera, geometry, materials, lighting, and color settings while standardizing the rendering engine and output resolution. Let $\hat{I}_i$ and $I_i$ denote the evaluator render and reference image in 8-bit sRGB, with pixel domain $\Omega_i$. Appearance is the fraction of pixels whose error does not exceed a threshold $\delta$ in any channel:
\begin{equation}
A_i =
\frac{1}{|\Omega_i|}
\sum_{u\in\Omega_i}
\mathbf{1}\!\left[
    \left\|\hat{I}_i(u)-I_i(u)\right\|_{\infty}
    \leq \delta
\right].
\label{eq:appearance}
\end{equation}
This metric captures the combined effect of scene geometry, materials, lighting, and camera configuration on the rendered view. We set $\delta=30$ (on a 0-255 scale) based on our sensitivity analysis (Appendix~\ref{app:appearance_metric}).
\vspace{-1mm}
\paragraph{\textbf{Overall score.}}
Valid submissions receive the mean of Reconstruction and Appearance.
Artifact-invalid submissions and unresolved evaluator failures receive zero.
The benchmark score averages over all $N$ attempted cases, including failures:

\begin{equation}
S = \frac{1}{N}
\sum_{\substack{i=1}}^{N} \frac{1}{2} (R_i + A_i) * V_i.
\label{eq:overall_score}
\end{equation}

\subsection{Main Results}
\begin{table*}[t]
    \legotablestyle
    \caption{
        \textbf{Performance ($\%$) on LEGO-Bench.} Higher is better for every metric. As a group, general-purpose coding agents reliably produce valid artifacts across both Indoor and Outdoor splits, with \texttt{GPT-6-astra} achieving the strongest overall performance.
    }
    \label{tab:lego-bench_model_and_harness}
    \resizebox{\textwidth}{!}{
        \begin{tabular}{l|ccc:c|ccc:c}
            \legotabletoprule
            & \multicolumn{4}{c|}{\textbf{Indoor}}
            & \multicolumn{4}{c}{\textbf{Outdoor}} \\
            \cmidrule(lr){2-5}
            \cmidrule(lr){6-9}

            \textbf{Model (+ Harness)}
            & \textbf{V $\uparrow$}
            & \textbf{R $\uparrow$}
            & \textbf{A $\uparrow$}
            & \textbf{S $\uparrow$}
            & \textbf{V $\uparrow$}
            & \textbf{R $\uparrow$}
            & \textbf{A $\uparrow$}
            & \textbf{S $\uparrow$} \\
            \midrule

            \hiderowcolors
            \multicolumn{9}{c}{
                \textcolor{gray}{\textit{General-Purpose Coding Agents}}
            } \\
            \showrowcolors
            \texttt{GPT-6-astra} + \texttt{Codex}
                & \meanstd{\textbf{100.0}}{0.0}
                & \meanstd{52.4}{1.1}
                & \meanstd{\textbf{54.4}}{0.7}
                & \meanstd{\textbf{53.4}}{0.8}
                & \meanstd{98.0}{1.0}
                & \meanstd{34.0}{1.5}
                & \meanstd{\textbf{45.5}}{0.5}
                & \meanstd{\textbf{39.6}}{0.8} \\

            \texttt{GPT-6-sol} + \texttt{Codex}
                & \meanstd{99.4}{1.1}
                & \meanstd{22.2}{2.4}
                & \meanstd{42.5}{0.3}
                & \meanstd{32.3}{1.0}
                & \meanstd{99.7}{0.6}
                & \meanstd{19.8}{1.0}
                & \meanstd{28.7}{0.5}
                & \meanstd{24.2}{0.4} \\

            \texttt{GPT-6-luna} + \texttt{Codex}
                & \meanstd{99.7}{0.5}
                & \meanstd{15.8}{1.2}
                & \meanstd{30.5}{0.3}
                & \meanstd{23.2}{0.7}
                & \meanstd{99.7}{0.6}
                & \meanstd{13.2}{0.6}
                & \meanstd{21.4}{0.5}
                & \meanstd{17.3}{0.5} \\

            \texttt{GPT-5.6-sol} + \texttt{Codex}
                & \meanstd{97.8}{2.3}
                & \meanstd{9.8}{2.1}
                & \meanstd{19.8}{0.6}
                & \meanstd{14.8}{1.0}
                & \meanstd{98.7}{0.6}
                & \meanstd{12.8}{1.8}
                & \meanstd{17.7}{0.4}
                & \meanstd{15.3}{0.8} \\

            \texttt{GPT-5.6-terra} + \texttt{Codex}
                & \meanstd{99.7}{0.5}
                & \meanstd{9.5}{1.0}
                & \meanstd{21.3}{0.5}
                & \meanstd{15.4}{0.6}
                & \meanstd{98.0}{1.7}
                & \meanstd{8.1}{0.7}
                & \meanstd{15.0}{0.9}
                & \meanstd{11.5}{0.4} \\

            \texttt{GPT-5.6-luna} + \texttt{Codex}
                & \meanstd{99.1}{0.0}
                & \meanstd{10.2}{2.1}
                & \meanstd{18.0}{0.7}
                & \meanstd{14.1}{1.3}
                & \meanstd{97.0}{1.0}
                & \meanstd{7.8}{1.5}
                & \meanstd{16.6}{0.8}
                & \meanstd{12.1}{1.0} \\
            \midrule
            \hiderowcolors
            \multicolumn{9}{c}{
                \textcolor{gray}{\textit{Task-Specific LLM Systems}}
            } \\
            \showrowcolors
            \texttt{VIGA~\citep{yin2026visionasinversegraphicsagentinterleavedmultimodal}}
            & \meanstd{\textbf{100.0}}{0.0}
            & \meanstd{10.8}{1.5}
            & \meanstd{20.3}{0.7}
            & \meanstd{15.6}{1.1}
            & \meanstd{\textbf{100.0}}{0.0}
            & \meanstd{8.0}{0.4}
            & \meanstd{17.1}{0.0}
            & \meanstd{12.6}{0.2} \\

            \texttt{SceneConductor~\citep{kim2026sceneconductor3dscenegeneration}}
            & \meanstd{12.3}{2.3}
            & \meanstd{5.0}{2.1}
            & \meanstd{0.0}{0.0}
            & \meanstd{0.3}{0.2}
            & -- & -- & -- & -- \\

            \midrule
            \hiderowcolors
            \multicolumn{9}{c}{
                \textcolor{gray}{\textit{Single-Image Scene Construction Baselines}}
            } \\
            \showrowcolors
            \texttt{3D-RE-GEN~\citep{sautter20253dregen3dreconstructionindoor}}
            & \meanstd{\textbf{100.0}}{0.0}
            & \meanstd{1.2}{0.1}
            & \meanstd{11.3}{0.2}
            & \meanstd{6.3}{0.1}
            & -- & -- & -- & -- \\

            \texttt{Gen3DSR~\citep{ardelean2025gen3dsrgeneralizable3dscene}}
            & \meanstd{92.0}{1.1}
            & \meanstd{\textbf{65.4}}{0.7}
            & \meanstd{0.0}{0.0}
            & \meanstd{30.1}{0.4}
            & \meanstd{80.0}{1.0}
            & \meanstd{\textbf{35.0}}{0.5}
            & \meanstd{0.0}{0.0}
            & \meanstd{14.0}{0.1} \\

            \texttt{REST3D~\citep{ma2026rest3dreconstructingphysicallystable}}
            & \meanstd{77.5}{13.0}
            & \meanstd{7.2}{0.7}
            & \meanstd{0.0}{0.0}
            & \meanstd{2.8}{0.4}
            & -- & -- & -- & -- \\

            \texttt{SceneGen~\citep{xia2025scenegenagentpreciseindustrialscene}}
            & \meanstd{71.6}{3.0}
            & \meanstd{9.7}{1.8}
            & \meanstd{0.0}{0.0}
            & \meanstd{3.4}{0.5}
            & \meanstd{38.7}{1.5}
            & \meanstd{2.6}{1.1}
            & \meanstd{0.0}{0.0}
            & \meanstd{0.5}{0.2} \\

            \legotablebottomrule
        \end{tabular}
    }
\end{table*}

\paragraph{\textbf{Experimental setup.}} We evaluate GPT-series coding agents under the Codex harness~\citep{codexagentloop}, using each model's native harness behavior. All methods run in a shared execution environment based on the Harbor Framework~\citep{Harbor_Framework}, with \texttt{Blender 5.0.1}~\citep{blender501} and \texttt{Blender-MCP}~\citep{blender_mcp}. We report mean and standard deviation over three runs; Appendix~\ref{app:experiment_details} details baseline-specific configurations.
\paragraph{\textbf{How well do coding agents reconstruct 3D scenes?}}
Table~\ref{tab:lego-bench_model_and_harness} reveals a clear gap between artifact delivery and faithful scene reconstruction. All six GPT configurations achieve near-saturated Validity, yet fidelity differs sharply by model: overall scores range from $53.4\%$/$39.6\%$ for \texttt{GPT-6-astra} to $15.4\%$/$15.3\%$ for the strongest \texttt{GPT-5.6} configurations. Among baselines, Gen3DSR attains higher Reconstruction than \texttt{GPT-6-astra} ($65.4\%$ vs.\ $52.4\%$ indoors) but produces no evaluable appearance, while VIGA and 3D-RE-GEN deliver valid artifacts with low fidelity. Three of six baselines do not support outdoor scenes, whereas all coding agents run on both splits with stable Validity, although fidelity drops outdoors for every model.

\finding{\texttt{GPT-6-astra} achieves the best overall score, surpassing task-specific and single-image scene-construction baselines. Coding agents deliver valid artifacts across indoor and outdoor scenes.}

\Needspace{10\baselineskip}
\vspace{-1mm}
\paragraph{\textbf{How does scene complexity affect performance?}}

\begin{wraptable}[18]{r}{0.4\textwidth}
    \vspace{-0.5\baselineskip}
    \legotablestyle[2pt]
    \captionsetup{skip=3pt}
    \caption{\textbf{Performance across scene-complexity tiers.}}
    \label{tab:scene_complexity}
    \resizebox{\linewidth}{!}{
    \begin{tabular}{l|cc|cc}
        \legotabletoprule
        & \multicolumn{2}{c|}{\textbf{Indoor}}
        & \multicolumn{2}{c}{\textbf{Outdoor}} \\
        \cmidrule(lr){2-3}
        \cmidrule(lr){4-5}
        \textbf{Tier}
        & \textbf{R} & \textbf{A}
        & \textbf{R} & \textbf{A} \\
        \midrule
        Easy
        & \meanstd{23.3}{0.1} & \meanstd{30.4}{0.6}
        & \meanstd{18.4}{0.9} & \meanstd{25.3}{0.4} \\
        Medium
        & \meanstd{19.7}{1.2} & \meanstd{27.9}{0.6}
        & \meanstd{14.2}{1.0} & \meanstd{22.8}{0.9} \\
        Hard
        & \meanstd{18.8}{0.4} & \meanstd{27.4}{0.5}
        & \meanstd{12.7}{1.0} & \meanstd{22.4}{0.3} \\
        \legotablebottomrule
    \end{tabular}
    }
\end{wraptable}

Increasing scene complexity degrades fidelity rather than executability. Averaged over all \texttt{GPT-6} and \texttt{GPT-5.6} configurations, Validity stays at $99.5\%$ across tiers, while the overall score drops from $24.6\%$ (Easy) to $21.3\%$ (Medium) and $20.5\%$ (Hard), with consistent declines in Reconstruction and Appearance on both splits (Table~\ref{tab:scene_complexity}). The drop is steepest for outdoor Reconstruction ($18.4\% \rightarrow 12.7\%$).

\WFclear
\finding{Greater scene complexity lowers both Reconstruction and Appearance while leaving Validity intact, and outdoor scenes are harder than indoor scenes at every complexity tier.}

\Needspace{19\baselineskip}
\vspace{-1mm}
\paragraph{\textbf{Does more reasoning improve scene reconstruction?}}

\begin{wrapfigure}{r}{0.3\textwidth}
    \vspace{-0.5\baselineskip}
    \centering
    \includegraphics[width=\linewidth]{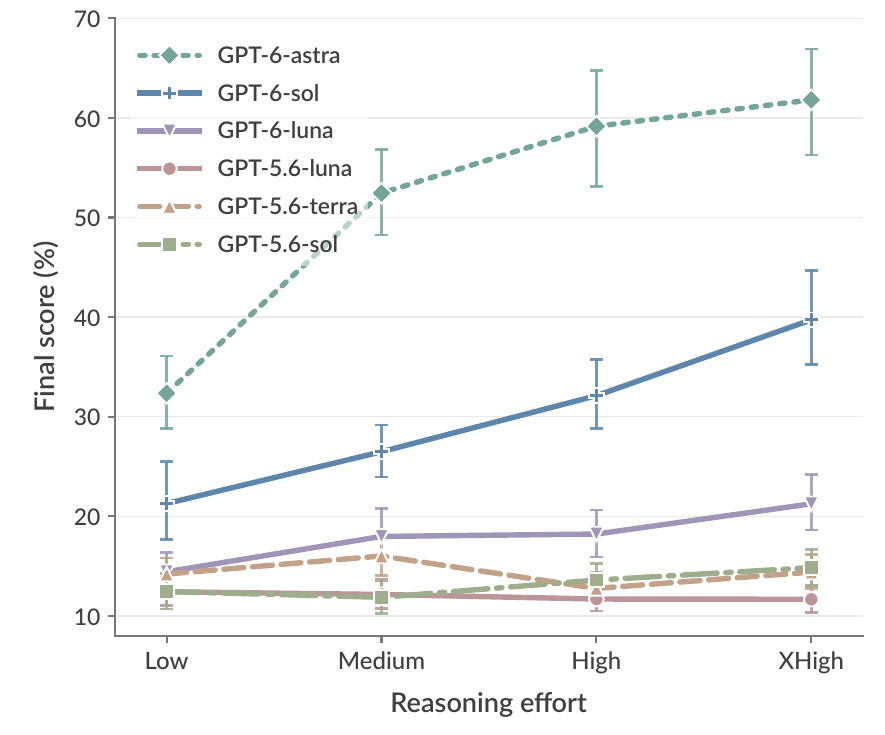}
    \captionsetup{skip=2pt}
    \caption{\textbf{Test-time scaling improves \texttt{GPT-6} performance.}}
    \label{fig:lego-bench_reasoning_efforts}
\end{wrapfigure}

We run a test-time scaling experiment on a fixed 42-case Office subset, varying the harness-native reasoning effort (Low, Medium, High, XHigh) while keeping inputs, prompts, execution budgets, and scoring fixed. For all three \texttt{GPT-6} models, the overall score rises with effort (Figure~\ref{fig:lego-bench_reasoning_efforts}): \texttt{astra} from $32.3\%$ to $61.8\%$, \texttt{sol} from $21.3\%$ to $39.7\%$, and \texttt{luna} from $14.4\%$ to $21.2\%$. Gains grow with model strength, and \texttt{astra} saturates toward XHigh while \texttt{sol} keeps improving. By contrast, \texttt{GPT-5.6} models show weak or non-monotonic changes (Appendix~\ref{app:reasoning_effort_protocol}).

\WFclear
\noindent\rule{0pt}{8mm}\par
\finding{Higher reasoning effort improves all three \texttt{GPT-6} models, with larger gains for stronger models, while \texttt{GPT-5.6} models show no consistent improvement.}

\section{LEGO-Plugin: From Failure Diagnosis to Reliable Scene Reconstruction}
\label{sec:lego_plugin}

\vspace{-1mm}
LEGO-Bench scores the submitted scene but collapses the construction process into a single outcome. A poor result may stem from a weak or delayed initial scene, edits that undo earlier progress, or an agent's failure to judge whether its scene is improving. We therefore analyze intermediate artifacts and test whether agents can reliably evaluate their own artifacts. These diagnoses motivate \textbf{LEGO-Plugin}, a training-free harness plugin whose three modules, \emph{Enhanced Initialization}, \emph{Version Control}, and \emph{Grounded Refinement}, each target one failure.
\vspace{-1mm}

\subsection{Diagnosing Construction Failures}
\label{sec:failure_diagnosis}
\begin{figure*}[t]
\centering
\includegraphics[width=\textwidth]{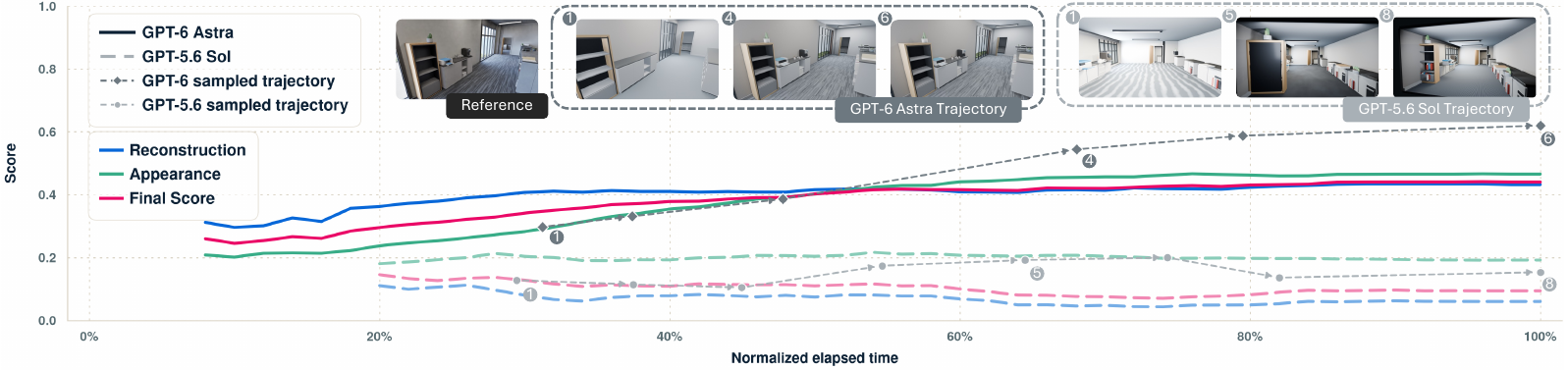}
\caption{\textbf{Scene quality over construction time.} \texttt{GPT-6-astra} reaches an evaluable scene earlier and refines it steadily, whereas \texttt{GPT-5.6-sol} starts later and regresses more often. The gap shows that reliable construction depends on fast initialization and protection against regressive edits.}
\label{fig:construction_trajectory}
\vspace{-3mm}
\end{figure*}

\paragraph{\textbf{Trajectory analysis.}}
We re-evaluate every renderable intermediate artifact in the trajectories from our main experiments with the LEGO-Bench metrics. This yields three trajectory-level quantities: time to the first evaluable scene, the frequency of score-decreasing edits, and the gap between the best intermediate score and the final submission. We contrast \texttt{GPT-6-astra} and \texttt{GPT-5.6-sol} here and report all four models in Appendix~\ref{app:four_model_trajectory}. As shown in Figure~\ref{fig:construction_trajectory}, \texttt{GPT-6-astra} reaches its first evaluable scene within roughly the first tenth of its budget, whereas \texttt{GPT-5.6-sol} needs about one fifth, leaving less budget for render-based refinement. After initialization, \texttt{GPT-6-astra} improves more steadily, while \texttt{GPT-5.6-sol} often stagnates or declines: 29.6\% of its edits decrease the score, and its final submission trails its best intermediate scene by 3.2 points. These regressions stem from later edits that damage geometry, camera settings, or a previously stronger scene.

\finding{Reconstruction quality is non-monotonic over the construction
trajectory: weaker agents take longer to produce an evaluable scene, and
subsequent edits can still degrade scene fidelity.}

\paragraph{\textbf{Can coding agents judge their own scenes?}}
\label{sec:vlm_judge}

\begin{wrapfigure}{r}{0.45\textwidth}
\vspace{-0.5\baselineskip}
\centering
\includegraphics[width=\linewidth]{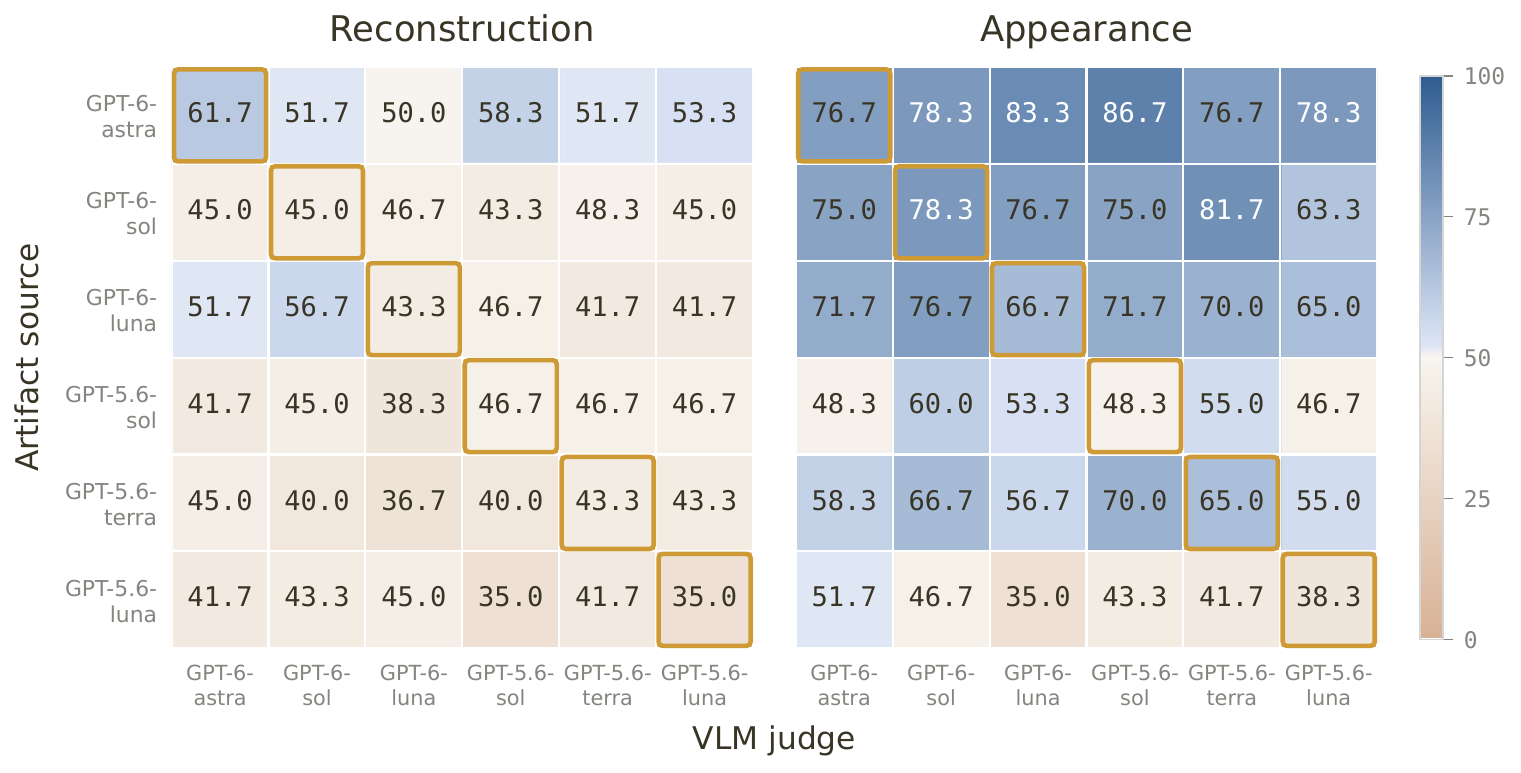}
\captionsetup{skip=2pt}
\caption{\textbf{Judge agreement with metric directions.} Reconstruction judgments remain near or below chance, and self-judging provides no consistent advantage over cross-model judging.}
\label{fig:judge_agreement}
\vspace{-0.5\baselineskip}
\end{wrapfigure}

We next test whether agents can tell when a scene improves. Given two renderings of the same case, a judge model picks the better one, and we measure agreement with the direction given by deterministic LEGO-Bench metrics (chance: $50\%$). Across all 36 builder--judge pairs of six models, self-judgments agree $45.8\%$ of the time on Reconstruction and $62.2\%$ on Appearance, versus $45.4\%$ and $63.9\%$ for cross-model judgments. Judges are thus near or below chance on geometry, and self-judging beats the mean of the other five judges for only 3 of 6 builders on Reconstruction and 2 of 6 on Appearance (Figure~\ref{fig:judge_agreement}; protocol in Appendix~\ref{app:vlm_judge_protocol}).

\WFclear
\noindent\rule{0pt}{3mm}\par
\finding{Coding agents cannot reliably judge scene quality, especially geometry, and self-judging offers no advantage over cross-model judging. Refinement should therefore be grounded in deterministic evidence rather than self-assessment.}

\subsection{LEGO-Plugin}
\label{sec:plugin_method}

\begin{figure}[tbp]
\centering
\includegraphics[width=\linewidth]{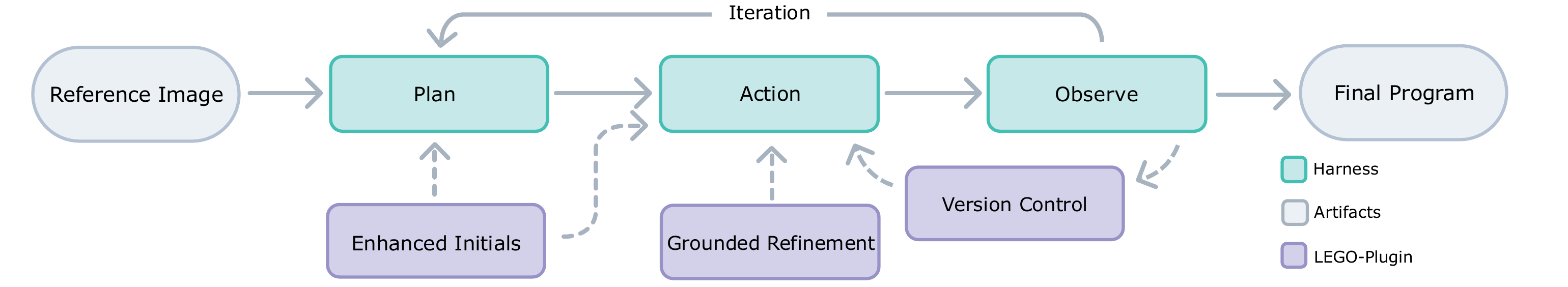}
\caption{\textbf{LEGO-Plugin.} A harness-compatible plugin adds Enhanced Initialization, Grounded Refinement, and Version Control to the vanilla coding-agent harness. These modules directly address weak initialization, unreliable self-evaluation, and regressive edits without changing the underlying agent.}
\label{fig:lego_plugin_design}
\end{figure}

LEGO-Plugin is a training-free control layer exposed as MCP tools, workflow skills, and runtime hooks, rather than a separate planner or memory module. The agent remains the planner and Blender MCP remains the scene editor, so LEGO-Plugin stays compatible with evolving harnesses while preserving native agent behavior. All plugin signals are derived from the reference image alone; private benchmark ground truth is never accessed. LEGO-Plugin comprises three modules, each targeting one failure mode from Section~\ref{sec:failure_diagnosis}. We illustrate the workflow in Figure~\ref{fig:lego_plugin_design} .

\paragraph{\textbf{Enhanced Initialization.}}
LEGO-Plugin recovers the scene frame and camera from the reference image with VGGT~\citep{wang2025vggt}, then derives visible-object layout cues to form an explicit initialization proposal. This gives the agent a reference-aligned starting scene instead of an unconstrained first placement.

\paragraph{\textbf{Grounded Refinement.}}
Since agents judge their own scenes unreliably (Section~\ref{sec:vlm_judge}), Grounded Refinement replaces free-form self-correction with tool-grounded measurement. It extracts reference object regions with SAM~3~\citep{carion2026sam3segmentconcepts} and relative depth with Depth Anything V2~\citep{yang2024depthanythingv2}, then compares the executed scene against these targets through explicit residuals, such as projected extent and relative depth, rather than global visual judgment.

\paragraph{\textbf{Version Control.}}
LEGO-Plugin treats each bounded update as a candidate revision and scores the resulting scene against reference-grounded residuals. Each update is accepted, repaired, or rolled back to the last accepted state, preventing later edits from silently overwriting earlier progress.

\Needspace{22\baselineskip}
\subsection{LEGO-Plugin Evaluation}
\label{sec:plugin_evaluation}

\noindent
\begin{minipage}[t]{0.57\textwidth}
\vspace{0pt}

\paragraph{\textbf{Protocol.}}
We evaluate LEGO-Plugin on the 42-case Office subset (Section~\ref{sec:LEGO-Bench}), comparing each model with and without the plugin under identical inputs, prompts, execution budgets, reasoning effort, and evaluator. As noted in Section~\ref{sec:lego_plugin}, the plugin sees only the reference image and the current Blender scene, never private ground truth or LEGO-Bench scores, so it guides construction without optimizing against the evaluator.
\vspace{4mm}
\paragraph{\textbf{Main comparison.}}
We compare each model with and without LEGO-Plugin on the Office subset, reporting the overall LEGO-Bench score for six model settings (Figure~\ref{fig:lego_plugin_model_comparison}). LEGO-Plugin improves all six models, with gains inversely related to base performance: the three \texttt{GPT-5.6} models improve by $55$--$63\%$ relative (about $7$--$8$ points), \texttt{GPT-6-luna} and \texttt{GPT-6-sol} by $27.5\%$ and $12.1\%$, and \texttt{GPT-6-astra} by only $2.1\%$. This pattern suggests that the plugin mainly compensates for the initialization, regression, and self-evaluation failures that weaker agents exhibit, while the strongest agent already avoids most of them.
\end{minipage}%
\hfill
\begin{minipage}[t]{0.4\textwidth}
\vspace{0pt}
\centering
\includegraphics[width=\linewidth]{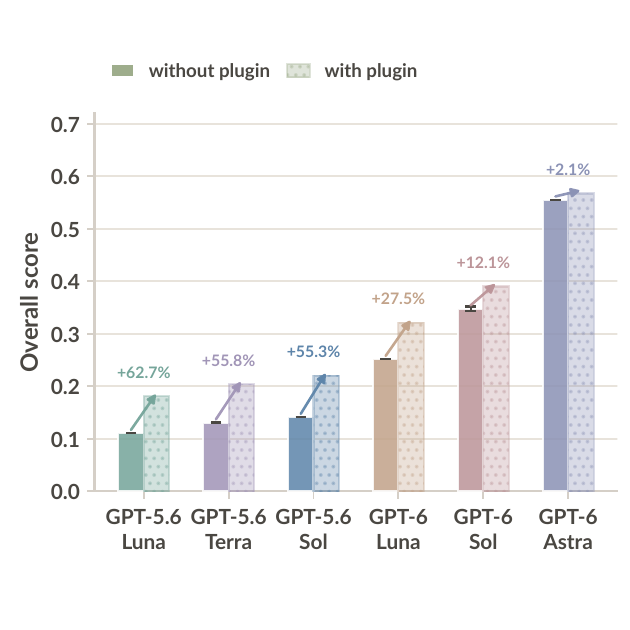}
\captionsetup{skip=2pt}
\captionof{figure}{\textbf{LEGO-Plugin comparison.} LEGO-Plugin improves all six models.}
\label{fig:lego_plugin_model_comparison}
\end{minipage}
\par
\vspace{0.5\baselineskip}

\finding{LEGO-Plugin improves every model without training, with the largest gains for weaker agents, narrowing but not closing the gap to the strongest model.}

\begin{figure}[!t]
\centering
\setlength{\tabcolsep}{0.5pt}
\renewcommand{\arraystretch}{0.9}
\begin{tabular}{@{}ccc:ccc@{}}
\multicolumn{3}{c:}{\scriptsize\textbf{Example 1: Reception}} &
\multicolumn{3}{c}{\scriptsize\textbf{Example 2: Copy room}} \\
{\scriptsize Ref.} & {\scriptsize w/o plugin} & {\scriptsize w/ plugin} &
{\scriptsize Ref.} & {\scriptsize w/o plugin} & {\scriptsize w/ plugin} \\
\includegraphics[width=0.162\linewidth]{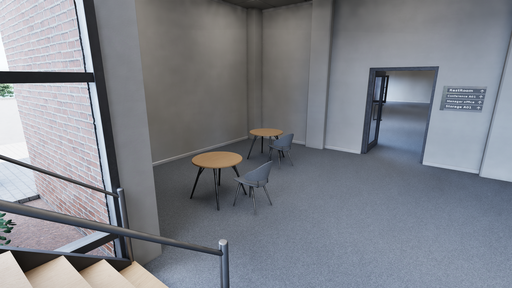} &
\includegraphics[width=0.162\linewidth]{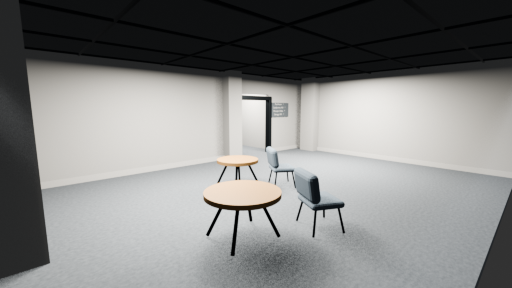} &
\includegraphics[width=0.162\linewidth]{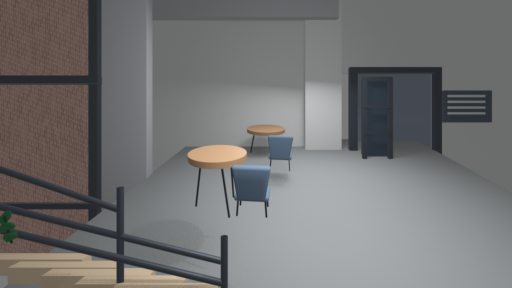} &
\includegraphics[width=0.162\linewidth]{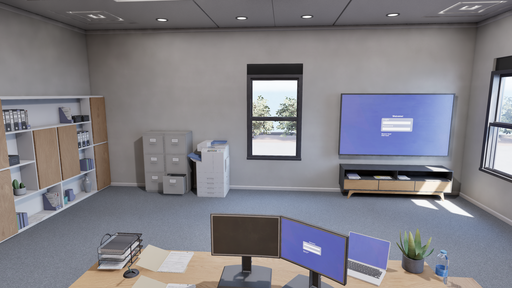} &
\includegraphics[width=0.162\linewidth]{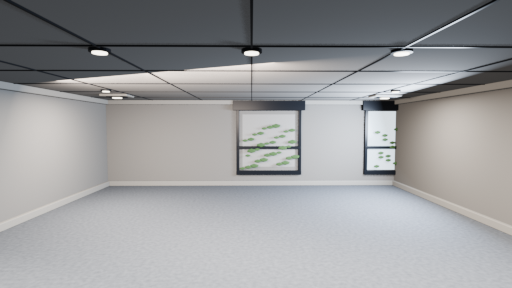} &
\includegraphics[width=0.162\linewidth]{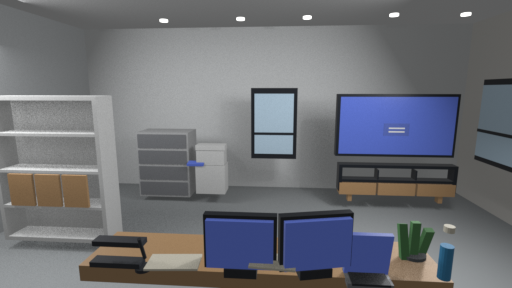}
\end{tabular}
\caption{\textbf{Qualitative LEGO-Plugin examples.} Relative to the best of three \texttt{GPT-5.6-sol} base runs, LEGO-Plugin better matches both references and more than doubles overall scores ($0.147{\rightarrow}0.325$; $0.109{\rightarrow}0.278$).}
\label{fig:lego_plugin_examples}
\vspace{-4mm}
\end{figure}

\Needspace{22\baselineskip}
\section{LEGO-World: Reconstructed Scenes as Visual Representations}
\label{sec:lego-world}

\begin{wraptable}[16]{r}{0.42\textwidth}
    \vspace{-0.5\baselineskip}
        \legotablestyle[3pt]
    \rowcolors{2}{gray!11}{white}
    \captionsetup{skip=3pt}
    \caption{
        \textbf{Task readout from frozen reconstructed scenes
        on three 100-image tracks.} Without task-specific training, the same scenes support detection, segmentation, and depth, but remain substantially behind specialist models.
    }
    \label{tab:universal_tasks}

    \resizebox{\linewidth}{!}{%
    \begin{tabular}{lll|c}
        \legotabletoprule
        \textbf{Task} & \textbf{Metric}
            & \textbf{Methods} & \textbf{Score} \\
        \midrule

        \hiderowcolors
        \multicolumn{4}{c}{%
            \textcolor{gray}{\textit{Detection}}%
        } \\
        \showrowcolors
            & AP $\uparrow$
            & \texttt{DINO} & 59.88 \\
        \multirow{-2}{*}{\shortstack[l]{COCO\\BBox}}
            &
            & \texttt{LEGO-Anything} & 30.14 \\
        \midrule

        \hiderowcolors
        \multicolumn{4}{c}{%
            \textcolor{gray}{\textit{Segmentation}}%
        } \\
        \showrowcolors
            & AP $\uparrow$
            & \texttt{Segment Anything~3} & 53.96 \\
        \multirow{-2}{*}{\shortstack[l]{LVIS\\Mask}}
            &
            & \texttt{LEGO-Anything} & 14.75 \\
        \midrule

        \hiderowcolors
        \multicolumn{4}{c}{%
            \textcolor{gray}{\textit{Depth Estimation}}%
        } \\
        \showrowcolors
            & AbsRel $\downarrow$
            & \texttt{Depth Anything~3} & 0.0783 \\
        \multirow{-2}{*}{\shortstack[l]{ETH3D\\Depth}}
            &
            & \texttt{LEGO-Anything} & 0.1554 \\
        \legotablebottomrule
    \end{tabular}%
    }

    \vspace{-0.5\baselineskip}
\end{wraptable}

We next ask whether reconstructed scenes can serve as representations of natural images. Because each reconstructed artifact is an executable scene program, object detection, instance segmentation, and relative depth can be obtained as deterministic readouts of the same frozen scene rather than as separate task-specific predictions (Figure~\ref{fig:main_graph}). For task $k$, a deterministic readout $\rho_k$ produces the prediction:
\begin{equation}
I \xrightarrow{\;\pi,\,\mathcal{E}\;} P \xrightarrow{\;\mathrm{Exec}_{\mathcal{E}}\;} S \xrightarrow{\;\rho_k\;} \hat{y}_k.
\label{eq:scene_readout}
\end{equation}
We test whether a single reconstructed scene supports three standard vision tasks. We reconstruct scenes with \texttt{GPT-6-astra}, the strongest agent on LEGO-Bench, without LEGO-Plugin, and let the agent label each object from the target dataset's category set. From each frozen scene, visible object instances are projected into 2D boxes for detection and instance masks for segmentation, and camera-space depth renderings provide relative depth. We evaluate on 100 randomly sampled images each from COCO val2017~\citep{lin2015microsoftcococommonobjects}, LVIS v1 val~\citep{gupta2019lvisdatasetlargevocabulary}, and ETH3D~\citep{schops2017multi}. Since scene exports provide no calibrated confidences, box and mask AP follow an equal-confidence protocol (Appendix~\ref{app:downstream_ap}); coverage is reported in Table~\ref{tab:appendix_downstream_coverage}. As shown in Table~\ref{tab:universal_tasks}, scene readouts reach $30.1$ box AP, $14.8$ mask AP, and $0.155$ AbsRel, compared with $59.9$, $54.0$, and $0.078$ for DINO~\citep{siméoni2025dinov3}, SAM~3~\citep{carion2026sam3segmentconcepts}, and Depth Anything~3~\citep{lin2025depth3recoveringvisual}.

\WFclear
\finding{Scenes reconstructed by current coding agents already support detection, segmentation, and depth without task-specific training, reaching about half of specialist box AP, but fall well short of specialized models. Executable scenes from current coding agents are thus a promising but not yet sufficiently precise representation of natural images.}

\section{Conclusion}

We presented \textbf{LEGO-Anything}, an Image-to-Code framework in which coding agents iteratively reconstruct 3D scenes from a single image as editable, executable scene programs. \textbf{LEGO-Bench} separately measures artifact validity, visible-surface geometry, and rendered appearance, revealing that current agents reliably deliver valid scene artifacts but remain limited in geometric fidelity. Trajectory analysis traces this gap to weak initialization, regressive edits, and unreliable self-evaluation, which \textbf{LEGO-Plugin} addresses through Enhanced Initialization, Version Control, and Grounded Refinement, improving all six evaluated models without training. Finally, \textbf{LEGO-World} shows that a single frozen scene already supports detection, segmentation, and depth readouts, though well below specialized models. Together, these results position executable scene programs as a promising, measurable, and diagnosable target for general-purpose coding agents, and we hope LEGO-Bench's extensible design supports future progress toward faithful scene reconstruction.

\section*{Acknowledgment}

We thank Longxuan Yu, Bale Chen, Jiyeon Kim, Janet Wang, Tianyi Zhou, Mustafa Kaba, and Hideo Kobayashi for their invaluable support and discussion.

\clearpage

\bibliographystyle{assets/plainnat}
\bibliography{ref}

@article{zhou2026digital,
  title={Digital Twin AI: Opportunities and Challenges from Large Language Models to World Models},
  author={Zhou, Rong and Chen, Dongping and Jia, Zihan and Su, Yao and Liu, Yixin and Lu, Yiwen and Shi, Dongwei and Huang, Yue and Xu, Tianyang and Pan, Yi and others},
  journal={arXiv preprint arXiv:2601.01321},
  year={2026}
}

@inproceedings{xiang2020sapien,
  title={Sapien: A simulated part-based interactive environment},
  author={Xiang, Fanbo and Qin, Yuzhe and Mo, Kaichun and Xia, Yikuan and Zhu, Hao and Liu, Fangchen and Liu, Minghua and Jiang, Hanxiao and Yuan, Yifu and Wang, He and others},
  booktitle={Proceedings of the IEEE/CVF conference on computer vision and pattern recognition},
  pages={11097--11107},
  year={2020}
}

@misc{kang2026simworldstudioautomaticenvironment,
      title={SimWorld Studio: Automatic Environment Generation with Evolving Coding Agent for Embodied Agent Learning}, 
      author={Haoqiang Kang and Xiaokang Ye and Yuhan Liu and Siddhant Hitesh Mantri and Lingjun Mao and James Fleming and Drishti Regmi and Lianhui Qin},
      year={2026},
      eprint={2605.09423},
      archivePrefix={arXiv},
      primaryClass={cs.AI},
      url={https://arxiv.org/abs/2605.09423}, 
}

@misc{deng2024citycraftrealcrafter3d,
      title={CityCraft: A Real Crafter for 3D City Generation}, 
      author={Jie Deng and Wenhao Chai and Junsheng Huang and Zhonghan Zhao and Qixuan Huang and Mingyan Gao and Jianshu Guo and Shengyu Hao and Wenhao Hu and Jenq-Neng Hwang and Xi Li and Gaoang Wang},
      year={2024},
      eprint={2406.04983},
      archivePrefix={arXiv},
      primaryClass={cs.CV},
      url={https://arxiv.org/abs/2406.04983}, 
}

@misc{raistrick2023infinitephotorealisticworldsusing,
      title={Infinite Photorealistic Worlds using Procedural Generation}, 
      author={Alexander Raistrick and Lahav Lipson and Zeyu Ma and Lingjie Mei and Mingzhe Wang and Yiming Zuo and Karhan Kayan and Hongyu Wen and Beining Han and Yihan Wang and Alejandro Newell and Hei Law and Ankit Goyal and Kaiyu Yang and Jia Deng},
      year={2023},
      eprint={2306.09310},
      archivePrefix={arXiv},
      primaryClass={cs.CV},
      url={https://arxiv.org/abs/2306.09310}, 
}

@misc{kim2026sceneconductor3dscenegeneration,
      title={SceneConductor: 3D Scene Generation from a Single Image with Multi-Agent Orchestration}, 
      author={Jeonghwan Kim and Yushi Lan and Yongwei Chen and Hieu Trung Nguyen and Chuanyu Pan and Xingang Pan},
      year={2026},
      eprint={2606.08402},
      archivePrefix={arXiv},
      primaryClass={cs.CV},
      url={https://arxiv.org/abs/2606.08402}, 
}

@misc{sam3dteam2026sam3d3dfyimages,
      title={SAM 3D: 3Dfy Anything in Images}, 
      author={SAM 3D Team and Xingyu Chen and Fu-Jen Chu and Pierre Gleize and Kevin J Liang and Alexander Sax and Hao Tang and Weiyao Wang and Michelle Guo and Thibaut Hardin and Xiang Li and Aohan Lin and Jiawei Liu and Ziqi Ma and Anushka Sagar and Bowen Song and Xiaodong Wang and Jianing Yang and Bowen Zhang and Piotr Dollár and Georgia Gkioxari and Matt Feiszli and Jitendra Malik},
      year={2026},
      eprint={2511.16624},
      archivePrefix={arXiv},
      primaryClass={cs.CV},
      url={https://arxiv.org/abs/2511.16624}, 
}

@misc{deitke2022procthorlargescaleembodiedai,
      title={ProcTHOR: Large-Scale Embodied AI Using Procedural Generation}, 
      author={Matt Deitke and Eli VanderBilt and Alvaro Herrasti and Luca Weihs and Jordi Salvador and Kiana Ehsani and Winson Han and Eric Kolve and Ali Farhadi and Aniruddha Kembhavi and Roozbeh Mottaghi},
      year={2022},
      eprint={2206.06994},
      archivePrefix={arXiv},
      primaryClass={cs.AI},
      url={https://arxiv.org/abs/2206.06994}, 
}

@misc{raistrick2024infinigenindoorsphotorealisticindoor,
      title={Infinigen Indoors: Photorealistic Indoor Scenes using Procedural Generation}, 
      author={Alexander Raistrick and Lingjie Mei and Karhan Kayan and David Yan and Yiming Zuo and Beining Han and Hongyu Wen and Meenal Parakh and Stamatis Alexandropoulos and Lahav Lipson and Zeyu Ma and Jia Deng},
      year={2024},
      eprint={2406.11824},
      archivePrefix={arXiv},
      primaryClass={cs.CV},
      url={https://arxiv.org/abs/2406.11824}, 
}

@misc{openai_codex_2026,
  author       = {{OpenAI}},
  title        = {Codex: AI Coding Partner from OpenAI},
  year         = {2026},
  howpublished = {\url{https://openai.com/codex/}},
  note         = {Accessed: 2026-06-27}
}

@misc{anthropic_claude_code_2026,
  author       = {{Anthropic}},
  title        = {Claude Code by Anthropic: AI Coding Agent, Terminal, IDE},
  year         = {2026},
  howpublished = {\url{https://claude.com/product/claude-code}},
  note         = {Accessed: 2026-06-27}
}

@misc{zhou2026articraftagenticscalablearticulated,
      title={Articraft: An Agentic System for Scalable Articulated 3D Asset Generation}, 
      author={Matt Zhou and Ruining Li and Xiaoyang Lyu and Zhaomou Song and Zhening Huang and Chuanxia Zheng and Christian Rupprecht and Andrea Vedaldi and Shangzhe Wu},
      year={2026},
      eprint={2605.15187},
      archivePrefix={arXiv},
      primaryClass={cs.CV},
      url={https://arxiv.org/abs/2605.15187}, 
}

@misc{le2025articulateanythingautomaticmodelingarticulated,
      title={Articulate-Anything: Automatic Modeling of Articulated Objects via a Vision-Language Foundation Model}, 
      author={Long Le and Jason Xie and William Liang and Hung-Ju Wang and Yue Yang and Yecheng Jason Ma and Kyle Vedder and Arjun Krishna and Dinesh Jayaraman and Eric Eaton},
      year={2025},
      eprint={2410.13882},
      archivePrefix={arXiv},
      primaryClass={cs.CV},
      url={https://arxiv.org/abs/2410.13882}, 
}

@misc{cao2025physxanythingsimulationreadyphysical3d,
      title={PhysX-Anything: Simulation-Ready Physical 3D Assets from Single Image}, 
      author={Ziang Cao and Fangzhou Hong and Zhaoxi Chen and Liang Pan and Ziwei Liu},
      year={2025},
      eprint={2511.13648},
      archivePrefix={arXiv},
      primaryClass={cs.CV},
      url={https://arxiv.org/abs/2511.13648}, 
}

@misc{liu2023parispartlevelreconstructionmotion,
      title={PARIS: Part-level Reconstruction and Motion Analysis for Articulated Objects}, 
      author={Jiayi Liu and Ali Mahdavi-Amiri and Manolis Savva},
      year={2023},
      eprint={2308.07391},
      archivePrefix={arXiv},
      primaryClass={cs.CV},
      url={https://arxiv.org/abs/2308.07391}, 
}

@misc{jiang2022dittobuildingdigitaltwins,
      title={Ditto: Building Digital Twins of Articulated Objects from Interaction}, 
      author={Zhenyu Jiang and Cheng-Chun Hsu and Yuke Zhu},
      year={2022},
      eprint={2202.08227},
      archivePrefix={arXiv},
      primaryClass={cs.CV},
      url={https://arxiv.org/abs/2202.08227}, 
}

@misc{ning2026codeagentharness,
      title={Code as Agent Harness}, 
      author={Xuying Ning and Katherine Tieu and Dongqi Fu and Tianxin Wei and Zihao Li and Yuanchen Bei and Jiaru Zou and Mengting Ai and Zhining Liu and Ting-Wei Li and Lingjie Chen and Yanjun Zhao and Ke Yang and Bingxuan Li and Cheng Qian and Gaotang Li and Xiao Lin and Zhichen Zeng and Ruizhong Qiu and Sirui Chen and Yifan Sun and Xiyuan Yang and Ruida Wang and Rui Pan and Chenyuan Yang and Dylan Zhang and Liri Fang and Zikun Cui and Yang Cao and Pan Chen and Dorothy Sun and Ren Chen and Mahesh Srinivasan and Nipun Mathur and Yinglong Xia and Hong Li and Hong Yan and Pan Lu and Lingming Zhang and Tong Zhang and Hanghang Tong and Jingrui He},
      year={2026},
      eprint={2605.18747},
      archivePrefix={arXiv},
      primaryClass={cs.CL},
      url={https://arxiv.org/abs/2605.18747}, 
}

@misc{jimenez2024swebenchlanguagemodelsresolve,
      title={SWE-bench: Can Language Models Resolve Real-World GitHub Issues?}, 
      author={Carlos E. Jimenez and John Yang and Alexander Wettig and Shunyu Yao and Kexin Pei and Ofir Press and Karthik Narasimhan},
      year={2024},
      eprint={2310.06770},
      archivePrefix={arXiv},
      primaryClass={cs.CL},
      url={https://arxiv.org/abs/2310.06770}, 
}

@software{openclaw2026,
  author       = {{OpenClaw contributors}},
  title        = {{OpenClaw}: Personal AI Assistant},
  year         = {2026},
  version      = {2026.6.10},
  license      = {MIT},
  url          = {https://github.com/openclaw/openclaw},
  urldate      = {2026-06-30},
  note         = {GitHub repository}
}

@misc{liu2026diveclaudecodedesign,
      title={Dive into Claude Code: The Design Space of Today's and Future AI Agent Systems}, 
      author={Jiacheng Liu and Xiaohan Zhao and Xinyi Shang and Zhiqiang Shen},
      year={2026},
      eprint={2604.14228},
      archivePrefix={arXiv},
      primaryClass={cs.SE},
      url={https://arxiv.org/abs/2604.14228}, 
}

@misc{lee2026metaharnessendtoendoptimizationmodel,
      title={Meta-Harness: End-to-End Optimization of Model Harnesses}, 
      author={Yoonho Lee and Roshen Nair and Qizheng Zhang and Kangwook Lee and Omar Khattab and Chelsea Finn},
      year={2026},
      eprint={2603.28052},
      archivePrefix={arXiv},
      primaryClass={cs.AI},
      url={https://arxiv.org/abs/2603.28052}, 
}

@software{karpathy_autoresearch_2026,
  author       = {Karpathy, Andrej},
  title        = {autoresearch: Autonomous pretraining research swarm},
  year         = {2026},
  version      = {0.1.0},
  url          = {https://github.com/karpathy/autoresearch},
  urldate      = {2026-06-30},
  note         = {GitHub repository}
}

@misc{li2026clawenvkitautomaticenvironmentgeneration,
      title={ClawEnvKit: Automatic Environment Generation for Claw-Like Agents}, 
      author={Xirui Li and Ming Li and Ion Stoica and Cho-Jui Hsieh and Tianyi Zhou},
      year={2026},
      eprint={2604.18543},
      archivePrefix={arXiv},
      primaryClass={cs.AI},
      url={https://arxiv.org/abs/2604.18543}, 
}

@misc{fu2026capxframeworkbenchmarkingimproving,
      title={CaP-X: A Framework for Benchmarking and Improving Coding Agents for Robot Manipulation}, 
      author={Max Fu and Justin Yu and Karim El-Refai and Ethan Kou and Haoru Xue and Huang Huang and Wenli Xiao and Guanzhi Wang and Fei-Fei Li and Guanya Shi and Jiajun Wu and Shankar Sastry and Yuke Zhu and Ken Goldberg and Linxi "Jim" Fan},
      year={2026},
      eprint={2603.22435},
      archivePrefix={arXiv},
      primaryClass={cs.RO},
      url={https://arxiv.org/abs/2603.22435}, 
}

@misc{liu2026guavaeffectiveuniversalharness,
      title={Guava: An Effective and Universal Harness for Embodied Manipulation}, 
      author={Haowen Liu and Xirui Li and Shaoxiong Yao and Peng Shi and Tianyi Zhou and Jia-Bin Huang and Furong Huang and Jiayuan Mao},
      year={2026},
      eprint={2606.18363},
      archivePrefix={arXiv},
      primaryClass={cs.RO},
      url={https://arxiv.org/abs/2606.18363}, 
}

@misc{yu2026agentichardwaredesignrepositorylevel,
      title={Agentic Hardware Design as Repository-Level Code Evolution}, 
      author={Cunxi Yu and Chenhui Deng and Nathaniel Pinckney and Brucek Khailany},
      year={2026},
      eprint={2606.28279},
      archivePrefix={arXiv},
      primaryClass={cs.AR},
      url={https://arxiv.org/abs/2606.28279}, 
}

@misc{yang2026p3dbenchbenchmarkingmllmsparametric,
      title={P3D-Bench: Benchmarking MLLMs for Parametric 3D Generation and Structural Reasoning}, 
      author={Yikang Yang and Zhanpeng Hu and Youtian Lin and Mengqi Zhou and Jingxi Xu and Feihu Zhang and Jiaheng Liu and Yao Yao},
      year={2026},
      eprint={2606.11152},
      archivePrefix={arXiv},
      primaryClass={cs.CV},
      url={https://arxiv.org/abs/2606.11152}, 
}

@misc{chen2026sandboxedcodingagentscompetitive,
      title={Sandboxed Coding Agents are Competitive Omni-modal Task Solvers}, 
      author={Dongping Chen and Xuanao Huang and Zhihan Hu and Qingyuan Shi and Dianqi Li and Tianyi Zhou},
      year={2026},
      eprint={2606.00579},
      archivePrefix={arXiv},
      primaryClass={cs.CL},
      url={https://arxiv.org/abs/2606.00579}, 
}

@misc{cao2026collaborativemultimodalcodinghighquality,
      title={Collaborative Multi-Modal Coding for High-Quality 3D Generation}, 
      author={Ziang Cao and Zhaoxi Chen and Liang Pan and Ziwei Liu},
      year={2026},
      eprint={2508.15228},
      archivePrefix={arXiv},
      primaryClass={cs.CV},
      url={https://arxiv.org/abs/2508.15228}, 
}

@misc{gao20263dcodebenchbenchmarkingagenticprocedural,
      title={3DCodeBench: Benchmarking Agentic Procedural 3D Modeling Via Code}, 
      author={Yipeng Gao and Lei Shu and Genzhi Ye and Xi Xiong and Ameesh Makadia and Meiqi Guo and Laurent Itti and Jindong Chen},
      year={2026},
      eprint={2606.01057},
      archivePrefix={arXiv},
      primaryClass={cs.CV},
      url={https://arxiv.org/abs/2606.01057}, 
}

@article{lu2026worldcoder,
  title={WorldCoder-Bench: Benchmarking Physically Grounded 3D World Synthesis},
  author={Lu, Shuo and Xu, Yinuo and Yu, Kecheng and Jiang, Siru and Yu, Yongcan and Wang, Yubin and Yang, Haitao and Zhang, Yuxiang and Wang, Bin and He, Ran and others},
  journal={arXiv preprint arXiv:2606.01869},
  year={2026}
}

@misc{xia2025scenegenagentpreciseindustrialscene,
      title={SceneGenAgent: Precise Industrial Scene Generation with Coding Agent}, 
      author={Xiao Xia and Dan Zhang and Zibo Liao and Zhenyu Hou and Tianrui Sun and Jing Li and Ling Fu and Yuxiao Dong},
      year={2025},
      eprint={2410.21909},
      archivePrefix={arXiv},
      primaryClass={cs.CL},
      url={https://arxiv.org/abs/2410.21909}, 
}

@misc{huang2026literealitygraphicsready3dscene,
      title={LiteReality: Graphics-Ready 3D Scene Reconstruction from RGB-D Scans}, 
      author={Zhening Huang and Xiaoyang Wu and Fangcheng Zhong and Hengshuang Zhao and Matthias Nießner and Joan Lasenby},
      year={2026},
      eprint={2507.02861},
      archivePrefix={arXiv},
      primaryClass={cs.CV},
      url={https://arxiv.org/abs/2507.02861}, 
}

@misc{zhang2026simartdecomposingmonolithicmeshes,
      title={SIMART: Decomposing Monolithic Meshes into Sim-ready Articulated Assets via MLLM}, 
      author={Chuanrui Zhang and Minghan Qin and Yuang Wang and Baifeng Xie and Hang Li and Ziwei Wang},
      year={2026},
      eprint={2603.23386},
      archivePrefix={arXiv},
      primaryClass={cs.CV},
      url={https://arxiv.org/abs/2603.23386}, 
}

@misc{wu2026urdfanythingendtoendgenerationsimulationready,
      title={URDF-Anything+: End-to-End Generation for Simulation-Ready Articulated Assets}, 
      author={Zhuangzhe Wu and Yue Xin and Chengkai Hou and Minghao Chen and Yaoxu Lyu and Jieyu Zhang and Shanghang Zhang},
      year={2026},
      eprint={2603.14010},
      archivePrefix={arXiv},
      primaryClass={cs.RO},
      url={https://arxiv.org/abs/2603.14010}, 
}

@misc{carion2026sam3segmentconcepts,
      title={SAM 3: Segment Anything with Concepts}, 
      author={Nicolas Carion and Laura Gustafson and Yuan-Ting Hu and Shoubhik Debnath and Ronghang Hu and Didac Suris and Chaitanya Ryali and Kalyan Vasudev Alwala and Haitham Khedr and Andrew Huang and Jie Lei and Tengyu Ma and Baishan Guo and Arpit Kalla and Markus Marks and Joseph Greer and Meng Wang and Peize Sun and Roman Rädle and Triantafyllos Afouras and Effrosyni Mavroudi and Katherine Xu and Tsung-Han Wu and Yu Zhou and Liliane Momeni and Rishi Hazra and Shuangrui Ding and Sagar Vaze and Francois Porcher and Feng Li and Siyuan Li and Aishwarya Kamath and Ho Kei Cheng and Piotr Dollár and Nikhila Ravi and Kate Saenko and Pengchuan Zhang and Christoph Feichtenhofer},
      year={2026},
      eprint={2511.16719},
      archivePrefix={arXiv},
      primaryClass={cs.CV},
      url={https://arxiv.org/abs/2511.16719}, 
}

@misc{wang2025vggt,
      title={VGGT: Visual Geometry Grounded Transformer},
      author={Jianyuan Wang and Minghao Chen and Nikita Karaev and Andrea Vedaldi and Christian Rupprecht and David Novotny},
      year={2025},
      eprint={2503.11651},
      archivePrefix={arXiv},
      primaryClass={cs.CV},
      url={https://arxiv.org/abs/2503.11651},
}

@misc{yang2024depthanythingv2,
      title={Depth Anything V2},
      author={Lihe Yang and Bingyi Kang and Zilong Huang and Zhen Zhao and Xiaogang Xu and Jiashi Feng and Hengshuang Zhao},
      year={2024},
      eprint={2406.09414},
      archivePrefix={arXiv},
      primaryClass={cs.CV},
      url={https://arxiv.org/abs/2406.09414},
}

@software{blender_mcp,
  author  = {{ahujasid}},
  title   = {BlenderMCP: Blender Model Context Protocol Integration},
  url     = {https://github.com/ahujasid/blender-mcp},
  version = {1.8.0},
  note    = {GitHub repository, accessed 2026-08-06},
  year    = {2025}
}

@software{Harbor_Framework,
author = {{Harbor Framework Team}},
title = {{Harbor: A framework for evaluating and optimizing agents and models in container environments}},
year = {2026},
version = {v0.16.1},
doi = {10.5281/zenodo.20953922},
url = {https://doi.org/10.5281/zenodo.20953922}
}

@misc{nousHermesAgent,
  author       = {{Nous Research}},
  title        = {Hermes Agent},
  year         = {2026},
  howpublished = {\url{https://github.com/NousResearch/hermes-agent}},
  note         = {Pinned revision aa6f77596b, accessed 2026-08-20}
}

@misc{codexagentloop,
  author       = {Michael Bolin},
  title        = {Unrolling the Codex Agent Loop},
  year         = {2026},
  month        = jan,
  day          = {23},
  publisher    = {OpenAI},
  howpublished = {\url{https://openai.com/index/unrolling-the-codex-agent-loop/}},
  note         = {Accessed: 2026-08-06}
}

@software{blender501,
  author  = {{Blender Foundation}},
  title   = {Blender},
  version = {5.0.1},
  date    = {2025-12-16},
  url     = {https://www.blender.org/}
}

@misc{ma2026lychsimcontrollableinteractivesimulation,
      title={LychSim: A Controllable and Interactive Simulation Framework for Vision Research}, 
      author={Wufei Ma and Chloe Wang and Siyi Chen and Jiawei Peng and Patrick Li and Alan Yuille},
      year={2026},
      eprint={2605.12449},
      archivePrefix={arXiv},
      primaryClass={cs.CV},
      url={https://arxiv.org/abs/2605.12449}, 
}

@misc{ma2025p3samnative3dsegmentation,
      title={P3-SAM: Native 3D Part Segmentation}, 
      author={Changfeng Ma and Yang Li and Xinhao Yan and Jiachen Xu and Yunhan Yang and Chunshi Wang and Zibo Zhao and Yanwen Guo and Zhuo Chen and Chunchao Guo},
      year={2025},
      eprint={2509.06784},
      archivePrefix={arXiv},
      primaryClass={cs.CV},
      url={https://arxiv.org/abs/2509.06784}, 
}

@misc{yan2025xparthighfidelitystructure,
      title={X-Part: high fidelity and structure coherent shape decomposition}, 
      author={Xinhao Yan and Jiachen Xu and Yang Li and Changfeng Ma and Yunhan Yang and Chunshi Wang and Zibo Zhao and Zeqiang Lai and Yunfei Zhao and Zhuo Chen and Chunchao Guo},
      year={2025},
      eprint={2509.08643},
      archivePrefix={arXiv},
      primaryClass={cs.GR},
      url={https://arxiv.org/abs/2509.08643}, 
}

@misc{zhao2026sceneactbenchagentsact3d,
      title={SceneActBench: Can Agents Act on the 3D Scenes They See?},
      author={Yifei Zhao and Xiangxin Zhou and Wenhao Yang and Jiaqi Tang and Pu Jian and Huanjin Yao and Jiarui Yao and Haowei Lin and Chunchao Guo and Zhuo Chen and Wenkai Lyu and Jianzhu Ma and Xueqian Wang and Wenxi Zhu},
      year={2026},
      eprint={2607.22393},
      archivePrefix={arXiv},
      primaryClass={cs.AI},
      url={https://arxiv.org/abs/2607.22393},
}

@misc{lu2026worldcoderbenchbenchmarkingphysicallygrounded,
      title={WorldCoder-Bench: Benchmarking Physically Grounded 3D World Synthesis}, 
      author={Shuo Lu and Yinuo Xu and Kecheng Yu and Siru Jiang and Yongcan Yu and Yubin Wang and Haitao Yang and Yuxiang Zhang and Bin Wang and Ran He and Jian Liang},
      year={2026},
      eprint={2606.01869},
      archivePrefix={arXiv},
      primaryClass={cs.AI},
      url={https://arxiv.org/abs/2606.01869}, 
}

@misc{yin2026visionasinversegraphicsagentinterleavedmultimodal,
      title={Vision-as-Inverse-Graphics Agent via Interleaved Multimodal Reasoning}, 
      author={Shaofeng Yin and Jiaxin Ge and Zora Zhiruo Wang and Chenyang Wang and Xiuyu Li and Michael J. Black and Trevor Darrell and Angjoo Kanazawa and Haiwen Feng},
      year={2026},
      eprint={2601.11109},
      archivePrefix={arXiv},
      primaryClass={cs.CV},
      url={https://arxiv.org/abs/2601.11109}, 
}

@misc{liu20263dprimitivesspatiallanguage,
      title={3D Primitives are a Spatial Language for VLMs}, 
      author={Junze Liu and Kun Qian and Florian Dubost and Kai Zhong and Arvind Srinivasan and Nan Chen and Anping Wang and Sam Zhang and Alejandro Mottini and Qingjun Cui and Tian Wang},
      year={2026},
      eprint={2605.12586},
      archivePrefix={arXiv},
      primaryClass={cs.CV},
      url={https://arxiv.org/abs/2605.12586}, 
}

@misc{sautter20253dregen3dreconstructionindoor,
      title={3D-RE-GEN: 3D Reconstruction of Indoor Scenes with a Generative Framework}, 
      author={Tobias Sautter and Jan-Niklas Dihlmann and Hendrik P. A. Lensch},
      year={2025},
      eprint={2512.17459},
      archivePrefix={arXiv},
      primaryClass={cs.CV},
      url={https://arxiv.org/abs/2512.17459}, 
}

@misc{ardelean2025gen3dsrgeneralizable3dscene,
      title={Gen3DSR: Generalizable 3D Scene Reconstruction via Divide and Conquer from a Single View}, 
      author={Andreea Ardelean and Mert Özer and Bernhard Egger},
      year={2025},
      eprint={2404.03421},
      archivePrefix={arXiv},
      primaryClass={cs.CV},
      url={https://arxiv.org/abs/2404.03421}, 
}

@misc{ma2026rest3dreconstructingphysicallystable,
      title={REST3D: Reconstructing Physically Stable 3D Scenes from a Single Image}, 
      author={Xiaoxuan Ma and Jiashun Wang and Nicolas Ugrinovic and Yehonathan Litman and Kris Kitani},
      year={2026},
      eprint={2605.30338},
      archivePrefix={arXiv},
      primaryClass={cs.CV},
      url={https://arxiv.org/abs/2605.30338}, 
}

@misc{siméoni2025dinov3,
      title={DINOv3}, 
      author={Oriane Siméoni and Huy V. Vo and Maximilian Seitzer and Federico Baldassarre and Maxime Oquab and Cijo Jose and Vasil Khalidov and Marc Szafraniec and Seungeun Yi and Michaël Ramamonjisoa and Francisco Massa and Daniel Haziza and Luca Wehrstedt and Jianyuan Wang and Timothée Darcet and Théo Moutakanni and Leonel Sentana and Claire Roberts and Andrea Vedaldi and Jamie Tolan and John Brandt and Camille Couprie and Julien Mairal and Hervé Jégou and Patrick Labatut and Piotr Bojanowski},
      year={2025},
      eprint={2508.10104},
      archivePrefix={arXiv},
      primaryClass={cs.CV},
      url={https://arxiv.org/abs/2508.10104}, 
}

@misc{lin2025depth3recoveringvisual,
      title={Depth Anything 3: Recovering the Visual Space from Any Views}, 
      author={Haotong Lin and Sili Chen and Junhao Liew and Donny Y. Chen and Zhenyu Li and Guang Shi and Jiashi Feng and Bingyi Kang},
      year={2025},
      eprint={2511.10647},
      archivePrefix={arXiv},
      primaryClass={cs.CV},
      url={https://arxiv.org/abs/2511.10647}, 
}

@misc{lin2015microsoftcococommonobjects,
      title={Microsoft COCO: Common Objects in Context}, 
      author={Tsung-Yi Lin and Michael Maire and Serge Belongie and Lubomir Bourdev and Ross Girshick and James Hays and Pietro Perona and Deva Ramanan and C. Lawrence Zitnick and Piotr Dollár},
      year={2015},
      eprint={1405.0312},
      archivePrefix={arXiv},
      primaryClass={cs.CV},
      url={https://arxiv.org/abs/1405.0312}, 
}

@misc{gupta2019lvisdatasetlargevocabulary,
      title={LVIS: A Dataset for Large Vocabulary Instance Segmentation}, 
      author={Agrim Gupta and Piotr Dollár and Ross Girshick},
      year={2019},
      eprint={1908.03195},
      archivePrefix={arXiv},
      primaryClass={cs.CV},
      url={https://arxiv.org/abs/1908.03195}, 
}

@inproceedings{schops2017multi,
  title={A multi-view stereo benchmark with high-resolution images and multi-camera videos},
  author={Schops, Thomas and Schonberger, Johannes L and Galliani, Silvano and Sattler, Torsten and Schindler, Konrad and Pollefeys, Marc and Geiger, Andreas},
  booktitle={Proceedings of the IEEE conference on computer vision and pattern recognition},
  pages={3260--3269},
  year={2017}
}

@misc{ye2026clawevaltrustworthyevaluationautonomous,
      title={Claw-Eval: Towards Trustworthy Evaluation of Autonomous Agents},
      author={Bowen Ye and Rang Li and Qibin Yang and Yuanxin Liu and Linli Yao and Hanglong Lv and Zhihui Xie and Chenxin An and Lei Li and Lingpeng Kong and Qi Liu and Zhifang Sui and Tong Yang},
      year={2026},
      eprint={2604.06132},
      archivePrefix={arXiv},
      primaryClass={cs.AI},
      url={https://arxiv.org/abs/2604.06132},
}

@misc{li2026scenixsparseview3dscene,
      title={Scenix: Sparse-View 3D Scene Reconstruction via Executable Scene Programs}, 
      author={Kai Li and Lutao Jiang and Zhenyang Li and Jiayu Dong and Jierui Zhang and Yingda Yin and Runze Zhang and Kai Yan and Xiaoyang Huang and Keyang Luo and Xin Wang and Xiangyu Zhao and Weikai Chen},
      year={2026},
      eprint={2608.07012},
      archivePrefix={arXiv},
      primaryClass={cs.CV},
      url={https://arxiv.org/abs/2608.07012}, 
}

@misc{wang2026codeworldsagenticdiscovery,
      title={Code as Worlds: Agentic Discovery of Executable World Representations for Physical Reasoning}, 
      author={Hanyang Wang and Yimo Cai and Weiliang Chen and Jiawei Chi and Haowen Sun and Qiyu Dai and Yi-Hsin Hung and Xingzhuo Guo and Jinshan Ren and Runmao Yao and Ziwei Liu and Mingsheng Long and Yueqi Duan and Jun Gao and Jiangran Lyu and Fangfu Liu and Jialong Wu},
      year={2026},
      eprint={2608.27549},
      archivePrefix={arXiv},
      primaryClass={cs.CV},
      url={https://arxiv.org/abs/2608.27549}, 
}

@misc{he2026thinkingblenderstagedexecutable,
      title={Thinking in Blender: Staged Executable Inverse Graphics with Vision-Language Models}, 
      author={Guangzhao He and Rundong Luo and Wei-Chiu Ma and Hadar Averbuch-Elor},
      year={2026},
      eprint={2606.02580},
      archivePrefix={arXiv},
      primaryClass={cs.CV},
      url={https://arxiv.org/abs/2606.02580}, 
}

@misc{wu2025ll3mlargelanguage3d,
      title={{LL3M}: Large Language 3D Modelers},
      author={Sining Lu and Guan Chen and Nam Anh Dinh and Itai Lang and Ari Holtzman and Rana Hanocka},
      year={2025},
      eprint={2508.08228},
      archivePrefix={arXiv},
      primaryClass={cs.GR},
      url={https://arxiv.org/abs/2508.08228},
}

@misc{chen2025blenderfusion3dgrounded,
      title={BlenderFusion: 3D-Grounded Visual Editing and Generative Compositing},
      author={Jiacheng Chen and Ramin Mehran and Xuhui Jia and Saining Xie and Sanghyun Woo},
      year={2025},
      eprint={2506.17450},
      archivePrefix={arXiv},
      primaryClass={cs.CV},
      url={https://arxiv.org/abs/2506.17450},
}

@misc{zadaianchuk2026reconstructiongeneration,
      title={Reconstruction by Generation: 3D Multi-Object Scene Reconstruction from Sparse Observations},
      author={Andrii Zadaianchuk and Leonardo Barcellona and Lennard Schuenemann and Christian Gumbsch and Zehao Wang and Muhammad Zubair Irshad and Fabien Despinoy and Rahaf Aljundi and Stratis Gavves and Sergey Zakharov},
      year={2026},
      eprint={2604.27106},
      archivePrefix={arXiv},
      primaryClass={cs.CV},
      url={https://arxiv.org/abs/2604.27106},
}

@misc{dahnert2021panoptic3dscene,
      title={Panoptic 3D Scene Reconstruction From a Single RGB Image},
      author={Manuel Dahnert and Ji Hou and Matthias Nie{\ss}ner and Angela Dai},
      year={2021},
      eprint={2111.02444},
      archivePrefix={arXiv},
      primaryClass={cs.CV},
      url={https://arxiv.org/abs/2111.02444},
}

@inproceedings{boittiaux2025buildee3dsimulation,
      title={Buildee: A 3D Simulation Framework for Scene Exploration and Reconstruction with Understanding},
      author={Cl{\'e}mentin Boittiaux and Vincent Lepetit},
      booktitle={NeurIPS 2025 Workshop on Synthetic Data for Computer Vision},
      year={2025},
      url={https://openreview.net/forum?id=1LmsiOaMTy},
}

@misc{zhu2022scenegraphgenerationcomprehensive,
      title={Scene Graph Generation: A Comprehensive Survey}, 
      author={Guangming Zhu and Liang Zhang and Youliang Jiang and Yixuan Dang and Haoran Hou and Peiyi Shen and Mingtao Feng and Xia Zhao and Qiguang Miao and Syed Afaq Ali Shah and Mohammed Bennamoun},
      year={2022},
      eprint={2201.00443},
      archivePrefix={arXiv},
      primaryClass={cs.CV},
      url={https://arxiv.org/abs/2201.00443}, 
}

@misc{wang2024dust3rgeometric3dvision,
      title={DUSt3R: Geometric 3D Vision Made Easy}, 
      author={Shuzhe Wang and Vincent Leroy and Yohann Cabon and Boris Chidlovskii and Jerome Revaud},
      year={2024},
      eprint={2312.14132},
      archivePrefix={arXiv},
      primaryClass={cs.CV},
      url={https://arxiv.org/abs/2312.14132}, 
}

@misc{gkioxari2020meshrcnn,
      title={Mesh R-CNN}, 
      author={Georgia Gkioxari and Jitendra Malik and Justin Johnson},
      year={2020},
      eprint={1906.02739},
      archivePrefix={arXiv},
      primaryClass={cs.CV},
      url={https://arxiv.org/abs/1906.02739}, 
}

@misc{hu2024scenecraftllmagentsynthesizing,
      title={SceneCraft: An LLM Agent for Synthesizing 3D Scene as Blender Code}, 
      author={Ziniu Hu and Ahmet Iscen and Aashi Jain and Thomas Kipf and Yisong Yue and David A. Ross and Cordelia Schmid and Alireza Fathi},
      year={2024},
      eprint={2403.01248},
      archivePrefix={arXiv},
      primaryClass={cs.CV},
      url={https://arxiv.org/abs/2403.01248}, 
}

@inproceedings{zhang2025scene,
  title={The scene language: Representing scenes with programs, words, and embeddings},
  author={Zhang, Yunzhi and Li, Zizhang and Zhou, Matt and Wu, Shangzhe and Wu, Jiajun},
  booktitle={2025 IEEE/CVF Conference on Computer Vision and Pattern Recognition (CVPR)},
  pages={24625--24634},
  year={2025},
  organization={IEEE}
}

\clearpage
\tableofcontents
\clearpage

\clearpage
\beginappendix
\section{Additional Related Work}
\label{app:related_work}

The main paper focuses on work most directly related to end-to-end
image-conditioned 3D scene reconstruction. Here, we briefly situate
LEGO-Anything within adjacent lines on procedural scene generation,
object and asset reconstruction, coding agents, and executable 3D
benchmarks.

\subsection{Procedural Scene Generation and Simulation Environments}
\label{app:rw_procedural_scenes}

Procedural systems construct large scene collections from rules and reusable
asset libraries. ProcTHOR~\citep{deitke2022procthorlargescaleembodiedai}
targets embodied indoor environments; Infinigen and Infinigen
Indoors~\citep{raistrick2023infinitephotorealisticworldsusing,
raistrick2024infinigenindoorsphotorealisticindoor} generate photorealistic
natural and indoor scenes; and CityCraft~\citep{deng2024citycraftrealcrafter3d}
extends procedural construction to cities. These systems provide scalable
generation once structured specifications are available. LEGO-Anything instead
studies whether an agent can infer such an executable specification from a
single reference image.

\subsection{Object Reconstruction and Asset Generation}
\label{app:rw_asset_generation}

Object-level image-to-3D methods recover geometry and appearance from
visual observations. Representative systems include
TriMM~\citep{cao2026collaborativemultimodalcodinghighquality},
SAM 3D~\citep{sam3dteam2026sam3d3dfyimages}, and
LiteReality~\citep{huang2026literealitygraphicsready3dscene}.
Such models provide increasingly capable geometric and appearance
priors, making them useful components of larger scene-construction
pipelines. However, reconstructing an isolated object does not by itself
determine how multiple objects should be selected, positioned,
articulated, and integrated into a coherent executable scene.

Related work also studies articulated and simulation-compatible assets.
Ditto~\citep{jiang2022dittobuildingdigitaltwins} and
PARIS~\citep{liu2023parispartlevelreconstructionmotion} recover movable
parts and articulation from observations. More recent systems generate
or refine structured articulated assets, including
URDF-Anything+~\citep{wu2026urdfanythingendtoendgenerationsimulationready},
SIMART~\citep{zhang2026simartdecomposingmonolithicmeshes}, and
Articulate-Anything~\citep{le2025articulateanythingautomaticmodelingarticulated}.
PhysX-Anything~\citep{cao2025physxanythingsimulationreadyphysical3d}
further targets physical attributes and simulation-ready representations. ArtiCraft~\citep{zhou2026articraftagenticscalablearticulated} develops an
agentic coding interface for articulated-object generation, illustrating
how executable programs can expose structured modeling operations to a
general-purpose coding agent.
These lines are complementary to our setting: they strengthen object-level
geometry, appearance, or functionality, whereas LEGO-Anything evaluates
image-conditioned assembly of full executable scenes.

\subsection{Scene Reconstruction and Editable 3D Intermediates}
\label{app:rw_scene_reconstruction}

Holistic scene-reconstruction methods recover multiple objects, geometry,
semantics, or layout from partial visual evidence. Panoptic 3D scene
reconstruction from a single RGB image jointly reasons about geometry,
semantics, and instances in a fixed scene
representation~\citep{dahnert2021panoptic3dscene}. RecGen reconstructs multi-object scenes
from sparse RGB-D observations through a generative framework, directly
addressing ambiguity, occlusion, and uncertainty under incomplete
views~\citep{zadaianchuk2026reconstructiongeneration}. These works are closely
related in task motivation, but they differ from LEGO-Anything in output
interface: we ask whether a general coding agent can produce an executable
Blender scene program from a single RGB image, rather than predicting a fixed
panoptic or generative scene representation.

Editable 3D intermediates are also useful beyond reconstruction. BlenderFusion
uses 3D-grounded Blender scenes for visual editing and generative compositing,
showing that explicit scene state, camera control, and object manipulation can
support image-conditioned visual tasks~\citep{chen2025blenderfusion3dgrounded}.
This is adjacent to our LEGO-World motivation: executable scenes are valuable
not only as reconstruction artifacts, but also as inspectable representations
that can be queried, edited, and reused by downstream procedures.

\subsection{Coding Agents and Harnesses}
\label{app:rw_agent_harnesses}

Coding-agent systems increasingly use code as the operational interface
for planning, tool use, environment interaction, and execution-based
verification
\citep{jimenez2024swebenchlanguagemodelsresolve,
anthropic_claude_code_2026,
openai_codex_2026,
openclaw2026,
liu2026diveclaudecodedesign}.
Agent harnesses organize these interactions by exposing tools, maintaining
execution state, and returning environmental feedback to the underlying
model~\citep{ning2026codeagentharness}. Recent work has studied automatic
harness optimization~\citep{lee2026metaharnessendtoendoptimizationmodel}
and extended coding agents to scientific, hardware, simulated, and
physical environments
\citep{karpathy_autoresearch_2026,
li2026clawenvkitautomaticenvironmentgeneration,
yu2026agentichardwaredesignrepositorylevel,
chen2026sandboxedcodingagentscompetitive,
fu2026capxframeworkbenchmarkingimproving,
liu2026guavaeffectiveuniversalharness,
kang2026simworldstudioautomaticenvironment}.
LL3M is especially relevant to our formulation because it uses multiple
language-model agents to write, debug, and refine Blender Python scripts for
3D modeling~\citep{wu2025ll3mlargelanguage3d}. LEGO-Anything differs in the
task and evaluation target: it evaluates end-to-end image-conditioned scene
reconstruction under hidden scene ground truth, rather than focusing on
language-driven asset generation and editing.

\subsection{3D Construction Benchmarks}
\label{app:rw_3d_benchmarks}

Existing benchmarks study the generation of executable 3D content through code. 3DCodeBench~\citep{gao20263dcodebenchbenchmarkingagenticprocedural} evaluates agentic procedural modeling, while
WorldCoder-Bench~\citep{lu2026worldcoder} and P3D-Bench~\citep{yang2026p3dbenchbenchmarkingmllmsparametric} examine
code-based generation of structured, parametric, or physically grounded
3D content. These benchmarks establish code generation as a practical
interface between foundation models and graphics environments.
Buildee provides a Blender-based simulation framework for scene exploration,
reconstruction, and understanding with simulator-grounded geometry and
semantics~\citep{boittiaux2025buildee3dsimulation}. It is complementary to
LEGO-Bench: Buildee emphasizes interactive exploration and simulator tasks,
whereas LEGO-Bench fixes a single RGB observation, hides evaluator geometry and
semantics, and evaluates final executable scene artifacts across diverse
scene families and controlled difficulty levels.

Compared with prior executable-3D benchmarks, LEGO-Bench is
\emph{image-conditioned}, \emph{scene-level}, and \emph{end-to-end}: the
agent must infer a complete executable scene program from visual evidence
rather than generate code from a textual or parametric specification.

\section{Supplementary Analyses}
\label{app:additional_evaluation}

This section collects supplementary analyses on the scope and limits of
LEGO-Anything beyond the paper's main evaluations. We place the strongest
benchmark-adjacent evidence first and retain more exploratory extensions only
as secondary context.

\subsection{Cost--Performance Pareto Frontier}
\label{app:cost_performance_frontier}

\begin{figure}[t]
    \centering
    \includegraphics[width=0.68\linewidth]{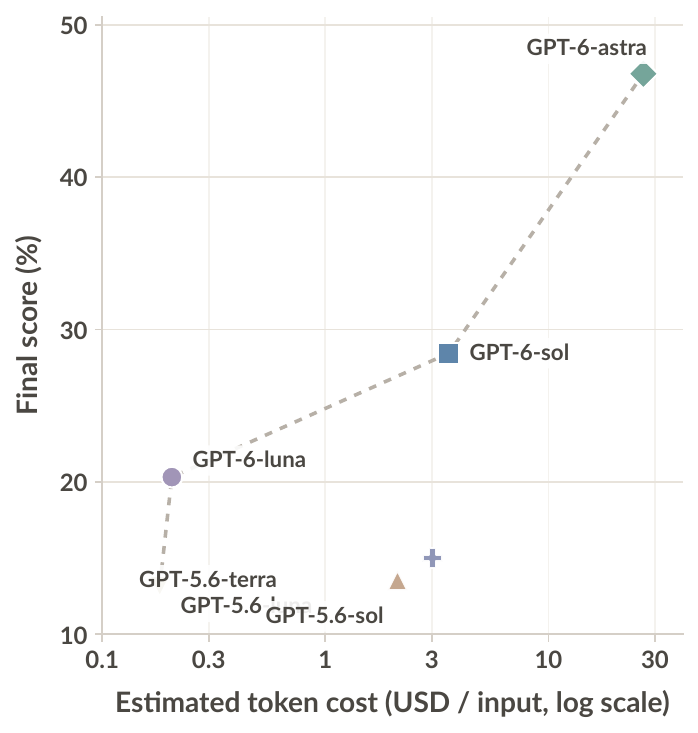}
    \caption{\textbf{Cost--performance frontier for coding agents on LEGO-Bench.} Each point is one coding-agent configuration, using full-benchmark mean Final score and a standardized token-cost proxy per input. The dashed line connects the non-dominated frontier.}
    \label{fig:cost_performance_pareto}
\end{figure}

Figure~\ref{fig:cost_performance_pareto} summarizes the six coding-agent
configurations with one point per model. The horizontal axis uses a
standardized token-cost proxy reconstructed from logged uncached input,
cached input, and output tokens for the retained source attempts; the vertical
axis reports the corresponding benchmark-wide validity-gated Final score.

\finding{\texttt{GPT-5.6-luna} is the absolute cheapest point, but most of the
quality--cost frontier is defined by the GPT-6 family: \texttt{GPT-6-luna}
improves sharply over the lowest-cost regime, \texttt{GPT-6-sol} adds another
quality tier at moderate cost, and \texttt{GPT-6-astra} reaches the highest
score at substantially higher spend. \texttt{GPT-5.6-sol} and
\texttt{GPT-5.6-terra} lie off the frontier.}

\subsection{Construction Trajectory Diagnostics}
\label{app:four_model_trajectory}

\begin{figure*}[t]
    \centering
    \includegraphics[width=0.92\textwidth]{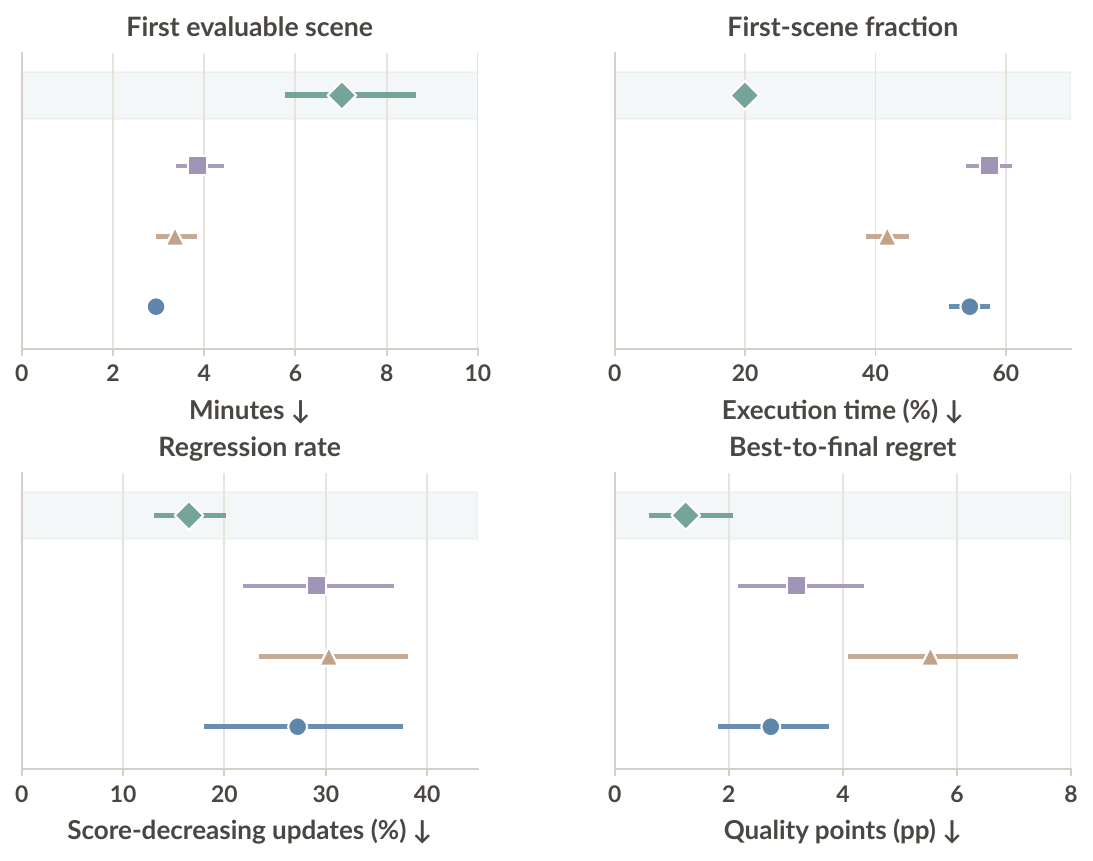}
    \caption{\textbf{Trajectory diagnostics on the common-task intersection.}
    \texttt{GPT-6-astra} reaches the first evaluable scene later in wall-clock
    time but earlier relative to its own runtime, and it has the lowest
    regression rate and best-to-final regret. Points and bars denote means and
    scene-family bootstrap 95\% intervals.}
    \label{fig:trajectory_summary}
\end{figure*}

We analyze the first archived native, high-reasoning-effort runs of
\texttt{GPT-6-astra}, \texttt{GPT-5.6-sol}, \texttt{GPT-5.6-terra}, and
\texttt{GPT-5.6-luna}. Each cohort targets 198 Indoor/Outdoor tasks; the 10
Bird's-Eye tasks are excluded. An eligible checkpoint has a valid Blender
scene, geometry export, and both Reconstruction and Appearance scores.
Checkpoint quality is $Q=(R+A)/2$, computed offline and never exposed to the
actor. Figure~\ref{fig:trajectory_summary} summarizes matched common-task
statistics for first-scene timing, trajectory regression, and final regret.

\finding{\texttt{GPT-6-astra} reaches its first evaluable scene later in
absolute time, but earlier relative to its own runtime, and it shows the
lowest regression rate and best-to-final regret. Construction remains
non-monotonic for all four models, so final submissions should be interpreted
as trajectory endpoints rather than guaranteed best states.}

\begin{figure}[htbp]
    \centering
    \includegraphics[width=\linewidth]{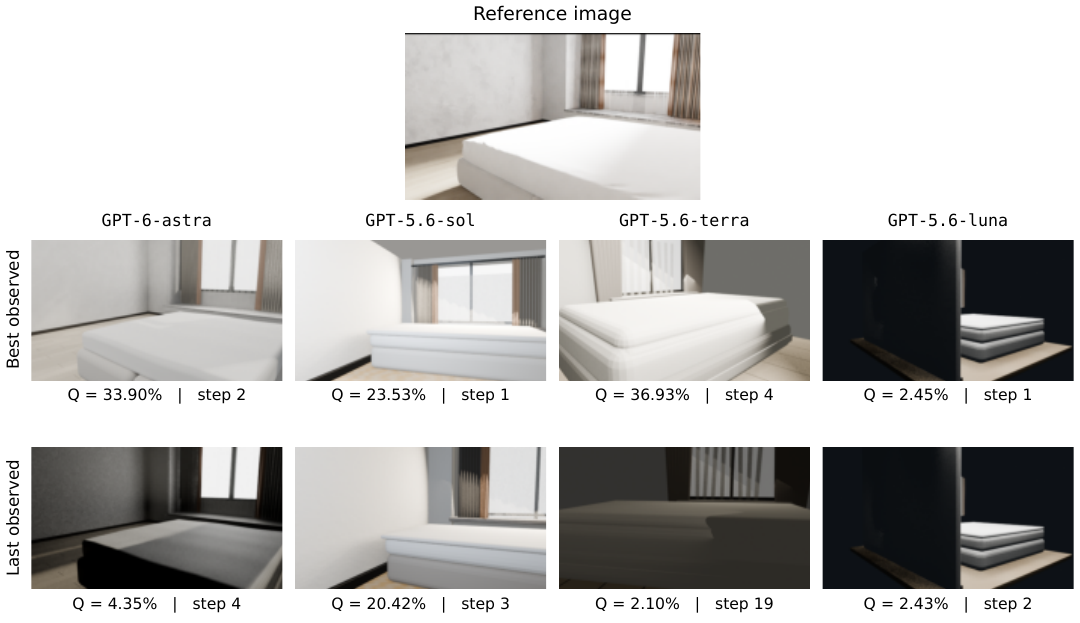}
    \caption{\textbf{A severe best-to-last regression example.}
    Each model is shown on the same House bedroom task at its best and last
    observed checkpoints. Even the strongest model can regress late in the
    trajectory. Displayed $Q$ is the ungated average of Reconstruction and
    Appearance.}
    \label{fig:four_model_trajectory_failure}
\end{figure}

\paragraph{\textbf{Severity-selected failure example.}}
To make this non-monotonicity concrete,
Figure~\ref{fig:four_model_trajectory_failure} shows the common task with the
largest mean best-to-last regret across the four models. We use this case only
as a severity example. It shows that even the strongest model can still
regress sharply late in the construction process.

\subsection{Bird's-Eye Reconstruction Stress Test}
\label{app:birds_eye_stress}

The unpaired NYC Bird's-Eye split contains 10 overhead-view cases per complete
run. The main result table pools these cases with the 90 ground-level Outdoor
views, but we report them separately here only as a stricter large-scale
reconstruction stress test rather than as a matched benchmark split. It is
therefore excluded from the paired scene-complexity analysis. Table
~\ref{tab:lego-bench_birds_eye} reports the three GPT-6 agents under artifact
validity, Reconstruction F@2\%, Appearance, and the corresponding
validity-gated Final score.

\begin{table}[htbp]
    \legotablestyle
    \caption{
        \textbf{Performance ($\%$) on the Bird's-Eye reconstruction stress
        split.} Higher is better for every metric.
    }
    \label{tab:lego-bench_birds_eye}
    \begin{tabular}{l|ccc:c}
        \legotabletoprule
        \textbf{Model + Harness}
        & \textbf{V $\uparrow$}
        & \textbf{R $\uparrow$}
        & \textbf{A $\uparrow$}
        & \textbf{S $\uparrow$} \\
        \midrule
        \texttt{GPT-5.6-sol} + \texttt{Codex}
        & \meanstd{93.3}{5.8}
        & \meanstd{11.2}{7.9}
        & \meanstd{24.3}{3.3}
        & \meanstd{16.4}{3.6} \\

        \texttt{GPT-5.6-terra} + \texttt{Codex}
        & \meanstd{86.7}{5.8}
        & \meanstd{0.9}{1.3}
        & \meanstd{23.0}{4.2}
        & \meanstd{10.7}{2.8} \\

        \texttt{GPT-5.6-luna} + \texttt{Codex}
        & \meanstd{80.0}{10.0}
        & \meanstd{7.6}{6.0}
        & \meanstd{21.1}{2.9}
        & \meanstd{11.4}{2.8} \\

        \midrule
        \texttt{GPT-6-astra} + \texttt{Codex}
        & \meanstd{\textbf{100.0}}{0.0}
        & \meanstd{\textbf{24.0}}{5.0}
        & \meanstd{\textbf{46.6}}{0.3}
        & \meanstd{\textbf{35.3}}{2.5} \\

        \texttt{GPT-6-sol} + \texttt{Codex}
        & \meanstd{100.0}{0.0}
        & \meanstd{17.8}{2.6}
        & \meanstd{40.0}{0.8}
        & \meanstd{28.9}{1.0} \\

        \texttt{GPT-6-luna} + \texttt{Codex}
        & \meanstd{100.0}{0.0}
        & \meanstd{6.7}{4.7}
        & \meanstd{34.4}{0.7}
        & \meanstd{20.0}{2.7} \\
        \legotablebottomrule
    \end{tabular}
\end{table}

\finding{\texttt{GPT-6-astra} is the strongest agent on the Bird's-Eye stress
test, and the GPT-6 family substantially outperforms the GPT-5.6 family.
However, all models still show a clear gap between artifact delivery and much
weaker strict-geometry reconstruction under F@2\%.}

\subsection{Pilot Scene-Level Articulation Evaluation on LEGO-Bench}
\label{app:articulation_evaluator_pilot}

\paragraph{\textbf{Motivation.}}
Interactive 3D scenes must capture not only geometry and appearance, but also
which objects can move and how they move
~\citep{zhou2026digital,xiang2020sapien}. Prior work such as
Articulate-Anything and ArtiCraft
~\citep{le2025articulateanythingautomaticmodelingarticulated,zhou2026articraftagenticscalablearticulated}
focuses on individual objects. Scene-level evaluation is harder
because it must identify articulatable instances, recover their movable
structure, and avoid assigning false motion to static objects.

\paragraph{\textbf{Canonical GT coverage.}}
This pilot uses the canonical articulation registry described in Appendix
~\ref{app:articulated_asset_annotation}. LEGO-Bench provides one
human-validated reference articulation for each articulated asset together
with confirmed-static controls; Table~\ref{tab:articulation_gt_coverage}
summarizes the resulting coverage of the canonical manifest.

\paragraph{\textbf{Pilot diagnostic and result.}}
As an initial scene-level diagnostic, each eligible canonical joint
requirement is classified as \emph{object missing}, \emph{part missing},
\emph{joint missing}, \emph{wrong joint type}, or \emph{correct}. For one
articulated object, the provisional score is
\begin{equation}
    S_{\mathrm{Art}}^{\mathrm{pilot}}
    =
    \frac{2C}
    {2C + O + P + J + 2W + E},
\end{equation}
where $C$, $O$, $P$, $J$, $W$, and $E$ denote correct requirements, missing
objects, missing parts, missing joints, wrong joint types, and extra joints,
respectively. Scene scores macro-average articulated target objects and
predictions that place movable joints on matched human-confirmed static
objects receive zero.

We reevaluate one frozen Office submission without changing its predicted
Blender scene. Under the earlier partial GT adapter, only two articulated
targets were scoreable and 12 were unavailable. The complete canonical
manifest makes all 14 articulated targets scoreable, retains 12
human-confirmed static targets as false-motion controls, and reduces the
unavailable count to zero. The submission still obtains
$S_{\mathrm{Art}}^{\mathrm{pilot}}=0$, so the zero reflects the prediction
rather than missing articulation GT.

\finding{Under the complete canonical articulation manifest, the pilot Office
submission still obtains $S_{\mathrm{Art}}^{\mathrm{pilot}}=0$. A more concrete scenario for tenable scene-level articulation needs further exploration.}

\section{Demos on LEGO-Bench}
\label{app:lego_bench_demos}

Figure~\ref{fig:appendix_c_examples} shows qualitative examples from the same
\texttt{run\_01} evaluation split used in the main LEGO-Bench experiments.
Each row uses one reference image on the left and the evaluator-rendered final
artifacts from six coding-agent configurations on the right.

\begin{center}
    \centering
    \setlength{\tabcolsep}{1pt}
    \renewcommand{\arraystretch}{0.96}
    \scriptsize
    \begin{tabular}{@{}ccccccc@{}}
        \textbf{Reference} & \textbf{5.6 Luna} & \textbf{5.6 Terra} &
        \textbf{5.6 Sol} & \textbf{6 Luna} & \textbf{6 Sol} &
        \textbf{6 Astra} \\
        \multicolumn{7}{@{}l}{\textbf{Example 1: Havana rum shop street}} \\
        \includegraphics[width=0.135\linewidth]{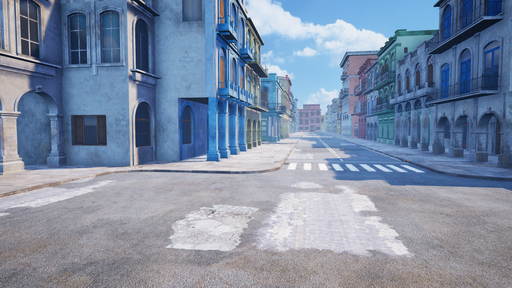} &
        \includegraphics[width=0.135\linewidth]{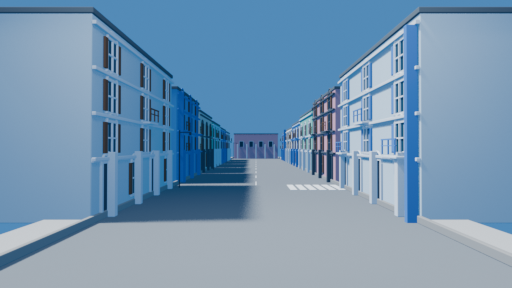} &
        \includegraphics[width=0.135\linewidth]{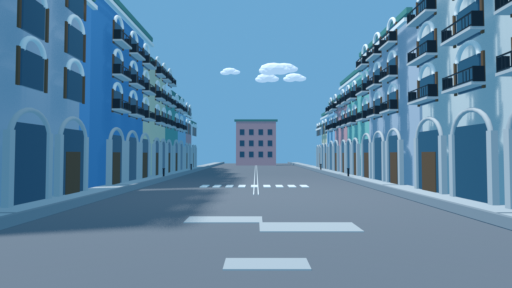} &
        \includegraphics[width=0.135\linewidth]{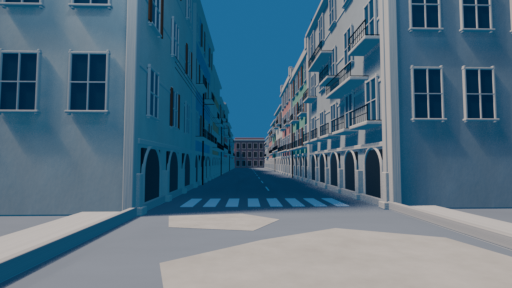} &
        \includegraphics[width=0.135\linewidth]{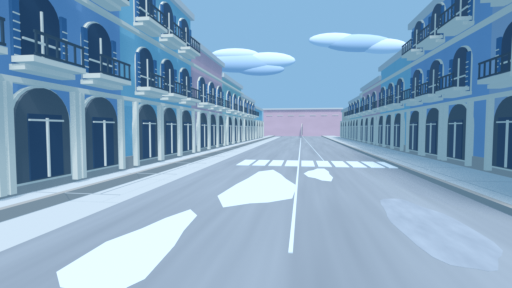} &
        \includegraphics[width=0.135\linewidth]{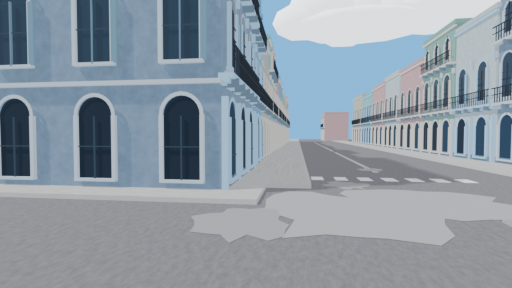} &
        \includegraphics[width=0.135\linewidth]{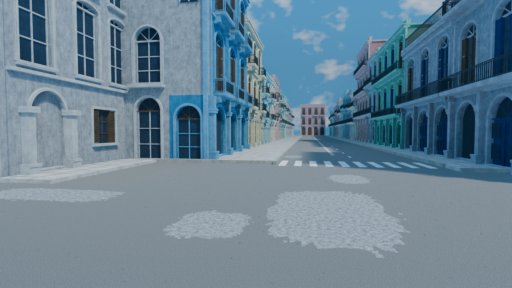} \\
        \multicolumn{7}{@{}l}{\textbf{Example 2: NYC East Village}} \\
        \includegraphics[width=0.135\linewidth]{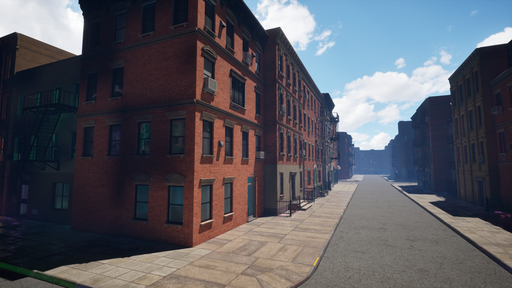} &
        \includegraphics[width=0.135\linewidth]{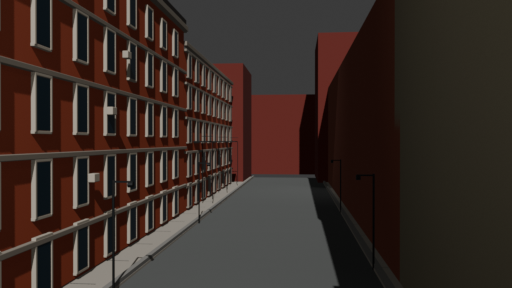} &
        \includegraphics[width=0.135\linewidth]{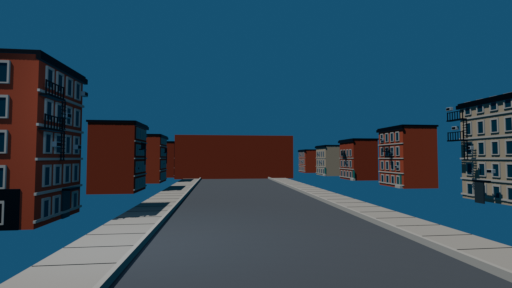} &
        \includegraphics[width=0.135\linewidth]{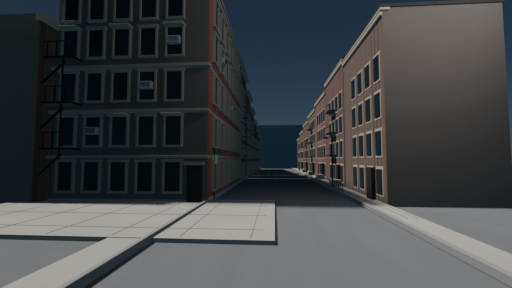} &
        \includegraphics[width=0.135\linewidth]{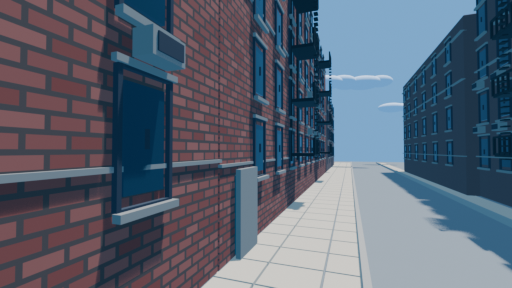} &
        \includegraphics[width=0.135\linewidth]{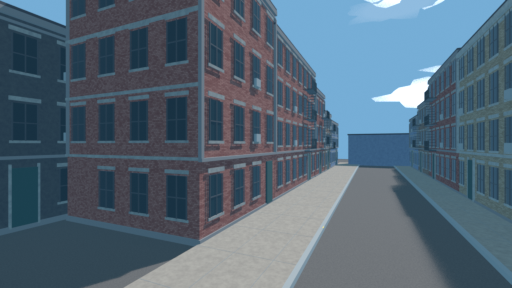} &
        \includegraphics[width=0.135\linewidth]{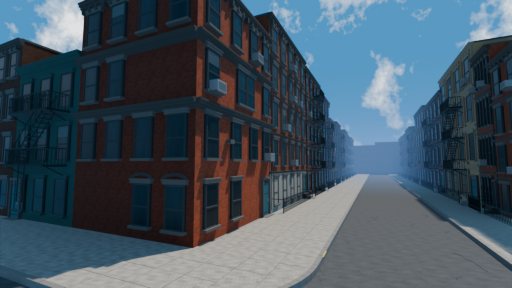} \\
        \multicolumn{7}{@{}l}{\textbf{Example 3: House bedroom}} \\
        \includegraphics[width=0.135\linewidth]{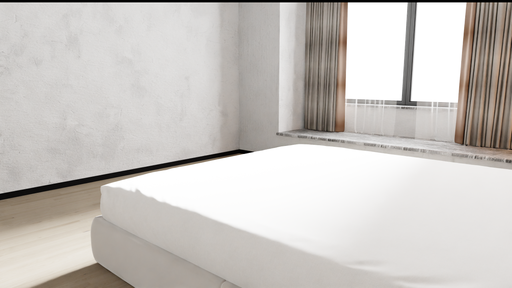} &
        \includegraphics[width=0.135\linewidth]{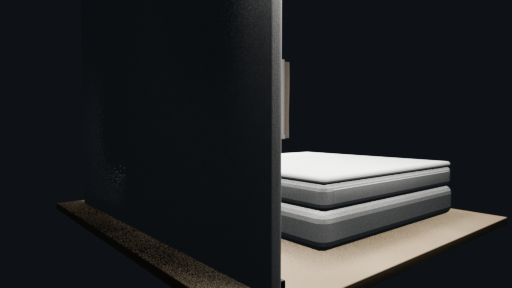} &
        \includegraphics[width=0.135\linewidth]{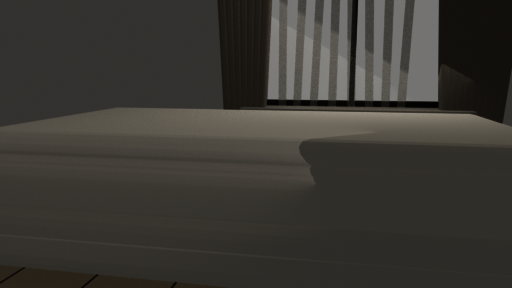} &
        \includegraphics[width=0.135\linewidth]{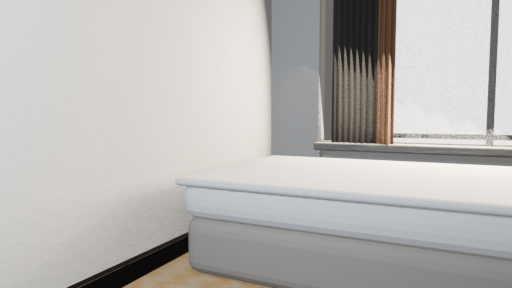} &
        \includegraphics[width=0.135\linewidth]{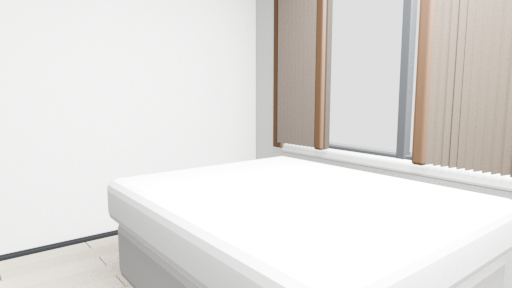} &
        \includegraphics[width=0.135\linewidth]{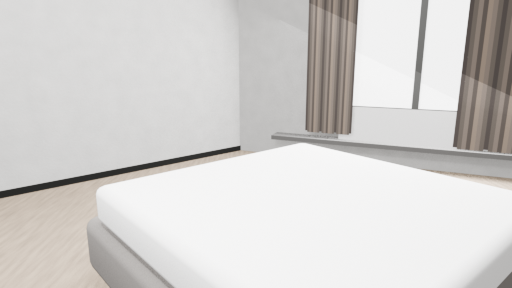} &
        \includegraphics[width=0.135\linewidth]{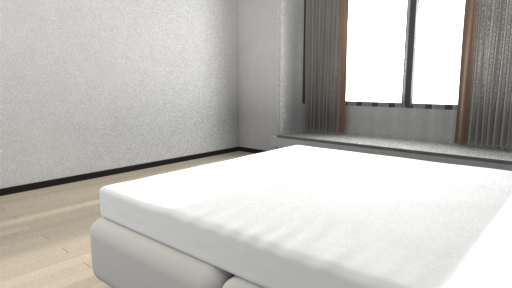} \\
        \multicolumn{7}{@{}l}{\textbf{Example 4: House kitchen}} \\
        \includegraphics[width=0.135\linewidth]{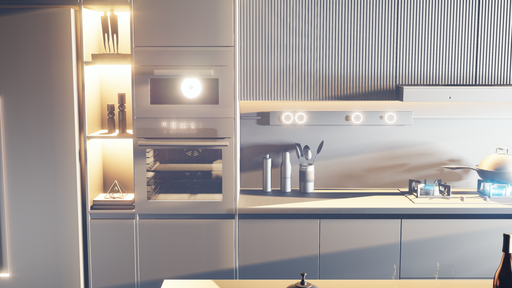} &
        \includegraphics[width=0.135\linewidth]{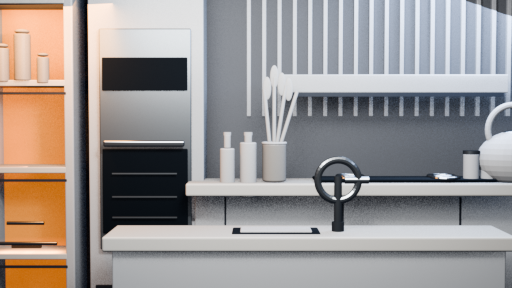} &
        \includegraphics[width=0.135\linewidth]{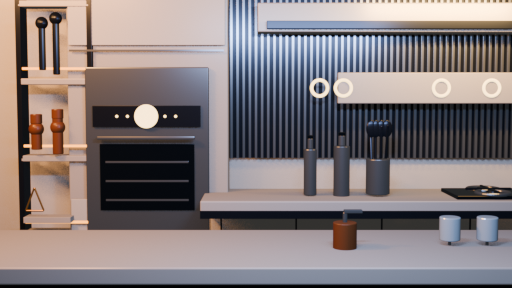} &
        \includegraphics[width=0.135\linewidth]{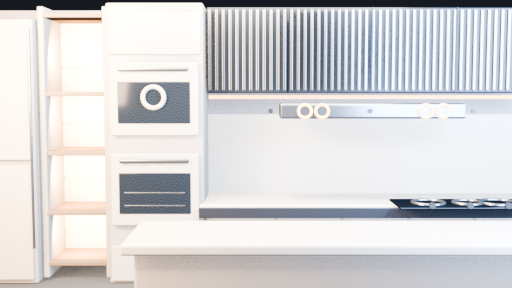} &
        \includegraphics[width=0.135\linewidth]{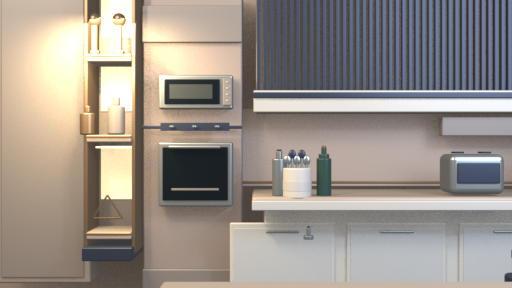} &
        \includegraphics[width=0.135\linewidth]{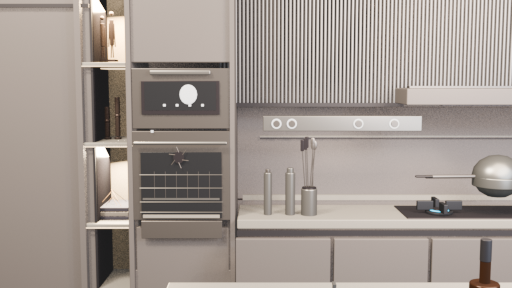} &
        \includegraphics[width=0.135\linewidth]{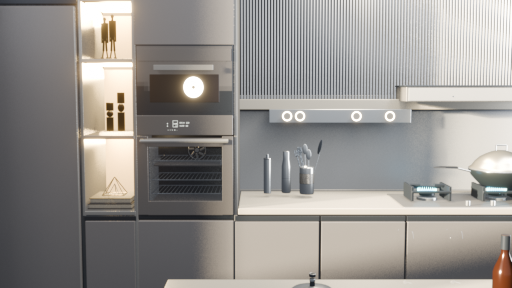} \\
        \multicolumn{7}{@{}l}{\textbf{Example 5: Office open-plan}} \\
        \includegraphics[width=0.135\linewidth]{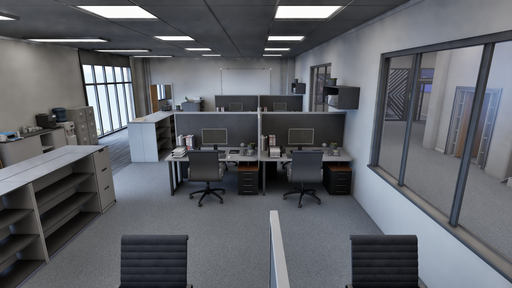} &
        \includegraphics[width=0.135\linewidth]{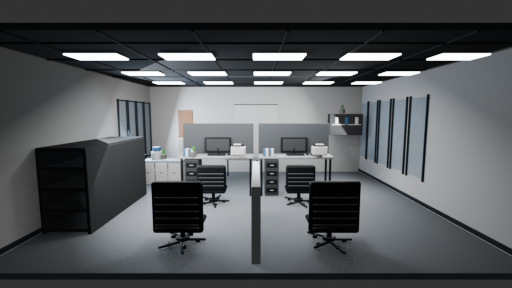} &
        \includegraphics[width=0.135\linewidth]{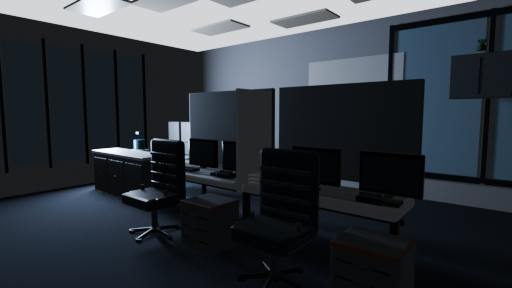} &
        \includegraphics[width=0.135\linewidth]{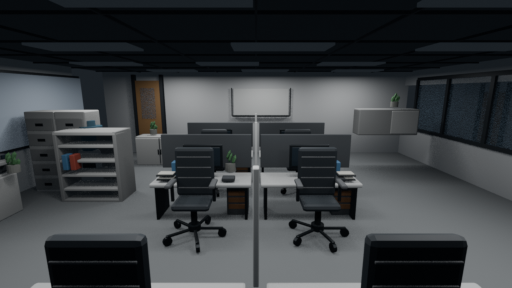} &
        \includegraphics[width=0.135\linewidth]{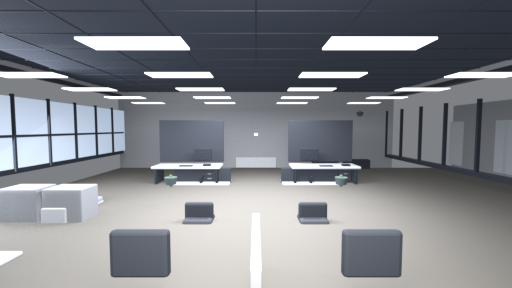} &
        \includegraphics[width=0.135\linewidth]{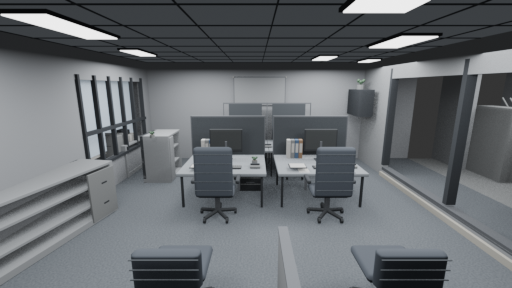} &
        \includegraphics[width=0.135\linewidth]{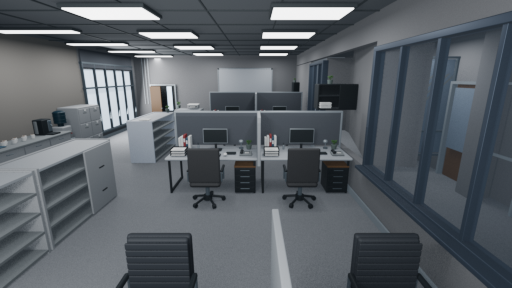} \\
        \multicolumn{7}{@{}l}{\textbf{Example 6: Warehouse}} \\
        \includegraphics[width=0.135\linewidth]{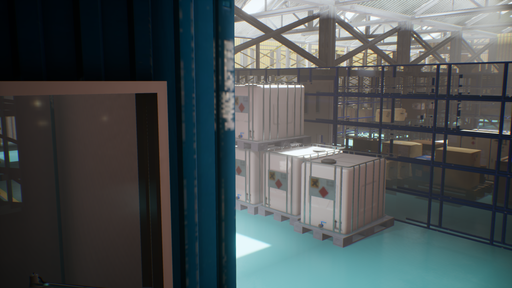} &
        \includegraphics[width=0.135\linewidth]{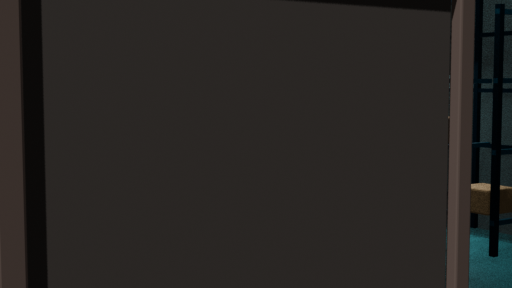} &
        \includegraphics[width=0.135\linewidth]{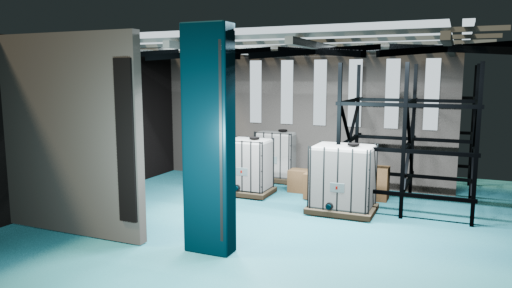} &
        \includegraphics[width=0.135\linewidth]{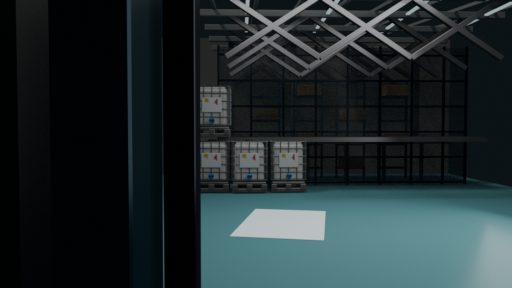} &
        \includegraphics[width=0.135\linewidth]{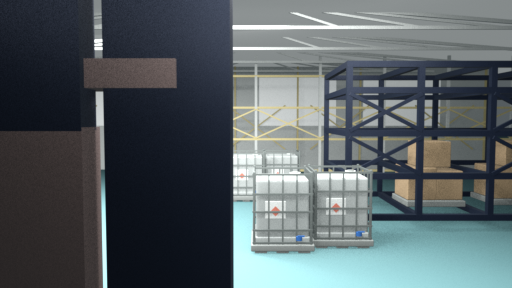} &
        \includegraphics[width=0.135\linewidth]{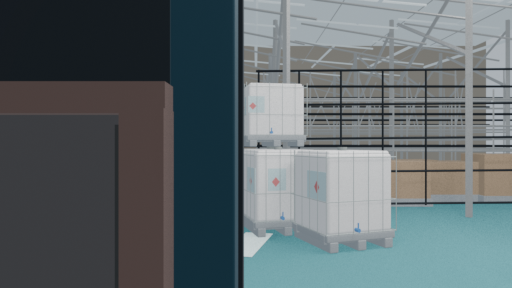} &
        \includegraphics[width=0.135\linewidth]{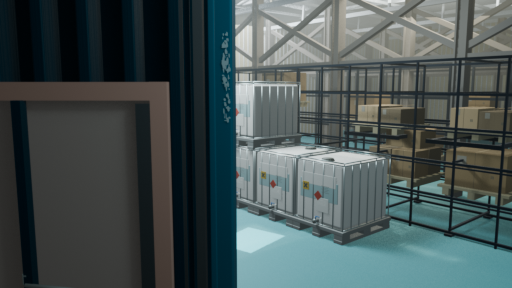}
    \end{tabular}
    \captionof{figure}{\textbf{Qualitative LEGO-Bench examples.}
    Columns show the reference image followed by final evaluator rerenders from
    \texttt{GPT-5.6-luna}, \texttt{GPT-5.6-terra}, \texttt{GPT-5.6-sol},
    \texttt{GPT-6-luna}, \texttt{GPT-6-sol}, and \texttt{GPT-6-astra}.}
    \label{fig:appendix_c_examples}
\end{center}

\section{LEGO-Bench Construction and Statistics}
\label{app:benchmark_details}

\subsection{Scene Construction and Capture}

LychSim provides the scene-construction, simulation, and ground-truth capture
backend for LEGO-Bench. We use it to assemble scene geometry, object layout,
support relations, lighting, and materials, then render benchmark inputs and
record the hidden annotations used for evaluation. Candidate scenes are
included only after collision and stability checks and human review of the
resulting scene composition and capture quality.

\paragraph{\textbf{Task inputs and ground truth.}}
Each task exposes only the public input needed for end-to-end reconstruction,
including the reference image, basic camera metadata, and the prediction
schema. The geometric and correspondence-level ground truth used for
evaluation remains private to the benchmark.

\paragraph{\textbf{Evaluator coordinate convention.}}
For evaluation, simulator outputs are converted from LychSim's left-handed
centimetre coordinates to the evaluator's right-handed metric coordinates with
$+Z$ up:
\begin{equation}
    \mathbf{p}_{B}
    =
    \mathbf{D}\mathbf{p}_{U}/100,
    \qquad
    \mathbf{R}_{B}
    =
    \mathbf{D}\mathbf{R}_{U}\mathbf{D},
    \qquad
    \mathbf{D}=\operatorname{diag}(1,-1,1).
\end{equation}
Subscripts $U$ and $B$ denote the simulator and evaluation coordinate systems,
respectively.

\subsection{Controlled Scene Complexity}

LEGO-Bench controls difficulty within matched scene families rather than by a
global object-count threshold. For each paired family, object membership is
nested, with
\(
\mathcal{O}_{E}\subset\mathcal{O}_{M}\subset\mathcal{O}_{H},
\)
so higher tiers add visible scene content while keeping shared content fixed.
Across the three tiers, architecture, lighting, materials, camera, and the
asset identity, transform, scale, and bounding box of every shared object are
held constant. Every retained object must preserve support consistency and be
visible in at least one reference view; families that violate these conditions
are rejected. Difficulty is therefore defined by controlled additions within a
scene family rather than by raw object count alone. The Aerial XHard split has
no matched family and is excluded from the paired complexity analysis.

\subsection{Dataset Statistics}
\label{app:benchmark_statistics}

\begin{figure}[t]
    \centering
    \includegraphics[width=\linewidth]{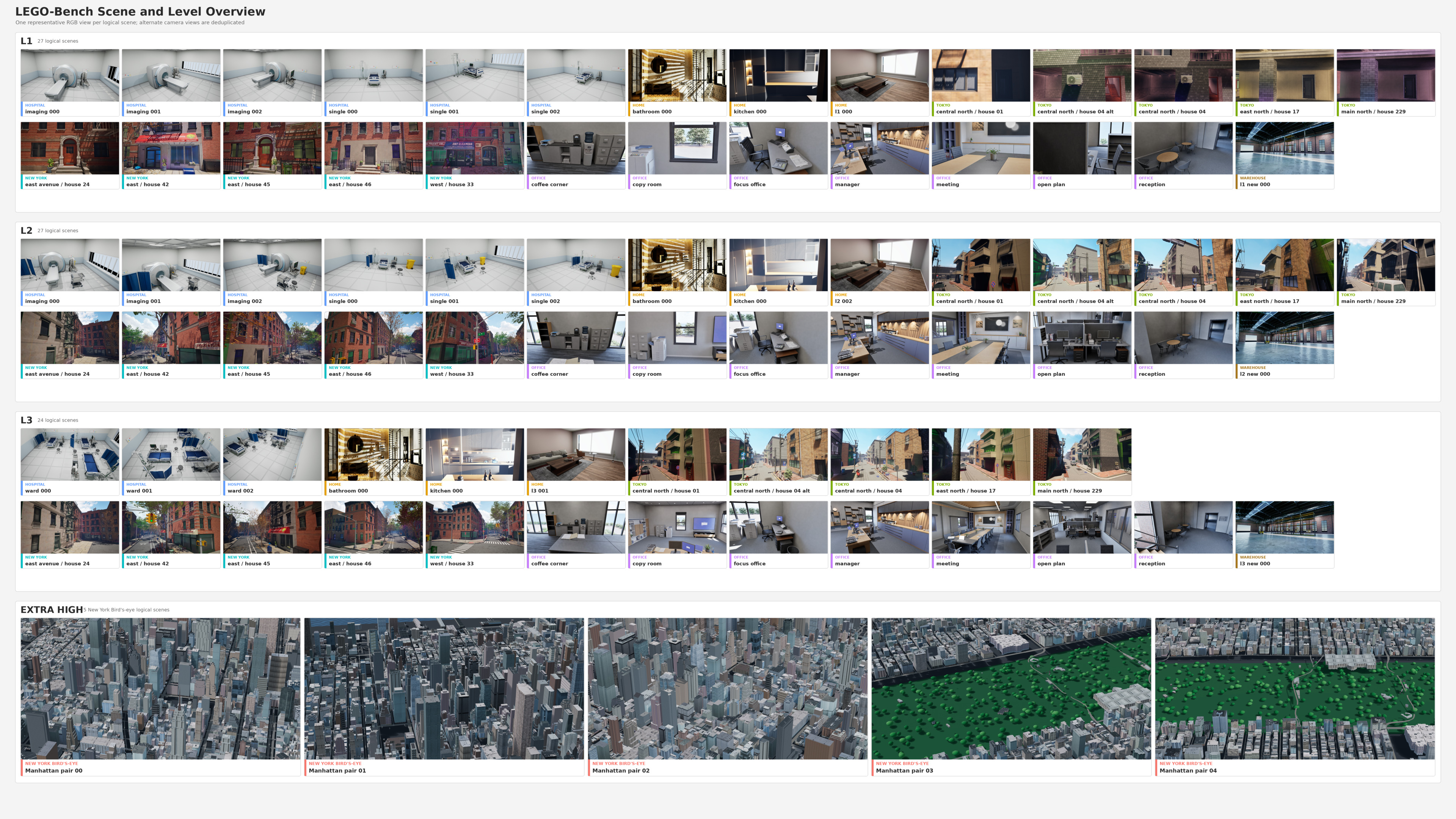}
    \caption{\textbf{Representative LEGO-Bench inputs.}
    LEGO-Bench includes indoor, outdoor, and Bird's-Eye reconstruction
    settings with controlled scene complexity.}
    \label{fig:overview}
\end{figure}

\begin{figure}[t]
    \centering
    \includegraphics[width=\linewidth]{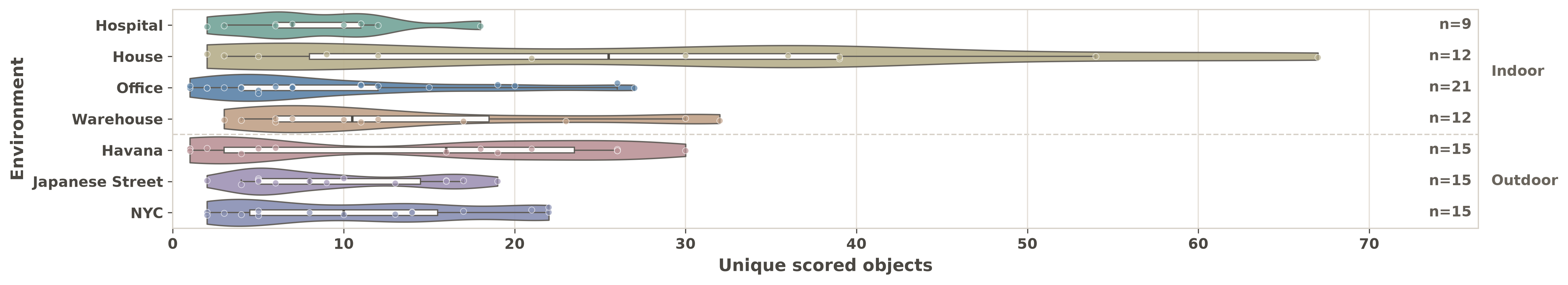}
    \caption{\textbf{Scored-object count per logical scene.}
    Counts increase within paired families, but global count is not used as the
    formal difficulty definition.}
    \label{fig:distribution_object_number}
\end{figure}

\begin{figure}[t]
    \centering
    \includegraphics[width=\linewidth]{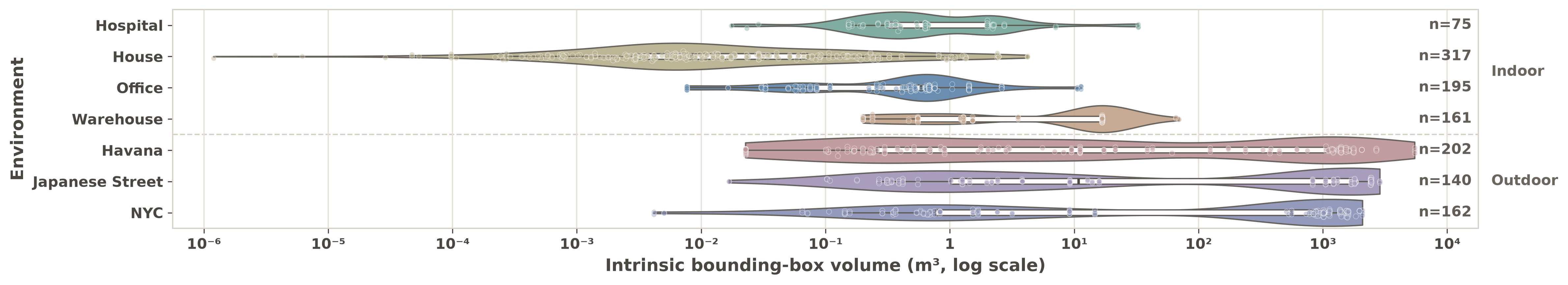}
    \caption{\textbf{Distribution of scored-object size.}
    Object scale spans small manipulable items through architectural and
    city-scale structures.}
    \label{fig:distribution_object_size}
\end{figure}

Figure~\ref{fig:overview} shows representative benchmark inputs.
Figures~\ref{fig:distribution_object_number} and
\ref{fig:distribution_object_size} report scene inventory and object-size
distributions. 

\subsection{Articulated-Asset Annotation}
\label{app:articulated_asset_annotation}

\paragraph{\textbf{Annotated assets.}}
The frozen benchmark registry contains 443 canonical assets, each of which
received a human articulation judgment. The archived review log additionally
contains 597 historical records, including superseded asset identifiers kept
for provenance. We map these records back to the canonical registry before
computing statistics, so all counts below are reported over canonical assets
rather than raw historical entries. This is the same 443-asset registry
summarized in Table~\ref{tab:lego_bench_statistics}. Execution records for the planned
\texttt{GPT-5.6-terra} proposal workflow are tracked separately in
the source manifest and do not affect the reported benchmark statistics.

\paragraph{\textbf{Proposal and verification workflow.}}
Annotation is human-verified throughout. For each candidate asset, annotators
inspect the reference observation together with the available source geometry.
\texttt{P3-SAM}
~\citep{ma2025p3samnative3dsegmentation} initializes the part segmentation,
while \texttt{X-Part}~\citep{yan2025xparthighfidelitystructure} provides candidate part
decompositions and articulation structures. Before joint inference, an
object-first stage groups mesh components into logical rigid parts and labels
each part as fixed, movable, or uncertain. Candidate joints then specify
parent--child relations, joint type, axis, origin, motion limits, and rest
state. Deterministic checks validate the URDF structure, mesh references, and
executable joint motion. A reviewer then compares the complete object, part
decomposition, candidate structure, and rendered motion evidence. Acceptance
requires complete coverage of the object's functional movable components,
correct rigid-part grouping and mechanism geometry, and individually
observable and executable joint motion. Reviewers may mark an object as
static, validate its articulation, or assign repair labels such as missing
joint, incorrect grouping, joint type, axis, pivot, or motion range.
Before acceptance, each motion is executed in the simulator and visually
checked, including the asset's return to its rest state. All automatically
generated outputs remain proposals and are never promoted directly to benchmark
ground truth without human review.

\begin{figure}[t]
    \centering
    \includegraphics[width=\linewidth]
        {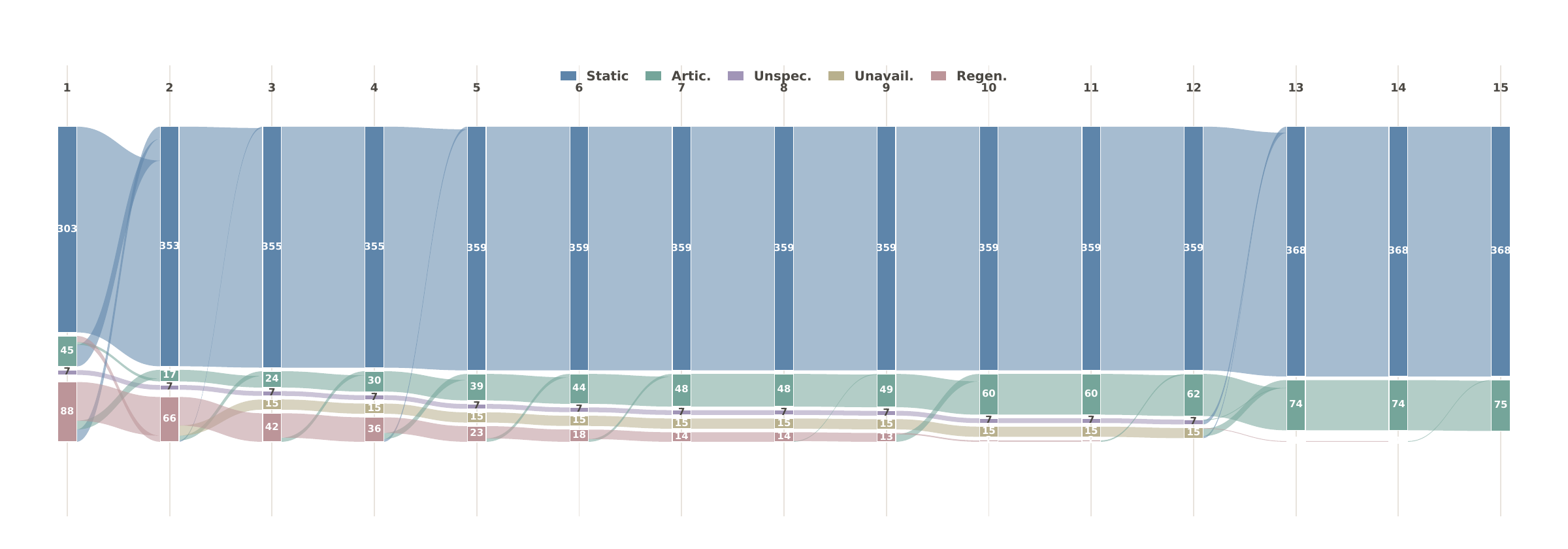}
    \caption{\textbf{Human-in-the-loop articulation annotation process.}
    The figure traces how the 443 canonical assets move across review states
    over 15 human annotation rounds, from initial review to static acceptance,
    articulated acceptance, or regeneration. The second round deliberately
    re-examines both first-pass articulation positives and unresolved assets,
    so the articulated count need not increase monotonically. After 15 rounds,
    all assets are resolved as 368 static and 75 articulated.}
    \label{fig:annotation_flow}
\end{figure}

\paragraph{\textbf{Annotation results.}}
Figure~\ref{fig:annotation_flow} provides a Sankey-style view of the
round-by-round annotation flow, showing how assets move from initial review to
static acceptance, articulated acceptance, or regeneration as evidence is
refined across rounds.
After 15 human annotation rounds, all 443 canonical assets are resolved:
368 (83.1\%) are confirmed static and 75 (16.9\%) have validated articulation,
with no unspecified or unresolved assets remaining. Each accepted annotation is
linked to the reviewed asset revision and its motion-validation evidence. Only
human-reviewed and execution-verified annotations enter Articulation
evaluation, using the matching and aggregation rules in
Appendix~\ref{app:metric_details}.

\begin{table}[t]
    \legotablestyle
    \caption{\textbf{Coverage of the canonical articulation GT manifest.}
    Joint requirements are parsed from the exact human-accepted candidate
    URDFs.}
    \label{tab:articulation_gt_coverage}
    \begin{tabular}{l|r}
        \legotabletoprule
        \textbf{Manifest statistic} & \textbf{Count} \\
        \midrule
        Canonical assets & 443 \\
        Confirmed static & 368 \\
        Validated articulated & 75 \\
        Joint requirements & 327 \\
        \hspace{1em}Continuous & 75 \\
        \hspace{1em}Revolute & 55 \\
        \hspace{1em}Prismatic & 197 \\
        GT unavailable & 0 \\
        \legotablebottomrule
    \end{tabular}
\end{table}

\section{Metric Implementation and Validation}
\label{app:metric_details}

The primary comparison uses Validity, Reconstruction, Appearance, and their
validity-gated Final score. Layout and articulation are auxiliary protocols
and are documented separately after the headline metrics and validation plan.

\begin{figure*}[t]
\centering
\includegraphics[width=\textwidth]{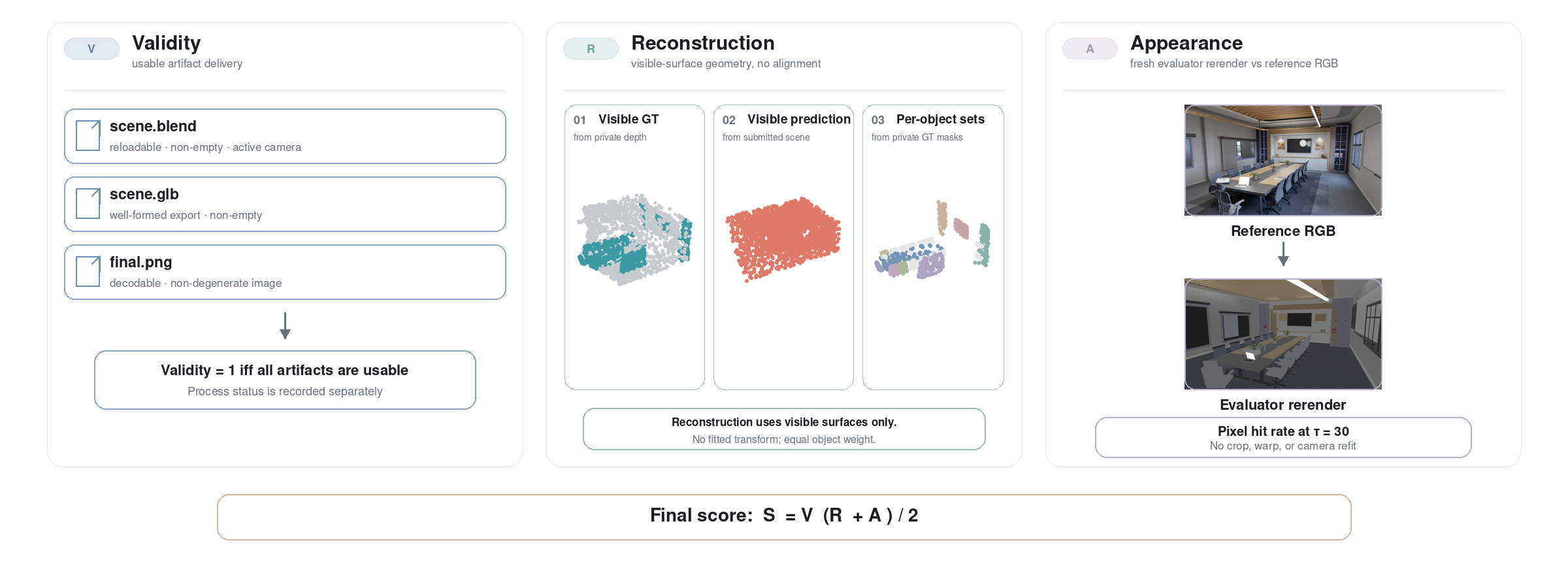}
\caption{\textbf{Overview of the LEGO-Bench metrics.} Validity checks usable scene artifacts; the main table also treats unresolved headline-evaluation failures as invalid. Reconstruction is illustrated with step-wise point-cloud states from the evaluator: visible GT surfaces from private depth, visible prediction surfaces from the submitted scene, and per-object sets induced by private masks. Appearance compares a fresh evaluator rerender against the reference image without geometric warping or camera refitting.}
\label{fig:metric_overview}
\end{figure*}

\subsection{Artifact Checks and Reported Validity}

The raw artifact-validity check measures deliverability rather than reconstruction
quality. Artifacts pass this check iff the submission contains a reloadable non-empty
\texttt{scene.blend} with at least one mesh and an active camera, a non-empty
\texttt{scene.glb}, and a decodable non-degenerate render. This check is
determined from the submitted artifacts themselves rather than from process
status: a timeout or interrupted run may still leave a valid submission, while
a clean process exit does not by itself establish validity. Artifact-invalid
attempts remain in the denominator and receive zero validity-gated Final score.
For the main table, reported Validity additionally requires completed
headline evaluation: an unresolved verifier error receives zero penalties
for all four columns. Historical diagnostic analyses retain their stated
artifact-validity and missingness conventions.
This distinction is important in agent evaluation more broadly: environment-
construction and agent-benchmark work increasingly treats usable artifact
delivery as a first-class outcome, rather than assuming that nominal task
completion or process exit is sufficient evidence of success
~\citep{li2026clawenvkitautomaticenvironmentgeneration,ye2026clawevaltrustworthyevaluationautonomous}.

\subsection{Reconstruction Alignment and Scoring}
\label{app:reconstruction_scoring}

The headline Reconstruction metric performs no registration. Submitted geometry
is evaluated directly in the camera frame induced by its active camera, and the
evaluator does not fit or apply a global or per-object transform. In
particular, it does not absorb errors through fitted translation, rotation,
scale, reflection, anisotropic scaling, or a Layout transform. The evaluator
rasterizes the submitted triangle geometry to obtain submitted visible
surfaces, back-projects private depth to obtain reference-visible surfaces, and
uses private instance masks to select the scored surface scopes.

\paragraph{\textbf{Point-to-instance assignment.}}
The implementation assigns surfaces to object scopes in the reference image
plane, not by matching predicted objects to ground-truth objects. Each scored
reference instance has a private RGB segmentation color, which defines a binary
mask $M_o$. The reference set $G_o$ is obtained by back-projecting private
depth samples whose pixels lie in $M_o$, after invalid/far-depth filtering and
reference-camera visibility. The submitted set is obtained by rasterizing the
submitted triangles from the fixed camera frame, retaining the visible
submitted surface, projecting each visible submitted point into the same
reference image plane, and keeping it in $P_o$ iff its rounded projected pixel
lies in $M_o$.

The evaluator therefore does not use predicted object names, categories, mesh
grouping, or a Hungarian/IoU matching step to establish object
correspondence. A camera, scale, boundary, occlusion, duplicate-object, or
category error is penalized through where the submitted surface projects: it
may be assigned to the wrong private instance mask, fall outside all scored
object masks, or fail the geometric nearest-neighbor test below. Visible
submitted geometry outside the scored object masks is excluded from the
headline per-object macro score, while companion object-union/visible-scene
diagnostics and the evaluator-rendered Appearance score expose broader-scope
errors.

\paragraph{\textbf{Nearest-neighbor tests.}}
For each scored instance $o$, let $G_o$ be the reference-visible point set and
$P_o$ be the predicted visible point set assigned by the rule above. Recall is
computed in the reference-to-prediction direction:
\[
    \mathrm{Rec}_{o}
    =
    \frac{1}{|G_o|}
    \sum_{g\in G_o}
    \mathbf{1}\!\left[
        \min_{p\in P_o}\|g-p\|_2 \leq \tau(g)
    \right],
\]
with zero recall when $G_o$ is non-empty and $P_o$ is empty. Precision uses the
opposite nearest-neighbor direction. For each $p\in P_o$, define
$g^\star(p)=\arg\min_{g\in G_o}\|p-g\|_2$; the reference-dependent tolerance is
then evaluated at this nearest reference point:
\[
    \mathrm{Prec}_{o}
    =
    \frac{1}{|P_o|}
    \sum_{p\in P_o}
    \mathbf{1}\!\left[
        \|p-g^\star(p)\|_2 \leq \tau(g^\star(p))
    \right].
\]
If $P_o$ is empty, precision is defined as zero for the F1 denominator. These
rules make unmatched visible predicted geometry inside a scored mask count as
false-positive surface mass for that instance, and missing reference-visible
surface count as false-negative mass.

For each eligible object, precision and recall use a pointwise tolerance
$\tau(g)=0.05z(g)$, where $z(g)$ is the reference point's positive forward
depth in the reference-camera frame. There is no fixed metre floor or cap in
the headline tolerance. The object score is the harmonic mean of precision and
recall, and the trial score is the equal-weight macro average over eligible
scored objects. Objects with no target points after sampling are marked
ineligible rather than included in the macro average; objects with target
points but no submitted points receive zero precision, recall, and F1. The
whole-scene scoring representation is deterministically capped at 100,000
points before the object masks are applied. Missing visible surfaces therefore
reduce recall, extra visible geometry inside a scored mask reduces precision,
and camera or scale errors are penalized directly rather than corrected after
the fact.

The evaluator also reports stricter 2\% and relaxed 10\% depth-relative
variants, fixed-distance and globally aligned diagnostics, normalized
symmetric Chamfer-$L_1$, and visible-surface voxel IoU. These are auxiliary
diagnostics only; the headline metric remains the no-alignment per-object
F@5\% score.

\paragraph{\textbf{Auxiliary reconstruction diagnostics.}}
The strict and relaxed depth-relative variants keep the same visible-surface
point sets and per-object macro aggregation, but replace the headline
tolerance by
\begin{equation}
    \tau_{\alpha}(g)=\alpha\,z(g), \qquad \alpha\in\{0.02,0.10\},
\end{equation}
yielding F@2\% and F@10\%, respectively. Historical fixed-distance diagnostics
instead use a constant threshold $\tau_{\delta}(g)=\delta$ with
$\delta\in\{0.05,0.10\}$ m. 

\subsection{Evaluator-Rendered Appearance}
\label{app:appearance_metric}

Appearance uses a fresh render of the submitted Blender scene. The evaluator
preserves its active camera, geometry, materials, lights, and color settings,
while fixing the rendering engine, output resolution, and full-frame output.
The reference is resized to that resolution without fitting a spatial warp,
crop, camera transform, or photometric correction. For 8-bit sRGB images,
a pixel is correct only if its largest absolute channel difference is at most
30. Appearance is the fraction of correct pixels, reported as a percentage.
The submitted \texttt{final.png} is checked for artifact delivery but does not
substitute for this evaluator render.

This measure tests agreement with the reference rendering under a fixed
protocol. It is sensitive to lighting, materials, camera error, and geometry,
and is not a perceptual similarity model or an independent measure of
geometric correctness. Reconstruction and Appearance are therefore reported
separately as well as through their validity-gated average.

\begin{figure*}[t]
\centering
\begin{subfigure}[t]{0.32\textwidth}
    \centering
    \includegraphics[width=\linewidth]{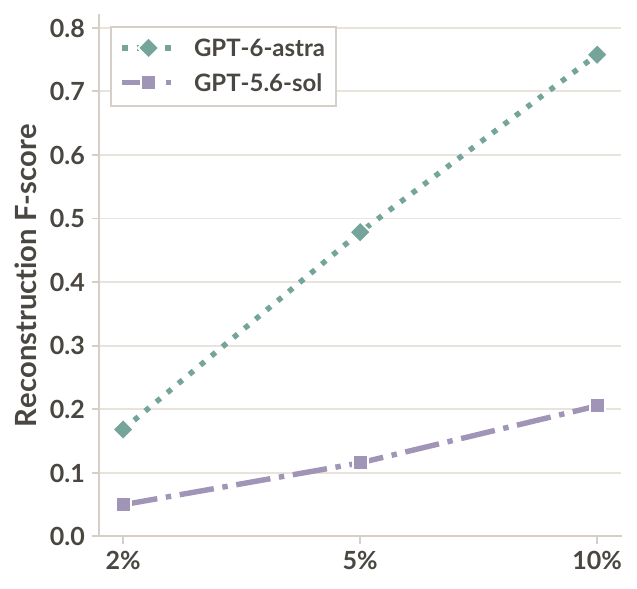}
    \caption{Reconstruction threshold.}
    \label{fig:metric_sensitivity_small_reconstruction}
\end{subfigure}
\hfill
\begin{subfigure}[t]{0.32\textwidth}
    \centering
    \includegraphics[width=\linewidth]{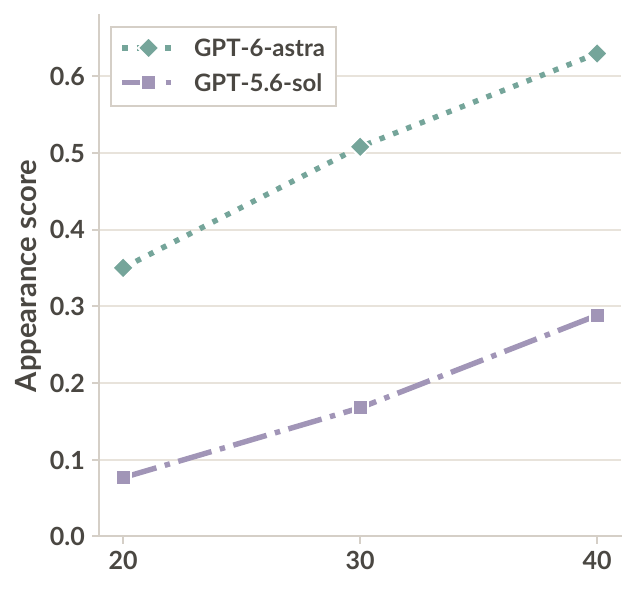}
    \caption{Appearance tolerance.}
    \label{fig:metric_sensitivity_small_appearance}
\end{subfigure}
\hfill
\begin{subfigure}[t]{0.32\textwidth}
    \centering
    \includegraphics[width=\linewidth]{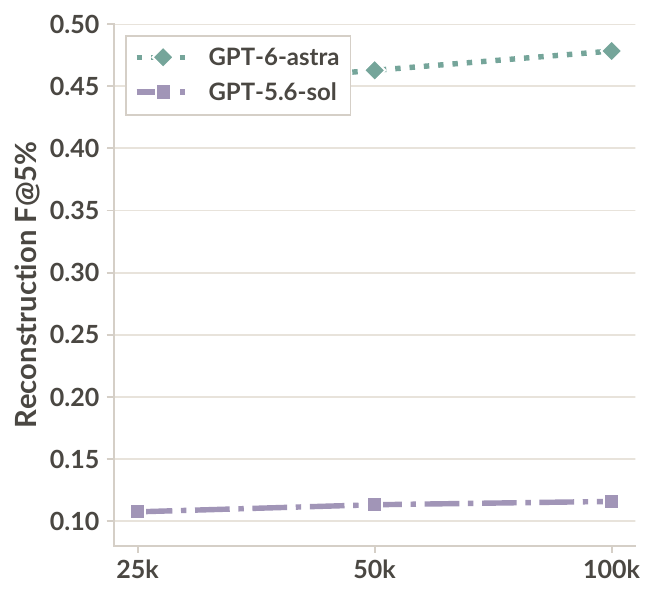}
    \caption{Point-budget sensitivity.}
    \label{fig:metric_sensitivity_small_points}
\end{subfigure}
\caption{\textbf{Small-scale metric sensitivity on a fixed shared subset.}
Left: no-alignment per-object Reconstruction under F@2\%, F@5\%, and F@10\%.
Middle: evaluator-rendered Appearance under RGB hit tolerances 20, 30, and 40.
Right: Reconstruction F@5\% under 25k, 50k, and 100k sampled-point budgets.
Across all three perturbations, \texttt{GPT-6-astra} remains ahead of
\texttt{GPT-5.6-sol}; the conclusions are therefore not driven by a single
threshold or point-sampling choice in this subset analysis.}
\label{fig:metric_sensitivity_small}
\end{figure*}

Figure~\ref{fig:metric_sensitivity_small} reports a small-scale sensitivity
check on a fixed six-task subset shared by \texttt{GPT-6-astra} and
\texttt{GPT-5.6-sol}. Varying reconstruction thresholds, appearance
tolerances, and reconstruction point budgets changes absolute values as
expected, but does not change the model ordering on this subset.

\subsection{Aggregation and Failure Accounting}
\label{app:aggregation}

The main table averages over all attempted benchmark views within each declared
Indoor or Outdoor split. For completed evaluations, $S_i = V_i(R_i + A_i)/2$;
artifact-invalid trials receive $S_i = 0$. In the main table, submissions with
usable artifacts but unresolved headline-evaluation failures are also assigned
$V_i = R_i = A_i = S_i = 0$ and remain in the denominator as explicit failure
penalties rather than measured scores. Conditional diagnostic analyses may use
restricted evaluated subsets, but must report both eligible and evaluated
counts.

\subsection{Human-Metric Alignment}
\label{app:human_metric_alignment}

To assess whether the headline metrics align with human judgment, we conduct a
blinded A/B/Tie study on fixed candidate pairs. Each task presents one natural
reference image and two full-scene candidate reconstructions. Reconstruction
tasks ask annotators to compare four numbered objects using shape, size, pose,
placement, visible structure, and missing geometry, while disregarding color,
material, texture, and lighting. Appearance tasks ask annotators to compare
four numbered circular regions using only local visual similarity. Each
annotator first completes one worked example for each task type and then
annotates 40 assigned scenes from controlled-overlap batches: 20
Reconstruction and 20 Appearance.

\begin{figure*}[t]
\centering
\includegraphics[width=\textwidth]{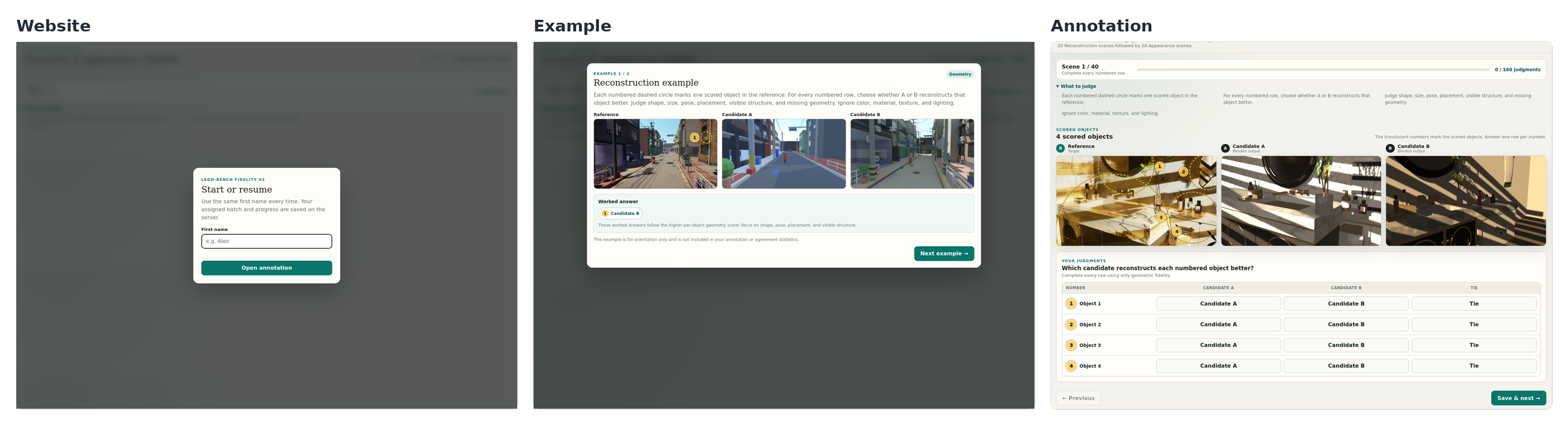}
\caption{\textbf{Human study interface used for metric validation.} Left:
annotator login. Middle: a worked example shown before annotation. Right: the
main annotation page, where the annotator compares two blinded full-scene
reconstructions against a shared reference and records one A/B/Tie judgment
for each marked object or region.}
\label{fig:human_metric_alignment_workflow}
\end{figure*}

We report results over six completed annotators in total 240 samples and 960 local area on those samples. Our primary inter-annotator
measure is \emph{compatibility agreement}, which counts \texttt{Tie-A},
\texttt{Tie-B}, and \texttt{Tie-Tie} as agreement and only a direct
\texttt{A-B} conflict as disagreement. Compatibility agreement is 86.6\%
overall, with 88.9\% on Reconstruction and 84.3\% on Appearance. 

For human-metric comparison, human labels are aggregated per scored target,
and an equal number of A and B votes is treated as \texttt{Both OK}. On the
metric side, we likewise assign \texttt{Both OK} whenever the absolute score
gap between the two candidates falls below a fixed uncertainty threshold,
$|s_A - s_B| < \varepsilon$, with $\varepsilon = 0.05$. Under this rule,
human-metric compatibility is 83.7\%, indicating that the headline
metrics track human preferences in most cases.

\subsection{Articulation Protocol}

Articulation evaluation asks whether a predicted scene recovers the
benchmark's canonical articulated structure for each matched object, rather
than whether it admits any visually plausible mechanism. Because global scene
pose is not part of the functional target, movable parts are first matched in
an object-centric frame. The articulation score then combines three factors:
movable-part structure F1, matched-part surface quality, and a
quality-weighted joint F1 that evaluates parent--child structure, joint type,
motion trajectory, and observed state:
\begin{equation}
    S_{\mathrm{Art}}
    =
    \operatorname{HMean}
    (S_{\mathrm{struct}},S_{\mathrm{geom}},S_{\mathrm{kin}}).
\end{equation}
Unmatched predicted or ground-truth parts contribute false positives or false
negatives, and assigning motion to a human-confirmed static object is counted
as an explicit error. We additionally report Worst-Mover, movable recall,
false-move rate, axis/state error, and complete-object success as diagnostics.

\section{Experimental Setup and Supplementary Results}
\label{app:experiment_details}

\subsection{Common Runtime and Artifacts}

All formal experiments run in isolated Harbor tasks with Blender~5.0.1 and
Blender-MCP. For each campaign, we freeze the agent, harness, prompt,
evaluator, timeout policy, and container image. The actor receives a single
public RGB image and no private depth, segmentation, or metric feedback.
The main LEGO-Bench coding-agent campaigns use three independent runs per
model configuration; main tables report mean and standard-deviation summaries
over these repeated runs.

All coding-agent campaigns use the same base user prompt shown in
Figure~\ref{prompt:user-prompt-0}.

\begin{promptfigure}
  {Vanilla user prompt}
  {The vanilla user prompt for LEGO-Bench.}
  {prompt:user-prompt-0}
  You are given a single RGB reference image.

  Reconstruct the depicted scene in Blender using the available Blender MCP tools.
  Create a valid, editable 3D scene that represents the visible environment and
  major objects in the reference image.

  The submission must contain actual 3D scene geometry rather than a rendered-image approximation. Major visible objects should exist as separate scene objects where appropriate.

  Save the final Blender scene and export all required submission artifacts before finishing.
\end{promptfigure}

Evaluation is performed on the submitted artifacts, including
\texttt{scene.blend}, \texttt{scene.glb}, and \texttt{final.png} when
available. Failed or incomplete attempts remain in scope. Main-table
aggregation follows Appendix~\ref{app:aggregation}; conditional diagnostic
subsets are reported separately.

\subsection{Method Configurations and Information Access}
\label{app:baseline_protocol}

The main table compares end-to-end systems: general-purpose coding agents,
task-specific LLM systems, and single-image scene-construction baselines.
These systems need not share the same tools, native outputs, or recovery
procedures. The shared component is the evaluation endpoint: each method is
scored by the same frozen evaluator on each supported task after any declared
artifact conversion. This equalizes scoring, but not inference budget or
information access.

For the non-Codex rows, we preserve each method's native generation pipeline
and report only the tasks it supports. VIGA is
evaluated in its procedural-only setting with external asset generation
disabled. SceneConductor and REST3D are evaluated only on Indoor tasks in our
setup, while SceneGen and Gen3DSR are evaluated on both Indoor and Outdoor
splits. 3D-RE-GEN is also Indoor-only; because it outputs calibrated PLY
geometry rather than a Blender scene, we convert its outputs to Blender using
public camera intrinsics, a fixed camera convention, and fixed area lighting
before Appearance evaluation.

\subsection{VLM Judge Setup and Reference Results}
\label{app:vlm_judge_protocol}

The VLM judge is read-only: it sees only the reference image and candidate
renders, not system identity, transcripts, private depth, segmentation, or
LEGO-Bench scores. We evaluate \texttt{GPT-6-astra}, \texttt{GPT-6-sol},
\texttt{GPT-6-luna}, and the three \texttt{GPT-5.6} models as both builders
and judges. The comparison contains 360 checkpoint pairs, 60 from each builder,
and 2,160 primary judgments from the full $6\times6$ matrix.

Each builder contributes one pair from each of 60 distinct, artifact-valid
main-experiment trajectories from runs two and three. We select 20 pairs each
with sequence gaps of 1--2, 3--5, and at least 6, balancing run, difficulty,
and environment deterministically. Both checkpoint scenes and renders must
exist, and their scene signatures and image hashes must differ. Pairs with
unavailable deterministic endpoint metrics are replaced under the declared
validity rule before judging, preserving builder and gap bucket. Selection
does not compare score values or judge preferences.

Agreement is an exact match to the deterministic metric direction, using
$\epsilon=10^{-6}$ for ties. \texttt{both\_bad} and terminal judge failures
count as incorrect; all 60 pairs remain in each cell's denominator.
There are 0 terminal failures among the 2,160 primary judgments.
An additional 36 pairs are evaluated in reversed order by every judge,
yielding 216 audit judgments. Order-swap consistency ranges from
$69.4\%$ to
$100.0\%$ for Reconstruction and
$61.1\%$ to
$83.3\%$ for Appearance.

\begin{table*}[htbp]
    \legotablestyle[4pt]
    \caption{\textbf{Builder--judge directional agreement (\%).}
    Each cell reports Reconstruction / Appearance agreement on 60
    archived checkpoint pairs. Rows are builders; columns are judges.
    \texttt{both\_bad} and failed judgments count as incorrect.}
    \label{tab:appendix_judge_matrix}
    \resizebox{\textwidth}{!}{
    \begin{tabular}{l|cccccc}
        \legotabletoprule
        \textbf{Builder / Judge} & \texttt{GPT-6-astra} & \texttt{GPT-6-sol} & \texttt{GPT-6-luna} & \texttt{GPT-5.6-sol} & \texttt{GPT-5.6-terra} & \texttt{GPT-5.6-luna} \\
        \midrule
        \texttt{GPT-6-astra} & 61.7 / 76.7 & 51.7 / 78.3 & 50.0 / 83.3 & 58.3 / 86.7 & 51.7 / 76.7 & 53.3 / 78.3 \\
        \texttt{GPT-6-sol} & 45.0 / 75.0 & 45.0 / 78.3 & 46.7 / 76.7 & 43.3 / 75.0 & 48.3 / 81.7 & 45.0 / 63.3 \\
        \texttt{GPT-6-luna} & 51.7 / 71.7 & 56.7 / 76.7 & 43.3 / 66.7 & 46.7 / 71.7 & 41.7 / 70.0 & 41.7 / 65.0 \\
        \texttt{GPT-5.6-sol} & 41.7 / 48.3 & 45.0 / 60.0 & 38.3 / 53.3 & 46.7 / 48.3 & 46.7 / 55.0 & 46.7 / 46.7 \\
        \texttt{GPT-5.6-terra} & 45.0 / 58.3 & 40.0 / 66.7 & 36.7 / 56.7 & 40.0 / 70.0 & 43.3 / 65.0 & 43.3 / 55.0 \\
        \texttt{GPT-5.6-luna} & 41.7 / 51.7 & 43.3 / 46.7 & 45.0 / 35.0 & 35.0 / 43.3 & 41.7 / 41.7 & 35.0 / 38.3 \\
        \legotablebottomrule
    \end{tabular}}
\end{table*}

\begin{tcolorbox}[
  promptboxstyle,
  breakable,
  title={Judge Prompt}
]
\small
All judges use the same prompt template:

You are a blind evaluator of a rendered 3D reconstruction. Image 1 is the
original reference. Image 2 is candidate A (left) and image 3 is candidate B
(right). Judge only visible evidence in these images. Model identity, run
identity, critique text, and benchmark metrics are unavailable.

Reconstruction: Which candidate better matches the reference geometry? Judge
camera/viewpoint, object presence, layout, scale, shape, silhouette, and
occlusion. Ignore color, material, texture, and lighting as much as possible.

Appearance: Which candidate better matches the reference appearance? Judge
color, material, texture, lighting, shadows, and overall visual character. Do
not reward geometry differences except where they prevent appearance judgment.

For each question choose exactly one of: \texttt{left}, \texttt{right},
\texttt{tie}, or \texttt{both\_bad}. Return only the requested JSON object.
Do not include an explanation, score, or confidence.
\end{tcolorbox}

\subsection{Reasoning-Effort Protocol}
\label{app:reasoning_effort_protocol}

We study test-time scaling with the native Low, Medium, High, and
XHigh reasoning settings. The sweep covers \texttt{GPT-5.6-luna},
\texttt{GPT-5.6-terra}, \texttt{GPT-5.6-sol}, \texttt{GPT-6-astra},
\texttt{GPT-6-sol}, and \texttt{GPT-6-luna} under the Codex harness on the
42-task Office subset, comprising seven room types,
three complexity levels, and two paired views per room--level group. This
produces a balanced $6\times4\times42$ design with one scored outcome per
task cell. When infrastructure failures occur, we rerun the same
model--effort--task configuration, but do not use retries for best-of
selection.
Within each model, task inputs, prompts, execution budgets, and scoring are
held fixed across effort settings. All 1,008 outcomes are rescored with the
same frozen evaluator and container image. Figure intervals are pointwise
95\% percentile bootstrap intervals for the mean over the 42 task inputs,
using 10,000 resamples; they do not measure variation across repeated runs.

The new \texttt{GPT-6-sol} and \texttt{GPT-6-luna} sweeps contribute 336
outcomes, all with completed evaluations. Their High results are generated
specifically for this sweep, whose prompts disable the extra camera guidance
used in the main-table campaigns. Thirteen Luna trials retain native
nonzero-exit flags but yield artifacts that are successfully rescored; these
outcomes remain in the fixed denominator. Terminal evaluation failures would
receive zero scores rather than being dropped. Worker concurrency and
shared-host disk contention varied during this campaign, so its elapsed
times are not treated as a controlled measure of reasoning cost.

\section{LEGO-Plugin: Implementation and Controlled Evaluation}
\label{app:plugin_details}

\subsection{Runtime Boundary}
LEGO-Plugin runs alongside the standard Blender MCP. Blender MCP remains the only general scene editor and the agent remains the planner; the plugin exposes only bounded operations for measurement, validation, edit transactions, and evidence retrieval. It adds no unrestricted code-execution path and never modifies the scene without an explicit agent call. All services communicate through fixed JSON schemas, and large payloads are stored by content hash.

\subsection{Shared Reference Evidence}
Both Enhanced Initialization and Grounded Refinement draw on evidence extracted from the reference image only: camera and scene geometry from \texttt{VGGT}~\citep{wang2025vggt}, and optionally object masks from \texttt{SAM~3}~\citep{carion2026sam3segmentconcepts} and relative depth from \texttt{Depth Anything V2}~\citep{yang2024depthanythingv2}. Providers return image regions and depth maps; the agent, not the provider, decides which Blender entity each region corresponds to.

\subsection{Enhanced Initialization}
The initialization service estimates a gravity-aligned, Z-up Manhattan frame from \texttt{VGGT}, constructs a landscape and installs the room and camera together after validation. It then produces a construction plan recording, for each visible object, its normalized centroid, projected extent, relative and ordinal depth, and observability flags. The plan deliberately avoids metric depth or absolute object size, which cannot be reliably recovered from a single monocular image. Outputs include the \texttt{VGGT} predictions, the recovered room and camera, a rendered room shell, and an initialization report.

\subsection{Version Control}
Each edit is wrapped in a transaction that declares editable entities, protected entities, allowed types of change, and acceptance requirements. Before the edit, the plugin snapshots the relevant scene state, including transforms, meshes, materials, hierarchy, visibility, active camera, and render settings. After the Blender MCP edit, it rejects any change to protected or undeclared entities. A transaction commits only if it produces a non-empty allowed change and all hard requirements pass on the current scene ; otherwise it is rolled back by removing new entities and restoring the snapshot. Evaluation runs allow at most eight transactions and two rollbacks per case to prevent unbounded repair loops.

\subsection{Grounded Refinement}
Validation checks named objects or collections against explicit requirements and returns \emph{pass}, \emph{fail}, or \emph{unknown}. Supported requirements cover existence, valid transforms and dimensions, support, containment, collision, facing direction, repeated layout, camera visibility, projected centroid and extent, reference depth, depth order, and rendered luminance. Failures of hard requirements, or missing evidence for them, block acceptance, while other failures are reported as warnings. Each report records the requirements, measurements, and residuals, and is tied to the exact scene state: any later scene change invalidates it. Each candidate receives at most two correction rounds before it must be committed or rolled back.

\subsection{Execution Records}
As supporting infrastructure rather than a separate module, every plugin call is logged with its request, response, Blender execution status, transaction state, and resulting scene signature, so that a failed backend call can never be recorded as a successful edit. Runtime hooks prevent the agent from finishing while validations are stale, transactions are unresolved, or renders are unreviewed. All camera renders are logged independently of the prompt. At the end of each run, the plugin exports \texttt{scene.blend}, \texttt{scene.glb}, and a $1920{\times}1080$ render without modifying scene geometry.
\section{LEGO-Anything: Natural-Image Task Protocols}
\label{app:lego_anything_details}

\subsection{Task Inputs and Frozen-Scene Readouts}

The main paper evaluates \texttt{GPT-6-astra} without LEGO-Plugin on three fixed
100-image tracks: COCO val2017 detection, LVIS v1 validation instance
segmentation, and ETH3D relative depth. The agent constructs a scene before
evaluation; task predictions are read from the frozen artifact.
These experiments test a common executable-scene interface across datasets,
not whether a single scene has already been evaluated against all three
annotation types on the same image.

The readout projects semantic/instance geometry into visible-object boxes and
masks and renders camera-space depth. Ground-truth annotations and metric
feedback belong to the verifier and must not be returned to the agent.
The intended category vocabulary and task schema are public inputs; exact
camera metadata and task-dependent prompt differences must be recorded.

\subsection{Detection and Segmentation: Confidence and Missing Predictions}
\label{app:downstream_ap}

The archived direct baselines are \texttt{DINO-R50-4scale-24ep} for boxes and \texttt{SAM~3}
for instance masks. The COCO comparison uses
\path{pycocotools.cocoeval.COCOeval} with at most 100 detections per image;
the LVIS comparison uses \path{lvis.LVISEval} with at most 300 detections
and its federated-label evaluation.

LEGO-Anything's scene schema does not provide a calibrated confidence score.
The AP export therefore assigns every predicted instance a confidence of
1.0 and uses a deterministic tie order based on image ID, category ID, and
the mask/box payload. The main table reports this \emph{equal-confidence AP}.
It measures the resulting ranking under that declared policy, not calibrated
detector confidence. Missing scene predictions are exported as no detections
rather than dropping their images.
Because LEGO-Anything emits deterministic scene-derived predictions
without calibrated confidence scores, AP should be read as a compatibility
diagnostic under a fixed ordering rather than as a calibrated ranking
comparison against detectors or segmenters with learned confidence scores.

Both scene-based AP tracks have 94 scored trials out of 100 attempted images.
All 100 images remain in the official subset AP evaluation. The separate
score-free matched-IoU diagnostic has 93 paired quality-valid images:
one image in each subset contains no scored ground-truth instances, and
the remaining missing predictions further restrict paired coverage.

\subsection{Relative Depth and Coverage}
\label{app:downstream_depth}

The direct depth baseline is \texttt{DA3MONO-LARGE}. Evaluation uses the frozen
inverse-depth scale-and-shift alignment protocol, so this is a relative-depth
comparison rather than a test of absolute metric scale.
The main table's direct AbsRel is averaged over 100 images, whereas
LEGO-Anything has 99 quality-valid depth outputs. A missing output cannot be
assigned zero error for a lower-is-better metric.

On the same 99 valid images, the archived direct and scene-based AbsRel
means are 0.07845 and 0.15540, respectively. This paired diagnostic addresses
the coverage difference while retaining the separate 99/100 completion
statement. Table~\ref{tab:appendix_downstream_coverage} makes the distinct
denominators explicit.

\begin{table}[htbp]
    \legotablestyle
    \caption{\textbf{Coverage of the natural-image comparisons.}
    Official box/mask AP retains all 100 subset images, including missing
    scene predictions. Matched diagnostics use only jointly valid pairs.}
    \label{tab:appendix_downstream_coverage}
    \begin{tabular}{l|ccc}
        \legotabletoprule
        \textbf{Track} & \textbf{Attempted images}
        & \textbf{Scene outputs scored} & \textbf{Matched diagnostic pairs} \\
        \midrule
        COCO boxes & 100 & 94 & 93 \\
        LVIS masks & 100 & 94 & 93 \\
        ETH3D depth & 100 & 99 & 99 \\
        \legotablebottomrule
    \end{tabular}
\end{table}

The raw standard-evaluator metrics, equal-confidence exports, and paired
comparison summaries are preserved with source hashes in the ancillary file
\path{appendix_verified_evidence.json}.

\section{Evaluation Infrastructure and Reproducibility}
\label{app:evaluation_infrastructure}

This section summarizes the evaluation infrastructure used by the reported
experiments. Adapter support alone does not imply that a corresponding model
configuration appears in the paper; reported runs are identified by the frozen
manifests in Appendix~\ref{app:experiment_details}.

\paragraph{\textbf{Harbor extension.}}
We build on Harbor~\citep{Harbor_Framework} for task expansion, repetitions,
logging, artifact collection, verifier execution, and aggregation. Our
\texttt{lego\_bench\_harbor} package adds the dataset converter, isolated
Blender runtime, agent adapters, verifier, and run manifest. Public inputs are
mounted for the actor, while private geometry, masks, depth, and
correspondences remain verifier-only.

\paragraph{\textbf{Common runtime.}}
Formal runs use isolated Docker/Compose environments with Blender~5.0.1,
Xvfb, and one Blender MCP listener per trial. Every harness receives the same
image, prompt, taxonomy, tool interface, limits, and artifact contract. The
runtime collects the submitted scene artifacts, transcripts, and verifier
reports, and applies the same prompt-independent finalization policy on
timeouts.

\paragraph{\textbf{Codex.}}
\texttt{ContainerCodex} subclasses Harbor's Codex adapter
~\citep{codexagentloop}, registers the MCP tools in an isolated
\texttt{CODEX\_HOME}, and attaches the input image. Final campaigns set
\texttt{model\_provider=amazon-bedrock}; GPT-family identifiers therefore do
not by themselves specify the serving path.

\paragraph{\textbf{Hermes.}}
Our wrapper uses Harbor's generic adapter and the upstream Nous Research
Hermes Agent~\citep{nousHermesAgent}, pinned to revision
\texttt{aa6f77596b}. \texttt{ContainerHermes} supports local-model
diagnostics, while \texttt{ContainerBedrockHermes} uses the Bedrock Converse
provider with the same MCP tools and session export. Only configurations with
frozen run manifests are reported.

\end{document}